\documentclass{article}

\usepackage[accepted]{icml2018}

\usepackage{color}
\usepackage[usenames,dvipsnames]{xcolor}

\usepackage{natbib}
\usepackage[utf8]{inputenc} 
\usepackage[T1]{fontenc}    

\usepackage{url}            
\usepackage{booktabs}       
\usepackage{amsfonts}       
\usepackage{nicefrac}       

\usepackage{algorithm}
\usepackage{algorithmic}
\usepackage{standalone}
\usepackage{tikz}
\usepackage{setspace}
\usepackage{amsmath}
\usepackage{amssymb}
\usepackage[makeroom]{cancel}
\usepackage[shortlabels]{enumitem}
\usepackage{subcaption}
\usepackage{refcount}
\usepackage{fixltx2e}
\usepackage{xspace}
\usepackage{thmtools}
\usepackage{thm-restate}
\usepackage{mathtools} 
\newtheorem{assumption}{Assumption}
\usepackage{hyperref}       
\hypersetup{
  pdftex,
  pdffitwindow=true,
  pdfstartview={FitH},
  pdfnewwindow=true,
  colorlinks,
  linktocpage=true,
  linkcolor=Green,
  urlcolor=Green,
  citecolor=Green
}

\usetikzlibrary{positioning,shadows,arrows,trees,shapes,fit,decorations.pathreplacing}

\usepackage[textsize=tiny,
]{todonotes}	

\newcommand{\sg}[1]{\todo[size=\scriptsize,color=green!50!white]{Steffen: #1}}

\newenvironment{proof}{\emph{Proof:}}{\hfill$\square$}

\usepackage[capitalize]{cleveref}
\usepackage{thmtools}
\usepackage{thm-restate}

\declaretheorem[name=Theorem,refname={Theorem,Theorems},Refname={Theorem,Theorems}]{theorem}

\declaretheorem[name=Lemma,refname={Lemma,Lemmas},Refname={Lemma,Lemmas},sibling=theorem]{lemma}

\declaretheorem[name=Corollary,refname={Corollary,Corollaries},Refname={Corollary,Corollaries},sibling=theorem]{corollary}

\declaretheorem[name=Proposition,refname={Proposition,Propositions},Refname={Proposition,Propositions},sibling=theorem]{proposition}

\Crefname{question}{Question}{Questions}
\creflabelformat{question}{(#2#1#3)}

\Crefname{problem}{Problem}{Problems}
\creflabelformat{problem}{(#2#1#3)}

\Crefname{equation}{Equation}{Equations}
\creflabelformat{equation}{(#2#1#3)}

\crefname{iCondition}{Condition}{Conditions}
\creflabelformat{iCondition}{(#2#1#3)}
\crefrangelabelformat{iCondition}{(#3#1#4) to (#5#2#6)}

\Crefname{item}{}{}
\creflabelformat{item}{(#2#1#3)}
\crefrangelabelformat{item}{(#3#1#4) to (#5#2#6)}

\newcommand{\MABDAAF}{\text{MABDA\!\!AF}\xspace}
\newcommand{\ODAAF}{\text{ODA\!\!AF}\xspace}
\newcommand{\one}{\mathbb{I}} 
\newcommand{\onep}[1]{\mathbb{I}\left\{#1\right\}} 
\newcommand{\Reg}{\mathfrak{R}} 
\newcommand{\PP}{\mathbb{P}}
\newcommand{\Prob}[1]{\mathbb{P}\left(#1\right)}
\newcommand{\E}{\mathbb{E}}
\newcommand{\EE}[1]{\mathbb{E}[#1]}

\newcommand{\Var}{\mathbb{V}}
\newcommand{\cA}{\mathcal{A}}
\newcommand{\cG}{\mathcal{G}}
\newcommand{\cF}{\mathcal{F}}

\newcommand{\timeset}{{ T}}

\newcommand{\event}{{\cal E}}
\newcommand{\Ex}{\mathbb{E}}
\newcommand{\removedtmp}[1]{}

\allowdisplaybreaks

\icmltitlerunning{Bandits with Delayed, Aggregated Anonymous Feedback}

\begin{document}

\twocolumn[
\icmltitle{Bandits with Delayed, Aggregated Anonymous Feedback}



\icmlsetsymbol{equal}{*}

\begin{icmlauthorlist}
\icmlauthor{Ciara Pike-Burke}{la}
\icmlauthor{Shipra Agrawal}{co}
\icmlauthor{Csaba Szepesv\'ari}{dm,uofa}
\icmlauthor{Steffen Gr\"unew\"alder}{la}
\end{icmlauthorlist}

\icmlaffiliation{la}{
Department of Mathematics and Statistics,
Lancaster University, Lancaster, UK}
\icmlaffiliation{dm}{
DeepMind, London, UK}
\icmlaffiliation{uofa}{
Department of Computing Science, University of Alberta, Edmonton, AB, Canada}
\icmlaffiliation{co}{
Department of Industrial Engineering and Operations Research, Columbia University, New York, NY, USA}

\icmlcorrespondingauthor{Ciara Pike-Burke}{ciara.pikeburke@gmail.com}

\icmlkeywords{Machine Learning, ICML}

\vskip 0.3in
]



\printAffiliationsAndNotice{}  

\begin{abstract}
We study 
a variant of the stochastic $K$-armed bandit problem, which we call
``bandits with delayed, aggregated anonymous feedback''.
In this problem, when the player pulls an arm, a reward is generated, however it is not immediately observed. Instead, at the end of each round the player observes only the sum of a number of previously generated rewards which happen to arrive in the given round. The rewards are stochastically delayed and due to the aggregated nature of the observations, the information of which arm led to a particular reward is lost. The question is what is the cost of the information loss due to this delayed, aggregated anonymous feedback? Previous works have studied bandits with stochastic, non-anonymous delays and found that the regret increases only by an additive factor relating to the expected delay. 
In this paper, we show that this additive regret increase can be maintained in the harder delayed, aggregated anonymous feedback setting when the expected delay (or a bound on it) is known. We provide an algorithm that matches the worst case regret of the non-anonymous problem exactly when the delays are bounded, and up to logarithmic factors or an additive variance term for unbounded delays. 
\end{abstract}

\section{Introduction}



\begin{figure*}
 \centering
  \begin{tikzpicture}
	 \draw (3,1) node[text width=5cm, align=center] {Multi-Armed Bandits \\ (eg. \citet{auer2002finite}) \\ $O(\sqrt{KT\log T})$};
	 \draw (7.75,1) node[text width=4.5cm, align=center] {Delayed Feedback Bandits (eg. \citet{joulani2013online}) \\ $O(\sqrt{KT\log T} + K\E[\tau])$};
	 \draw (12.75,1) node[text width=4.75cm, align=center] {Bandits with Delayed, Aggregated Anonymous Feedback \\ $O(\sqrt{KT\log K} + K\E[\tau])$};
	 \draw [->, line width= 2] (1,0.2)--(15,0.2);
	 \draw (7.5,-0.1) node[align=center] {\emph{Difficulty}};
\end{tikzpicture}
  \label{fig:difficulties}
  \caption{The relative difficulties and problem independent regret bounds of the different problems. For \MABDAAF, our algorithm uses knowledge of $\E[\tau]$ and a mild assumption on a delay bound, which is not required by \citet{joulani2013online}.}
  \label{fig:diff}
\end{figure*}

The stochastic multi-armed bandit (MAB) problem is a prominent framework for capturing the exploration-exploitation tradeoff in online decision making and experiment design.
The MAB problem proceeds in discrete sequential rounds, where in each round, the player pulls one of the $K$ possible arms. In the classic stochastic MAB setting, the player immediately observes stochastic feedback from the pulled arm in the form of a `reward' which can be used to improve the decisions in subsequent rounds. One of the main application areas of MABs is in online advertising. Here, the arms correspond to adverts, and the feedback would correspond to \emph{conversions}, that is users buying a product after seeing an advert. However, in practice, these conversions may not necessarily happen immediately after the advert is shown, and it may not always be possible to assign the credit of a sale to a particular showing of an advert. A similar challenge is encountered in many other applications, e.g., in personalized treatment planning, where the effect of a treatment on a patient's health may be delayed, and it may be difficult to determine which out of several past treatments caused the change in the patient's health; or, in content design applications, where the effects of multiple changes in the website design on website traffic and footfall may be delayed and difficult to distinguish. 

In this paper, we propose a new bandit model to handle online problems with such `delayed, aggregated and anonymous' feedback. 
In our model, a player interacts with an environment of $K$ actions (or arms) in a sequential fashion. 
At each time step the player selects an action which leads to a reward generated at random from the underlying reward distribution.
At the same time, a nonnegative random integer-valued delay 
is also generated i.i.d. from  an underlying delay distribution. 
Denoting this delay by $\tau\ge 0$ and the index of the current round by $t$, 
the reward generated in round $t$ will arrive at the end of the $(t+\tau)$th round.
At the end of each round, the player observes only the \emph{sum} of all the rewards that arrive in that round. Crucially, the player does not know which of the past plays have contributed to this aggregated reward. 
We call this problem \emph{multi-armed bandits with delayed, aggregated anonymous feedback} (\MABDAAF).
As in the standard MAB problem, in \MABDAAF, the goal is to maximize the cumulative reward from $T$ plays of the bandit, or equivalently to minimize the regret. The regret is the total difference between the reward of the optimal action and the actions taken.

If the delays are all zero, the \MABDAAF problem reduces to the standard (stochastic) MAB problem,
which has been studied considerably
\citep[e.g.,][]{thompson1933likelihood, lai1985asymptotically, auer2002finite, BC12}. 
Compared to the MAB problem, the job of the player in our problem appears to be significantly more difficult since the player has to deal with 
{\em (i)} that some feedback from the previous pulls may be \emph{missing} due to the delays, and
{\em (ii)} that the feedback takes the form of the sum of an \emph{unknown number} of rewards of \emph{unknown origin}.

\removedtmp{A natural question to ask is whether the problem would become easier if some of these obstacles were removed. 
For example, one can imagine that in some applications the rewards are individually observed (rather than only observing their sum), but the player still does not know which arm generated each reward.
In this setting, the player has more information. However, it is not clear that this will actually help reduce the regret.
Consider binary rewards and little knowledge of the delay
distributions. Here, attributing the rewards to the arms that generated them is still not possible. Hence, the only extra knowledge is how many rewards are received, and it is unclear how to use this.}

An easier problem is when the observations are delayed, but they are \emph{non-aggregated} and \emph{non-anonymous}: that is, the player has to only deal with challenge (i) and not (ii).
Here, the player receives delayed feedback in the shape of action-reward pairs that inform 
the player of both the individual reward and which action generated it. 
This problem, which we shall call the  \emph{(non-anonymous) delayed feedback bandit problem}, 
has been studied by \citet{joulani2013online}, and later followed up by \citet{mandel2015queue} (for bounded delays). 
Remarkably, they show that compared to the standard (non-delayed) stochastic MAB setting, 
the regret will increase only additively by a factor that scales with the expected delay.
For delay distributions with a finite expected delay, $\E[\tau]$, the worst case regret scales with
 $O(\sqrt{KT \log T } +K\E[\tau])$. Hence, the price to pay for the delay in receiving the observations is negligible.
QPM-D of \citet{joulani2013online} and SBD of \citet{mandel2015queue} place received rewards into queues for each arm, taking one whenever a base bandit algorithm suggests playing the arm. 
Throughout, we take UCB1 \cite{auer2002finite} as the base algorithm in QPM-D. \citet{joulani2013online} also present a direct modification of the UCB1 algorithm. All of these algorithms achieve the stated regret.
None of them require \emph{any} knowledge of the delay distributions, but they all rely heavily upon the non-anonymous nature of the observations. 

While these results are encouraging, 
the assumption that the rewards are observed individually in a non-anonymous fashion is limiting for most practical applications with delays (e.g., recall the applications discussed earlier).
How big is the price to be paid for receiving only aggregated anonymous feedback?
Our main result is to prove that essentially there is no extra price to be paid provided 
that the value of the expected delay (or a bound on it) is available.
In particular, this means that detailed knowledge of which action led to a particular delayed 
reward can be replaced by the much weaker requirement that the expected delay, or a bound on it, is known.
\cref{fig:diff} summarizes
the relationship between the non-delayed, the delayed and the new problem by showing the leading terms of the regret. 
In all cases, the dominant term is $\sqrt{KT}$. Hence,  asymptotically, 
the delayed, aggregated anonymous feedback problem is no more difficult than the standard multi-armed bandit problem.

\subsection{Our Techniques and Results}
We now consider what sort of algorithm will be able to achieve the aforementioned results for the \MABDAAF problem.
Since the player only observes delayed, aggregated anonymous rewards, the first problem we face
 is how to even estimate the mean reward of individual actions.
Due to the delays and anonymity, it appears that to be able to estimate the mean reward of an action,
the player wants to have played it consecutively for long stretches. 
Indeed, if the stretches are sufficiently long compared to the mean delay,
the observations received during the stretch will mostly consist of rewards of the action played
in that stretch.
This naturally leads to considering algorithms that \emph{switch actions rarely} and this is indeed the basis
of our approach.

Several popular MAB algorithms are based on choosing 
the action with the largest upper confidence bound (UCB) in each round 
\citep[e.g.,][]{auer2002finite, cappe2013kullback}. 
UCB-style algorithms tend to switch arms frequently and will only play the optimal 
arm for long stretches if a unique optimal arm exists.
Therefore, for \MABDAAF, we will consider alternative algorithms where arm-switching is more tightly controlled.
The design of such algorithms goes back at least to the work of \citet{agrawal1988asymptotically} where the problem of bandits with switching costs was studied.
The general idea of these rarely switching algorithms is to gradually eliminate suboptimal arms by playing arms in phases and comparing each arm's upper confidence bound to the lower confidence bound of a leading arm at the end of each phase.
Generally, this sort of rarely switching algorithm switches arms only $O(\log T)$ times.
We base our approach on one such algorithm, the so-called Improved UCB\footnote{The adjective ``Improved'' indicates that the algorithm improves upon the regret bounds achieved by UCB1. The improvement replaces $\log(T)/\Delta_j$ by $\log(T\Delta_j^2)/\Delta_j$ in the regret bound.}
 algorithm of \citet{auer2010ucb}.
 
Using a rarely switching algorithm alone will not be sufficient for \MABDAAF.
The remaining problem, and where the bulk of our contribution lies,
is to construct appropriate confidence bounds and adjust the length of the periods of playing each arm to account
for the delayed, aggregated anonymous feedback.
In particular, in the confidence bounds attention must be paid to fine details:
it turns out that unless the variance of the observations is dealt with, there is
a blow-up by a multiplicative factor of $K$. We avoid this by an improved analysis involving Freedman's inequality \citep{Fre75}. 
Further, to handle the dependencies between the number of plays of each arm and the past rewards, we combine Doob's optimal skipping theorem \citep{Doob53} and Azuma-Hoeffding inequalities. 
Using a rarely switching algorithm for \MABDAAF means we must also consider the dependencies between the elimination of arms in one phase and the corruption of observations in the next phase (ie. past plays can influence both whether an arm is still active and the corruption of its next plays). We deal with this through careful algorithmic design.

Using the above, we provide an algorithm that achieves worst case regret of $O(\sqrt{KT\log K} + K\E[\tau] \log T)$ using only knowledge of the expected delay, $\E[\tau]$. We then show that this regret can be improved by using a more careful martingale argument that exploits the fact that our algorithm is designed to remove most of the dependence between the corruption of future observations and elimination of arms.
 Particularly, if the delays are bounded with known bound $0\leq d\leq \sqrt{T/K}$, we can recover worst case regret of $O(\sqrt{KT\log K} + K\E[\tau])$, matching that of \citet{joulani2013online}. 
If the delays are unbounded but have known variance $\Var(\tau)$, 
we show that the problem independent regret can be reduced to $O(\sqrt{KT\log K} + K\E[\tau] + K\Var(\tau))$. 

\subsection{Related Work}
We have already discussed several of the most relevant works to our own. However, there has also been other work looking at different flavors of the bandit problem with delayed (non-anonymous) feedback. For example, \citet{neu2010online} and \citet{cesa2016delay} consider non-stochastic bandits with fixed constant delays; \citet{dudik2011efficient} look at stochastic contextual bandits with a constant delay and \citet{desautels2014parallelizing} consider Gaussian Process bandits with a bounded stochastic delay. The general observation that delay causes an additive regret penalty in stochastic bandits and a multiplicative one in adversarial bandits is made in \citet{joulani2013online}.
The empirical performance of $K$-armed stochastic bandit algorithms in delayed settings was investigated in \citet{chapelle2011empirical}.
A further related problem is the `batched bandit' problem studied by \citet{perchet2016batched}. Here the player must fix a set of time points at which to collect feedback on all plays leading up to that point. 
\citet{vernade2017stochastic} consider delayed Bernoulli bandits where some observations could also be censored (e.g., no conversion is ever actually observed if the delay exceeds some threshold) but require complete knowledge of the delay distribution.
Crucially, here and in all the aforementioned works, the feedback is always assumed to take the form of arm-reward pairs and knowledge of the assignment of rewards to arms underpins the suggested algorithms, rendering them unsuitable for \MABDAAF.
To the best of our knowledge, ours is the first work to develop algorithms to deal with delayed, aggregated anonymous feedback in the bandit setting.

\subsection{Organization}
The reminder of this paper is organized as follows:
In the next section (\cref{sec:def}) we give the formal problem definition.
We present our algorithm in \cref{sec:alg}. In \cref{sec:reg}, we discuss the performance of our algorithm under various delay assumptions; known expectation, bounded support with known bound and expectation, and known variance and expectation.
This is followed by a numerical illustration of our results in \cref{sec:experiments}. We conclude in \cref{sec:conc}.

\section{Problem Definition} \label{sec:def}
There are $K>1$ actions or arms in the set $\cA$.
Each action $j\in \cA$ is associated with a  reward distribution $\zeta_j$ and a delay distribution $\delta_j$.
The reward distribution is supported in $[0,1]$ and the delay distribution is supported on $\mathbb{N} \doteq \{0,1,\dots\}$.
We denote by $\mu_j$ the mean of $\zeta_j$, $\mu^* = \mu_{j^*} = \max_j \mu_j$ and define $\Delta_j = \mu^* - \mu_j$ to be the \emph{reward gap}, that is the expected loss of reward each time action $j$ is chosen instead of an optimal action. 
Let $(R_{l,j},\tau_{l,j})_{l \in \mathbb{N}, j\in \cA}$ be an infinite array of random variables defined on the probability space $(\Omega, \Sigma, P)$ which are mutually independent.
Further, $R_{l,j}$ follows the distribution $\zeta_j$ and $\tau_{l,j}$ follows the distribution $\delta_j$.
The meaning of these random variables is that 
if the player plays action $j$ at time $l$, a payoff of $R_{l,j}$ will be added to the aggregated feedback that the player receives at the end of the $(l + \tau_{l,j})$th play. 
Formally, if $J_l\in \cA$ denotes the action chosen by the player at time $l=1,2,\dots$, then the observation received 
at the end of the $t$th play is
\[ 
X_t = \sum_{l=1}^t \sum_{j=1}^K R_{l,j} \times \one \{l+\tau_{l,j} = t,  J_l = j  \}.
\]
For the remainder, we will consider i.i.d. delays across arms. We also assume discrete delay distributions, although most results hold for continuous delays by redefining the event $\{ \tau_{l,j} = t-l\}$ as $\{ t-l-1 < \tau_{l,j} \leq t-l\}$ in $X_t$. 
In our analysis, we will sum over stochastic index sets.
For a stochastic index set $I$ 
 and random variables $\{ Z_n \}_{n \in \mathbb{N}}$ we denote such sums as $\sum_{t \in I} Z_t \doteq \sum_{t \in \mathbb{N}} \one\{t \in I\} \times Z_t$.

\paragraph{Regret definition} In most bandit problems, the regret is the cumulative loss due to not playing an optimal action. In the case of delayed feedback, there are several possible ways to define the regret. One option is to consider only the loss of the rewards \emph{received} before horizon $T$ (as in \citet{vernade2017stochastic}). However, we will not use this definition.
Instead, as in \citet{joulani2013online}, we consider the loss of all \emph{generated} rewards and define the (pseudo-)regret by
\[ \Reg_T = \sum_{t=1}^T( \mu^* - \mu_{J_t}) =  T\mu^* - \sum_{t=1}^T \mu_{J_t}. \]
This includes the rewards received after the horizon $T$ and does not penalize large delays as long as an optimal 
action is taken. This definition is natural since, in practice, the player should eventually receive all outstanding
reward.

\citet{lai1985asymptotically} showed that the regret of any algorithm for the standard MAB problem must satisfy,
\begin{align}
\liminf_{T \to \infty} \frac{\E[\Reg_T]}{\log(T)} \geq \sum_{j: \Delta_j>0} \frac{\Delta_j}{KL(\zeta_j,\zeta^*)}, \label{eqn:lairobbins}
\end{align}
where $KL(\zeta_j, \zeta^*)$ is the KL-divergence between the reward distributions of arm $j$ and an optimal arm. 
Theorem 4 of \citet{vernade2017stochastic} shows that the lower bound in \eqref{eqn:lairobbins} also holds for delayed feedback bandits with no censoring and their alternative definition of regret.
We therefore suspect \eqref{eqn:lairobbins} should hold for \MABDAAF. However, due to the specific problem structure, finding a lower bound for \MABDAAF is non-trivial and remains an open problem. 

\paragraph{Assumptions on delay distribution} 
For our algorithm for \MABDAAF, we need some assumptions on the delay distribution. We assume that the expected delay, $\mathbb{E}[\tau]$, is bounded and known. This quantity is used in the algorithm.

\begin{assumption}
\label{assum:1}
The expected delay $\mathbb{E}[\tau]$ is bounded and known to the algorithm.
\end{assumption}

We then show that under some further mild assumptions on the delay, we can obtain better algorithms with even more efficient regret guarantees. We consider two settings: delay distributions with bounded support, and bounded variance. 
\begin{assumption}[Bounded support]
\label{assum:2}
There exists some constant $d > 0$ known to the algorithm such that the support of the delay distribution is bounded by $d$.
\end{assumption}

\begin{assumption}[Bounded variance]
\label{assum:3}
The variance, $\Var(\tau)$, of the delay is bounded and known to the algorithm.
\end{assumption}

In fact the known expected value and known variance assumption can be replaced by a `known upper bound' on the expected value and variance respectively. However, for simplicity, in the remaining, we use $\Ex[\tau]$ and $\Var(\tau)$ directly. The next sections provide algorithms and regret analysis for different combinations of the above assumptions.
\section{Our Algorithm} \label{sec:alg}

Our algorithm is a phase-based elimination algorithm based on the Improved UCB algorithm by \citet{auer2010ucb}. The general structure is as follows. In each phase, each arm is played multiple times consecutively. At the end of the phase, the observations received are used to update mean estimates, and any arm with an estimated mean below the best estimated mean by a gap larger than a `separation gap tolerance' is eliminated. This separation tolerance is decreased exponentially over phases, so that it is very small in later phases, eliminating all but the best arm(s) with high probability. An alternative formulation of the algorithm is that at the end of a phase, any arm with an upper confidence bound lower than the best lower confidence bound is eliminated. These confidence bounds are computed so that with high probability they are more (less) than the true mean, but within the separation gap tolerance. The phase lengths are then carefully chosen to ensure that the confidence bounds hold.
Here we assume that the horizon $T$ is known, but we expect that this can be relaxed as in \citet{auer2010ucb}.

\begin{algorithm}[t]
   \renewcommand{\algorithmicelsif}{\textbf{else while}}
\renewcommand{\algorithmicrequire}{\textbf{Input:}}
\renewcommand{\algorithmicensure}{\textbf{Initialization:}}
\renewcommand{\algorithmicprint}{\textbf{Select Algorithm:}}
\begin{algorithmic}
   \REQUIRE A set of arms, $\cA$; a horizon, $T$; choice of $n_m$ for each phase $m=1,2,\ldots$.
   \ENSURE Set $\tilde{\Delta}_1 = \nicefrac{1}{2}$ (tolerance), the set of active arms $\cA_1=\cA$.  Let $\timeset_i(1)=\emptyset, i\in A$, $m=1$ (phase index), $t=1$ (round index)
%
%
  \WHILE{$t\le T$}
   		\STATE \underline{Step 1: Play arms.}
   		\FOR{$j \in \cA_m$} 
            \STATE Let 
            				$\timeset_j(m)=\timeset_j(m-1)$
            \WHILE{$|\timeset_j(m)|\le n_m$ \textbf{ and } $t\le T$}
   			\STATE Play arm $j$, receive $X_t$. Add $t$ to $\timeset_j(m)$. Increment $t$ by $1$.\\
            \ENDWHILE          
		\ENDFOR
   		\STATE \underline{Step 2: Eliminate sub-optimal arms.}
		\STATE For every arm in $j\in \cA_m$, compute $\bar{X}_{m,j}$ as the average of observations at time steps $t\in \timeset_{j}(m)$. That is,
        $$\bar{X}_{m,j} = \frac{1}{|\timeset_j(m)|} \sum_{t\in \timeset_j(m)} X_t \,.$$
   		\STATE Construct $\cA_{m+1}$ by eliminating actions $j \in \cA_m$ with
   				\begin{align*}
                & \bar{X}_{m,j}+\tilde{\Delta}_m < \max_{ j' \in \cA_m} \bar{X}_{m,j'}  \,.
   				\end{align*}
		\vspace{-10pt}
   		\STATE \underline{Step 3: Decrease Tolerance.} \\
        Set $\tilde{\Delta}_{m+1} = \frac{\tilde{\Delta}_m}{2}$. \\
   		%
        \STATE \underline{Step 4: Bridge period.} \\
        Pick an arm $j\in \cA_{m+1}$ and play it $\nu_m = n_m - n_{m-1}$ times while incrementing $t \leq T$.
        Discard all observations from this period. Do not add $t$ to $T_j(m)$.
        \STATE Increment phase index $m$.
   	\ENDWHILE
\end{algorithmic}
   \caption{Optimism for Delayed, Aggregated Anonymous Feedback (\ODAAF)}
   \label{alg:skeleton}
\end{algorithm}

\paragraph{Algorithm overview}
Our algorithm, \ODAAF, is given in Algorithm~\ref{alg:skeleton}. It operates in phases $m=1,2,\ldots$. Define $\cA_m$ to be the set of active arms in phase $m$. The algorithm takes parameter $n_m$ which defines the number of samples of each active arm required by the end of phase $m$. 

In Step 1 of phase $m$ of the algorithm, each active arm $j$ is played repeatedly for $n_m-n_{m-1}$ steps.
We record all timesteps where arm $j$ was played in the first $m$ phases (excluding bridge periods) in the set $T_j(m)$.
The active arms are played in any arbitrary but fixed order.
 In Step 2, the $n_m$ observations from timesteps in $\timeset_j(m)$ are averaged to obtain a new estimate $\bar{X}_{m,j}$ of $\mu_j$. Arm $j$ is eliminated if $\bar{X}_{m,j}$ is further than $\tilde \Delta_m$ from $\max_{j' \in \cA_m} \bar{X}_{m,j'}$.

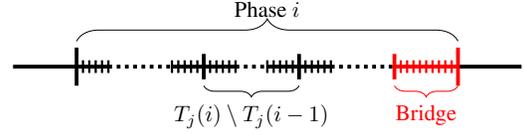
\begin{figure}
 \centering
  \begin{tikzpicture}
	\draw[-, line width=1.7] (1,1)--(2.5,1);
	\draw[dotted, line width=1.7] (2.5,1)--(3.5,1);
	\draw[-, line width=1.7] (3.5,1)--(4.5,1);
	\draw[dotted, line width=1.7] (4.5,1)--(5,1);
	\draw[-, line width=1.7] (5,1)--(6,1);
	\draw[dotted, line width=1.7] (6,1)--(7,1);
	\draw[-,line width=1.7,color=red] (7,1)--(8,1);
	\draw[-,line width=1.7] (8,1)--(9,1);
	\draw[-, line width=1.7] (2,1.3)-- (2,0.7); 
		\draw[-, line width=1] (2.1,1.1)-- (2.1,0.9);
		\draw[-, line width=1] (2.2,1.1)-- (2.2,0.9);
		\draw[-, line width=1] (2.3,1.1)-- (2.3,0.9);
		\draw[-, line width=1] (2.4,1.1)-- (2.4,0.9);
		\draw[-, line width=1] (2.5,1.1)-- (2.5,0.9);
	\draw[-, line width=1.7] (4,0.8)-- (4,1.2); 
		\foreach \x in {3.5,3.6,3.7,3.8,3.9,4,4.1,4.2,4.3,4.4,4.5} 
			\draw[shift={(\x,0)},color=black, line width=1] (0,1.1) -- (0,0.9);
		\foreach \x in {5,5.1,5.2,5.3,5.4,5.5,5.6,5.7,5.8,5.9,6}
			\draw[shift={(\x,0)},color=black, line width=1] (0,1.1) -- (0,0.9);
	\draw[-, line width=1.7] (5.5,0.8)-- (5.5,1.2); 
	\draw[-, line width=1.7, color=red, text=black] (7,1.2)-- (7,0.8); 
		\foreach \x in {7,7.1,7.2,7.3,7.4,7.5,7.6,7.7,7.8,7.9,8}
			\draw[shift={(\x,0)},color=red, line width=1] (0,1.1) -- (0,0.9);
	\draw[-, line width=1.7, color=red, text=black] (8,1.3)-- (8,0.7);
	
	\draw [decorate,decoration={brace,amplitude=10pt}]  (2,1.4) -- (8,1.4) node [midway, above, yshift=7.5] {Phase $i$};
	\draw [decorate,decoration={brace,mirror,amplitude=5pt}] (4,0.7) -- (5.5,0.7) node [midway, below, yshift=-5] {$T_j(i)\setminus T_j(i-1)$};
	\draw [decorate,decoration={brace,mirror,amplitude=5pt}, color=red]  (7,0.7) -- (8,0.7) node [midway, below, yshift=-5] {Bridge};
\end{tikzpicture}
  \caption{An example of phase $i$ of our algorithm.}
    \label{fig:phases2}
\end{figure}

A further nuance in the algorithm structure is the `{\it bridge period}' (see Figure~\ref{fig:phases2}). 
The algorithm picks an active arm $j\in \cA_{m+1}$ to play in this bridge period for $n_m-n_{m-1}$ steps. 
The observations received during the bridge period are discarded, and not used for computing confidence intervals. The significance of the bridge period is that it breaks the dependence between confidence intervals calculated in phase $m$ and the delayed payoffs seeping into phase $m+1$. Without the bridge period this dependence would impair the validity of our confidence intervals. However, we suspect that, in practice, it may be possible to remove it. 

\paragraph{Choice of $\bf n_m$} A key element of our algorithm design is the careful choice of $n_m$.
Since $n_m$ determines the number of times each active (possibly suboptimal) arm is played, it clearly has an impact on the regret.
Furthermore, $n_m$ needs to be chosen so that the confidence bounds on the estimation error hold with given probability. 
The main challenge is developing these confidence bounds from delayed, aggregated anonymous feedback. 
Handling this form of feedback involves a credit assignment problem of deciding which samples can be used for a given arm's mean estimation, since each sample is an aggregate of rewards from multiple previously played arms. This credit assignment problem would be hopeless in a passive learning setting without further information on how the samples were generated. Our algorithm utilizes the power of active learning to design the phases in such a way that the feedback can be effectively `decensored' without losing too many samples. 

A naive approach to defining the confidence bounds for delays bounded by a constant $d\geq 0$ would be to observe that,  
\[ \bigg| \sum_{t \in T_j(m) \setminus T_j(m-1)} X_t - \sum_{t \in T_j(m) \setminus T_j(m-1)} R_{t,j} \bigg| \leq d, \]
since all rewards are in $[0,1]$. Then we could use Hoeffding's inequality to bound $R_{t,J_t}$ (see Appendix~\ref{app:basicd}) and select
\[ n_m =  \frac{C_1\log(T\tilde \Delta_m^2)}{\tilde{\Delta}_m^2} + \frac{C_2 md}{\tilde \Delta_m} \]
 for some constants $C_1,C_2$. This corresponds to worst case regret of $O(\sqrt{KT\log K} + K \log(T) d)$. For $d \gg \E[\tau]$ and large $T$, this is significantly worse than that of \citet{joulani2013online}. 
In \cref{sec:reg}, we show that, surprisingly, it is possible to recover the same rate of regret as \citet{joulani2013online}, but this requires a significantly more nuanced argument to get tighter confidence bounds and smaller $n_m$.
In the next section, we describe this improved
choice of $n_m$ for every phase $m\in \mathbb{N}$ and its implications on the regret, for each of the three cases mentioned previously: (i) Known and bounded expected delay (Assumption \ref{assum:1}), (ii) Bounded delay with known bound and expected value (Assumptions \ref{assum:1} and \ref{assum:2}), (iii) Delay with known and bounded variance and expectation (Assumptions \ref{assum:1} and \ref{assum:3}). 

\section{Regret Analysis} \label{sec:reg}
In this section, we specify the choice of parameters $n_m$ and provide regret guarantees for Algorithm \ref{alg:skeleton} for each of the three previously mentioned cases.
\subsection{Known and Bounded Expected Delay}
First, we consider the setting with the weakest assumption on delay distribution: we only assume that the expected delay, $\Ex[\tau]$, is bounded and known. No assumption on the support or variance of the delay distribution is made. The regret analysis for this setting will not use the bridge period, so Step 4 of the algorithm could be omitted in this case. 

\paragraph{Choice of $\bf n_m$}
Here, we use Algorithm \ref{alg:skeleton} with 
\begin{equation}
\label{eq:nm}
n_m = \frac{C_1\log(T\tilde \Delta_m^2)}{\tilde{\Delta}_m^2} + \frac{C_2 m \mathbb{E}[\tau]}{\tilde \Delta_m}
\end{equation}
for some large enough constants $C_1, C_2$. The exact value of  $n_m$ is given in Equation~\eqref{eqn:ngen} in Appendix~\ref{app:gen}.
\paragraph{Estimation of error bounds}
We bound the error between $\bar{X}_{m,j}$ and $\mu_j$ by $\tilde{\Delta}_m/2$. In order to do this we first bound the corruption of the observations received during timesteps $\timeset_j(m)$ due to delays. 

Fix a phase $m$ and arm $j \in \cA_m$. Then the observations $X_t$ in the period ${t\in \timeset_j(m)\setminus \timeset_j(m-1)}$ are composed of two types of rewards: a subset of rewards from plays of arm $j$ in this period, and delayed rewards from some of the plays before this period. The expected value of observations from this period would be $(n_m-n_{m-1}) \mu_j$ but for the rewards entering and leaving this period due to delay. Since the reward is bounded by $1$, a simple observation is that expected discrepancy between the sum of observations in this period and the quantity $(n_m-n_{m-1}) \mu_j$ is bounded by the expected delay $\mathbb{E}[\tau]$,
\begin{equation}
\label{eq:intermediate1}
\mathbb{E}\left[\sum_{t\in \timeset_j(m)\setminus \timeset_j(m-1)} (X_t - \mu_j)\right] \le \Ex[\tau].
\end{equation}
Summing this over phases $\ell=1,\ldots m$ gives a bound
\begin{equation}
\label{eq:corruptionBound1}
|\mathbb{E}[\bar{X}_{m,j}] - \mu_j| \le \frac{m \mathbb{E}[\tau]}{|\timeset_j(m)|} =  \frac{m\mathbb{E}[\tau]}{n_m}.
\end{equation}
Note that given the choice of $n_m$ in \eqref{eq:nm}, the above is smaller than $\tilde \Delta_m/2$, when large enough constants are used. Using this, along with concentration inequalities and the choice of $n_m$ from \eqref{eq:nm}, 
we can obtain the following high probability bound. A detailed proof is provided in Appendix \ref{app:boundsgen}. 
\begin{restatable}{lemma}{lemebone}
\label{lem:eb1}
Under Assumption \ref{assum:1} and the choice of $n_m$ given by \eqref{eq:nm}, the estimates $\bar{X}_{m,j}$ constructed by Algorithm \ref{alg:skeleton} satisfy the following: For every fixed arm $j$ and phase $m$, with probability $1-\frac{3}{T\tilde{\Delta}_m^2}$, either $j\notin \cA_m, $ or:
$$\bar{X}_{m,j} -\mu_j \le \tilde{\Delta}_{m}/2\,.$$
\end{restatable}

\paragraph{Regret bounds} Using Lemma \ref{lem:eb1}, we derive the following regret bounds in the current setting.
\begin{restatable}{theorem}{thmreggenNew}
\label{thm:reggenNew}
Under Assumption \ref{assum:1}, the expected regret of Algorithm \ref{alg:skeleton} is upper bounded as
\begin{align} 
\E[\Reg_T] \leq  \sum_{\substack{j=1 \\ j \neq j^*}}^K O\bigg( \frac{\log(T\Delta_j^2)}{\Delta_j}   + \log(1/ \Delta_j)\E[\tau]\bigg). \label{eqn:reggenNew}
\end{align}
\end{restatable}
%
\begin{proof}
Given Lemma \ref{lem:eb1}, the proof of Theorem \ref{thm:reggenNew} closely follows the analysis of the Improved UCB algorithm of \citet{auer2010ucb}. Lemma \ref{lem:eb1} and the elimination condition in Algorithm \ref{alg:skeleton} ensure that, with high probability, any suboptimal arm $j$ will be eliminated by phase $m_j=\log(1/\Delta_j)$, thus incurring regret at most $n_{m_j}\Delta_j$  
We then substitute in $n_{m_j}$ from \eqref{eq:nm}, and sum over all suboptimal arms.
A detailed proof is in Appendix~\ref{app:reggen}. As in \citet{auer2010ucb}, we avoid a union bound over all arms (which would result in an extra $\log K$) by
{\em (i)} reasoning about the regret of each arm individually, and
{\em (ii)} bounding the regret resulting from erroneously eliminating the optimal arm by carefully controlling the probability it is eliminated in each phase.
\end{proof}

Considering the worst-case values of $\Delta_j$ (roughly $\sqrt{K/T}$), we obtain the following problem independent bound.
\begin{restatable}{corollary}{correggen}
\label{cor:reggen}
For any problem instance satisfying Assumption \ref{assum:1}, the expected regret of Algorithm \ref{alg:skeleton} satisfies
\[  \E[\Reg_T] \leq O(\sqrt{KT\log(K)} + K\E[\tau]\log(T)). \]
\end{restatable}

\subsection{Delay with Bounded Support}
If the delay is bounded by some constant $d \geq 0$ and a single arm is played repeatedly for long enough, we can restrict the number of arms corrupting the observation $X_t$ at a given time $t$. In fact, if each arm $j$ is played consecutively for more than $d$ rounds, then at any time $t \in T_j(m)$, the observation $X_t$ will be composed of the rewards from at most two arms: the current arm $j$, and previous arm $j'$. Further, from the elimination condition, with high probability, arm $j'$ will have been eliminated if it is clearly suboptimal. We can then recursively use the confidence bounds for arms $j$ and $j'$ from the previous phase to bound $|\mu_j - \mu_{j'}|$. 
Below, we formalize this intuition to obtain a tighter bound on $|\bar{X}_{m,j} - \mu_j|$ for every arm $j$ and phase $m$, when each active arm is played a specified number of times per phase.

\paragraph{Choice of $\bf n_m$}
Here, we define,
\begin{align}
\label{eq:nmBounded}
n_m =& \frac{C_1\log(T\tilde \Delta_m^2)}{\tilde{\Delta}_m^2} + \frac{C_2 \mathbb{E}[\tau]}{\tilde \Delta_m} 
\\ &+\min\bigg\{md, \frac{C_3\log(T\tilde \Delta_m^2)}{\tilde{\Delta}_m^2} + \frac{C_4 m\Ex[\tau]}{\tilde{\Delta}_m} \bigg \} \nonumber
\end{align}
for some large enough constants $C_1, C_2, C_3, C_4$ (see Appendix~\ref{app:bounded}, Equation~\eqref{eqn:nbounded} for the exact values).
This choice of $n_m$ means that for large $d$, we essentially revert back to the  choice of $n_m$ from \eqref{eq:nm} for the unbounded case, and we gain nothing by using the bound on the delay. However, if $d$ is not large, 
the choice of $n_m$ in \eqref{eq:nmBounded} is smaller than \eqref{eq:nm} since the second term now scales with $\E[\tau]$ rather than $m \E[\tau]$. 

\paragraph{Estimation of error bounds}
In this setting, by the elimination condition and bounded delays, the expectation of each reward entering $T_j(m)$ will be within $\tilde \Delta_{m-1}$ of $\mu_j$, with high probability.
Then, using knowledge of the upper bound of the support of $\tau$, we can obtain a tighter bound and get an error bound similar to Lemma \ref{lem:eb1} with the smaller value of $n_m$ in \eqref{eq:nmBounded}. 
We prove the following proposition. Since $\tilde{\Delta}_m = 2^{-m}$, this is considerably tighter than \eqref{eq:intermediate1}.
\begin{proposition} \label{prop:intermediate2}
Assume $n_i-n_{i-1}\ge d$ for phases $i= 1, \dots, m$. 
Define $\event_{m-1}$ as the event that all arms $j\in {\cal A}_{m}$ satisfy error bounds $|\bar{X}_{m-1,j} -\mu_j| \le \tilde{\Delta}_{m-1}/2$. Then, for every arm $j\in \cA_m$,
\begin{equation}
\label{eqintermediate2}
\Ex\left[\sum_{t\in \timeset_j(m) \setminus \timeset_j(m-1)} (X_t - \mu_j) \bigg| \event_{m-1} \right] \le \tilde{\Delta}_{m-1} \Ex[\tau]. \nonumber
\end{equation}
\end{proposition}
\begin{proof} (Sketch).
Consider a fixed arm $j \in \cA_m$. The expected value of the sum of observations $X_t$ for $t\in \timeset_j(m)\setminus \timeset_j(m-1)$ would be $(n_m-n_{m-1}) \mu_j$ were it not for some rewards entering and leaving this period due to the delays.
Because of the i.i.d. assumption on the delay, in expectation, the number of rewards leaving the period is roughly the same as the number of rewards entering this period, i.e., $\Ex[\tau]$. (Conditioning on $\event_{m-1}$ does not effect this due to the bridge period). Since $n_m-n_{m-1} \ge d$, the reward coming into the period $\timeset_j(m) \setminus \timeset_j(m-1)$ can only be from the previous arm $j'$. 
All rewards leaving the period are from arm $j$. Therefore the expected difference between rewards entering and leaving the period is $(\mu_j-\mu_{j'}) \Ex[\tau]$. Then, if $\mu_j$ is close to $\mu_{j'}$, the total reward leaving the period is compensated by total reward entering. 
Due to the bridge period, even when $j$ is the first arm played in phase $m$, $j'\in \cA_m$, so it was not eliminated in phase $m-1$. By the elimination condition in Algorithm \ref{alg:skeleton}, 
 if the error bounds $|\bar{X}_{m-1,j} -\mu_j| \le \tilde{\Delta}_{m-1}/2$ are satisfied for all arms in $\cA_m$, then $|\mu_j-\mu_{j'}|\le \tilde{\Delta}_{m-1}$. This gives the result.
\end{proof}

Repeatedly using Proposition~\ref{prop:intermediate2} we get, 
\[ \sum_{i=1}^m \Ex\left[\sum_{t\in \timeset_j(i)\setminus \timeset_j(i-1)} (X_t - \mu_j) \bigg| \event_{i-1} \right] \leq 2\Ex[\tau] \]
since $\sum_{i=1}^m \tilde \Delta_{i-1} = \sum_{i=0}^{m-1} 2^{-i} \leq 2$. Then, observe that $\PP(\event_i^C)$ is small.
This bound is an improvement of a factor of $m$ compared to \eqref{eq:corruptionBound1}. 
For the regret analysis, we derive a high probability version of the above result. 
Using this, and the choice of $n_m \ge \Omega\left(\frac{\log(T \tilde \Delta_m^2)}{\tilde \Delta_m^2} + \frac{\Ex[\tau]}{\tilde \Delta_m} \right)$ from \eqref{eq:nmBounded}, for large enough constants, we derive the following lemma. 
A detailed proof is given in Appendix \ref{app:cbbounded}. 
\begin{restatable}{lemma}{lemebtwo} 
\label{lem:eb2}
Under Assumptions \ref{assum:1} of known expected delay and \ref{assum:2} of bounded delays, and choice of $n_m$ given in \eqref{eq:nmBounded}, the estimates $\bar{X}_{m,j}$ obtained by Algorithm \ref{alg:skeleton} satisfy the following: For any arm $j$ and phase $m$, with probability at least $1-\frac{12}{T\tilde{\Delta}_m^2}$, either $j \notin \cA_m$ or
$$\bar{X}_{m,j} -\mu_j \le \tilde{\Delta}_{m}/2.$$
\end{restatable}
\paragraph{Regret bounds} We now give regret bounds for this case.

\begin{restatable}{theorem}{thmregboundedNew}
\label{thm:regboundedNew}
Under Assumption \ref{assum:1} and bounded delay Assumption \ref{assum:2}, the expected regret of Algorithm \ref{alg:skeleton} satisfies 
\begin{align*}
 \E[\Reg_T] \leq & \sum_{j=1; j \neq j^*}^K O\bigg( \frac{\log(T\Delta_j^2)}{\Delta_j}  +\E[\tau] 
 \\ &+ \min\bigg \{d,\frac{\log(T\Delta_j^2)}{\Delta_j} + \log(\frac{1}{\Delta_j})\Ex[\tau] \bigg\} \bigg). 
 \end{align*}
\end{restatable}
\begin{proof} (Sketch).
Given \cref{lem:eb2}, the proof is similar to that of \cref{thm:reggenNew}. 
 The full proof is in \cref{app:regbounded}.
\end{proof}

Then, if $d\leq \sqrt{\frac{T\log K}{K}} + \E[\tau]$, 
we get the following problem independent regret bound which matches that of \citet{joulani2013online}.
\begin{restatable}{corollary}{corregboundedNew}
\label{cor:regbounded}
For any problem instance satisfying Assumptions \ref{assum:1} and \ref{assum:2} with $d\leq \sqrt{\frac{T\log K}{K}} + \E[\tau]$, the expected regret of Algorithm \ref{alg:skeleton} satisfies 
\[ \E[\Reg_T] \leq O(\sqrt{KT\log(K)} + K\E[\tau]). \]
\end{restatable}

\subsection{Delay with Bounded Variance}
If the delay is unbounded but well behaved in the sense that we know (a bound on) the variance, then we can obtain similar regret bounds to the bounded delay case. Intuitively, delays from the previous phase will only corrupt observations in the current phase if their delays exceed the length of the bridge period. We control this by using the bound on the variance to bound the tails of the delay distributions.

\paragraph{Choice of $\bf n_m$}
Let $\Var(\tau)$ be the known variance (or bound on the variance) of the delay, as in Assumption \ref{assum:3}. Then, we use Algorithm \ref{alg:skeleton} with the following value of $n_m$,
\begin{equation}
\label{eq:nvar}
n_m = C_1 \frac{\log(T\tilde{\Delta}_m^2)}{\tilde \Delta_m^2} + C_2 \frac{\Ex[\tau] + \Var(\tau)}{\tilde \Delta_m}
\end{equation}
for some large enough constants $C_1, C_2$. The exact value of  $n_m$ is given in Appendix~\ref{app:var}, Equation~\eqref{eqn:nvarlong}.

\paragraph{Regret bounds} We get the following instance specific and problem independent regret bound in this case.
\begin{restatable}{theorem}{thmregvar}
\label{thm:regvar}
Under Assumption \ref{assum:1} and Assumption \ref{assum:3} of known (bound on) the expectation and variance of the delay, and choice of $n_m$ from \eqref{eq:nvar}, the expected regret of Algorithm \ref{alg:skeleton} can be upper bounded by,
\[ \E[\Reg_T] \leq \sum_{j=1: \mu_j\ne \mu^*}^K O\bigg( \frac{\log(T\Delta_j^2)}{\Delta_j} + \E[\tau] + \Var(\tau) \bigg).\]
\end{restatable}
\begin{proof} (Sketch). See Appendix~\ref{app:regvar}.
We use Chebychev's inequality to get a result similar to Lemma~\ref{lem:eb2} and then use a similar argument to the bounded delay case.
\end{proof} 
\begin{restatable}{corollary}{corregvar}
\label{cor:regvar}
For any problem instance satisfying Assumptions \ref{assum:1} and \ref{assum:3}, the expected regret of Algorithm \ref{alg:skeleton} satisfies 
\[ \E[\Reg_T] \leq O(\sqrt{KT\log(K)} + K\E[\tau] + K \Var(\tau)). \]
\end{restatable}
\paragraph{Remark} 
If $\E[\tau] \geq 1$, then the delay penalty can be reduced to $O( K\E[\tau] + K \Var(\tau)/\E[\tau])$ (see Appendix~\ref{app:var}).

Thus, it is sufficient to know a bound on variance to obtain regret bounds similar to those in bounded delay case. Note that this approach is not possible just using knowledge of the expected delay since we cannot guarantee that the reward entering phase $i$ is from an arm active in phase $i-1$.



\section{Experimental Results} \label{sec:experiments}

\begin{figure}[t]
    \centering
    \begin{subfigure}{0.42\textwidth}
        \centering
        \includegraphics[width=0.74\textwidth]{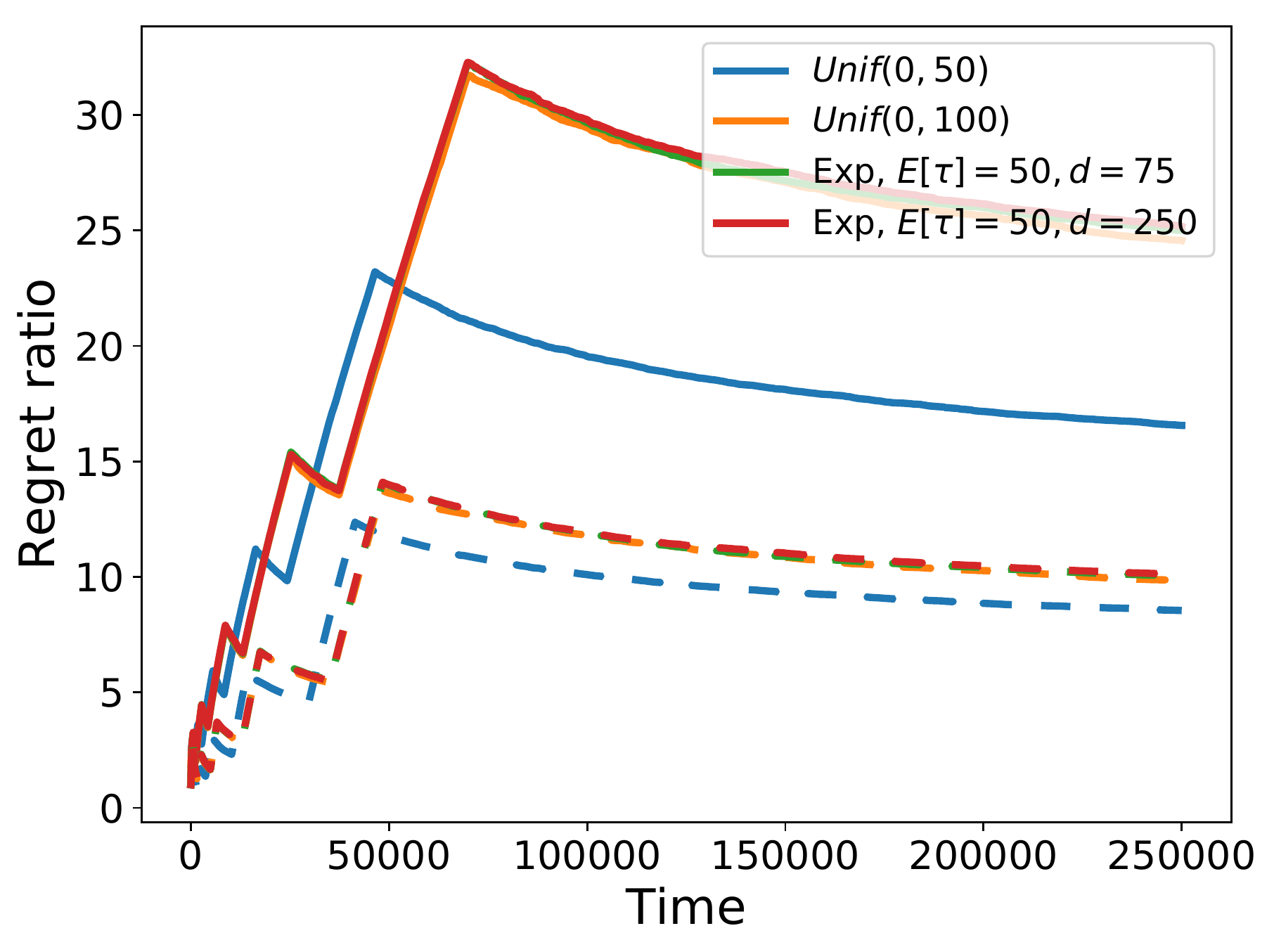}
        \caption{Bounded delays. Ratios of regret of \ODAAF (solid lines) and \ODAAF-B (dotted lines) to that of QPM-D.}
        \label{fig:bounded}
    \end{subfigure}
    \quad
    \begin{subfigure}{0.42\textwidth}
        \centering
        \includegraphics[width=0.74\textwidth]{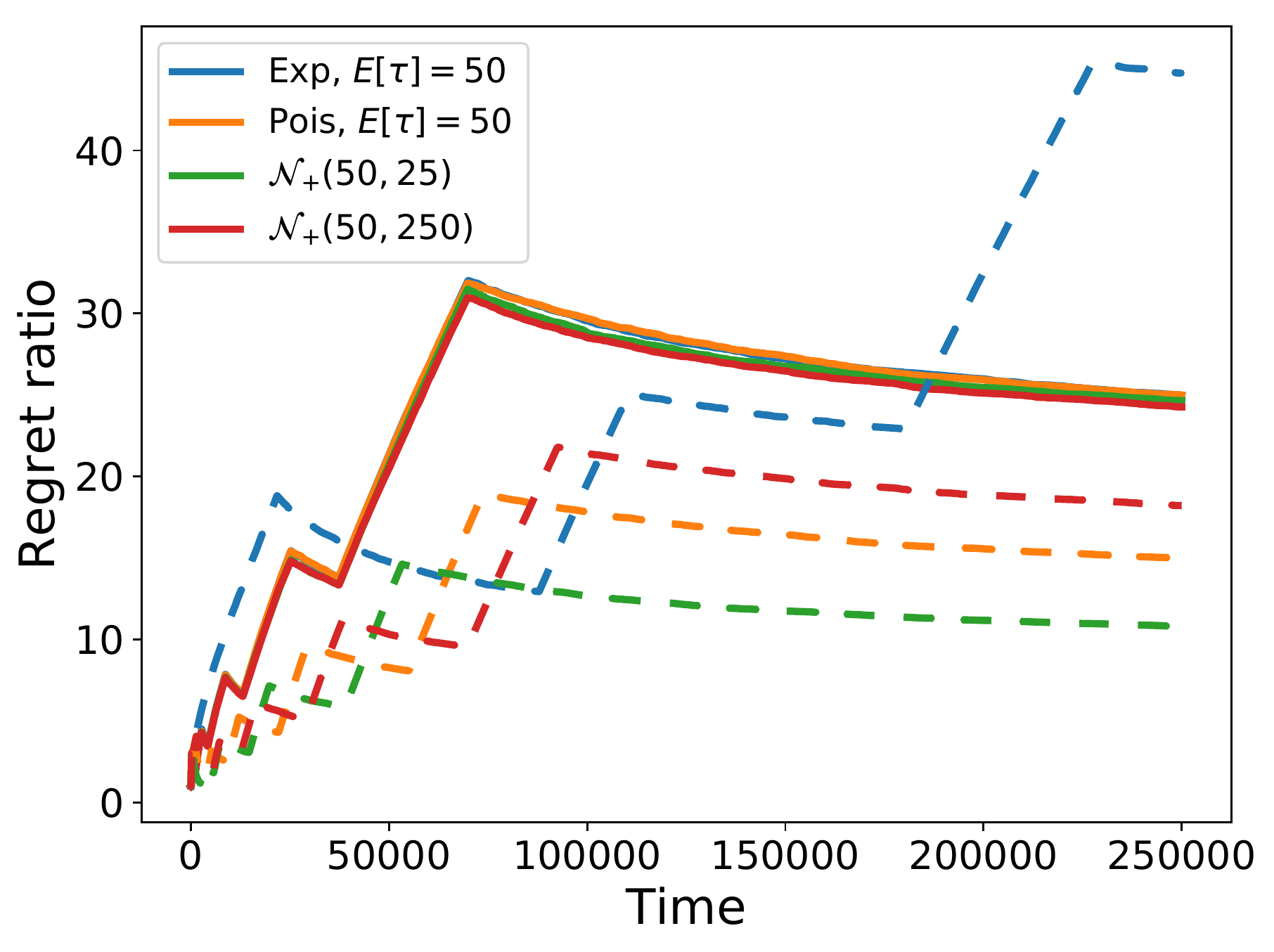}
        \caption{Unbounded delays. Ratios of regret of \ODAAF (solid lines) and \ODAAF-V (dotted lines) to that of QPM-D.}
        \label{fig:unbounded}
    \end{subfigure}
    \caption{The ratios of regret of variants of our algorithm to that of QPM-D for different delay distributions. } \label{fig:results}
\end{figure}

We compared the performance of our algorithm (under different assumptions) to QPM-D \citep{joulani2013online} in various experimental settings. In these experiments, our aim was to investigate the effect of the delay on the performance of the algorithms. In order to focus on this, we used a simple setup of two arms with Bernoulli rewards and ${\boldsymbol \mu}=(0.5, 0.6)$. In every experiment, we ran each algorithm to horizon $T=250000$ and used UCB1 \cite{auer2002finite} as the base algorithm in QPM-D. The regret was averaged over $200$ replications. For ease of reading, we define \ODAAF to be our algorithm using only knowledge of the expected delay, with $n_m$ defined as in  \labelcref{eq:nm} and run without a bridge period, and \ODAAF-B and \ODAAF-V to be the versions of Algorithm~\ref{alg:skeleton} that use a bridge period and information on the bounded support and the finite variance of the delay to define $n_m$ as in \labelcref{eq:nmBounded} and \labelcref{eq:nvar} respectively.

We tested the algorithms with different delay distributions. In the first case, we considered bounded delay distributions whereas in the second case, the delays were unbounded. 
In \cref{fig:bounded}, we plotted the ratios of the regret of \ODAAF and \ODAAF-B (with knowledge of $d$, the delay bound)  to the regret of QPM-D. We see that in all cases the ratios converge to a constant. This shows that the regret of our algorithm is essentially of the same order as that of QPM-D. Our algorithm predetermines the number of times to play each active arm per phase (the randomness appears in whether an arm is active), so the jumps in the regret are it changing arm. This occurs at the same points in all replications. 

\cref{fig:unbounded} shows a similar story for unbounded delays with mean $\E[\tau]=50$ (where $\mathcal{N}_+$ denotes the the half normal distribution). The ratios of the regret of \ODAAF and \ODAAF-V (with knowledge of the delay variance) to the regret of QPM-D again converge to constants. Note that in this case, these constants, and the location of the jumps, vary with the delay distribution and $\Var(\tau)$. When the variance of the delay is small, it can be seen that using the variance information leads to improved performance. However, for exponential delays where $\Var(\tau) = \E[\tau]^2$, the large variance causes $n_m$ to be large and so the suboptimal arm is played more, increasing the regret. In this case \ODAAF-V had only just eliminated the suboptimal arm at time $T$. 

It can also be illustrated experimentally that the regret of our algorithms and that of QPM-D all increase linearly in $\E[\tau]$. This is shown in Appendix~\ref{app:experiments}. We also provide an experimental comparison to \citet{vernade2017stochastic} in Appendix~\ref{app:experiments}.

\section{Conclusion}
\label{sec:conc}
We have studied an extension of the multi-armed bandit problem to bandits with delayed, aggregated anonymous feedback. Here, a sum of observations is received after some stochastic delay and we do not learn which arms contributed to each observation. In this more difficult setting, we have proven that, surprisingly, it is possible to develop an algorithm that performs comparably to those for the simpler delayed feedback bandits problem, where the assignment of rewards to plays is known. Particularly, using only knowledge of the expected delay, our algorithm matches the worst case regret of \citet{joulani2013online} up to a logarithmic factor. 
This logarithmic factors can be removed using an improved analysis and slightly more information about the delay; if the delay is bounded, we achieve the same worst case regret as \citet{joulani2013online}, and for unbounded delays with known finite variance, we have an extra additive $\Var(\tau)$ term. 
We supported these claims experimentally.
Note that while our algorithm matches the order of regret of QPM-D, the constants are worse. Hence, it is an open problem to find algorithms with better constants. 

\newpage

\section*{Acknowledgments}
CPB would like to thank the EPSRC funded EP/L015692/1 STOR-i centre for doctoral training and Sparx.
We would like to thank the reviewers for their helpful comments.

\bibliography{df_refs}

\begin{thebibliography}{19}
\providecommand{\natexlab}[1]{#1}
\providecommand{\url}[1]{\texttt{#1}}
\expandafter\ifx\csname urlstyle\endcsname\relax
  \providecommand{\doi}[1]{doi: #1}\else
  \providecommand{\doi}{doi: \begingroup \urlstyle{rm}\Url}\fi

\bibitem[Agrawal et~al.(1988)Agrawal, Hedge, and
  Teneketzis]{agrawal1988asymptotically}
Agrawal, R., Hedge, M., and Teneketzis, D.
\newblock Asymptotically efficient adaptive allocation rules for the multiarmed
  bandit problem with switching cost.
\newblock \emph{IEEE Transactions on Automatic Control}, 33\penalty0
  (10):\penalty0 899--906, 1988.

\bibitem[Auer \& Ortner(2010)Auer and Ortner]{auer2010ucb}
Auer, P. and Ortner, R.
\newblock {UCB} revisited: Improved regret bounds for the stochastic
  multi-armed bandit problem.
\newblock \emph{Periodica Mathematica Hungarica}, 61\penalty0 (1-2):\penalty0
  55--65, 2010.

\bibitem[Auer et~al.(2002)Auer, Cesa-Bianchi, and Fischer]{auer2002finite}
Auer, P., Cesa-Bianchi, N., and Fischer, P.
\newblock Finite-time analysis of the multiarmed bandit problem.
\newblock \emph{Journal of Machine Learning Research}, 47\penalty0
  (2-3):\penalty0 235--256, 2002.

\bibitem[Bubeck \& Cesa-Bianchi(2012)Bubeck and Cesa-Bianchi]{BC12}
Bubeck, S. and Cesa-Bianchi, N.
\newblock \emph{Regret Analysis of Stochastic and Nonstochastic Multi-armed
  Bandit Problems}.
\newblock Foundations and Trends in Machine Learning. 2012.

\bibitem[Capp{\'e} et~al.(2013)Capp{\'e}, Garivier, Maillard, Munos, and
  Stoltz]{cappe2013kullback}
Capp{\'e}, O., Garivier, A., Maillard, O.-A., Munos, R., and Stoltz, G.
\newblock Kullback--{L}eibler upper confidence bounds for optimal sequential
  allocation.
\newblock \emph{The Annals of Statistics}, 41\penalty0 (3):\penalty0
  1516--1541, 2013.

\bibitem[Cesa-Bianchi et~al.(2016)Cesa-Bianchi, Gentile, Mansour, and
  Minora]{cesa2016delay}
Cesa-Bianchi, N., Gentile, C., Mansour, Y., and Minora, A.
\newblock Delay and cooperation in nonstochastic bandits.
\newblock In \emph{Conference on Learning Theory}, pp.\  605--622, 2016.

\bibitem[Chapelle \& Li(2011)Chapelle and Li]{chapelle2011empirical}
Chapelle, O. and Li, L.
\newblock An empirical evaluation of {T}hompson sampling.
\newblock In \emph{Advances in Neural Information Processing Systems}, pp.\
  2249--2257, 2011.

\bibitem[Desautels et~al.(2014)Desautels, Krause, and
  Burdick]{desautels2014parallelizing}
Desautels, T., Krause, A., and Burdick, J.
\newblock Parallelizing exploration-exploitation tradeoffs in {G}aussian
  process bandit optimization.
\newblock \emph{Journal of Machine Learning Research}, 15\penalty0
  (1):\penalty0 3873--3923, 2014.

\bibitem[Doob(1953)]{Doob53}
Doob, J.~L.
\newblock \emph{Stochastic processes}.
\newblock John Wiley \& Sons, 1953.

\bibitem[Dudik et~al.(2011)Dudik, Hsu, Kale, Karampatziakis, Langford, Reyzin,
  and Zhang]{dudik2011efficient}
Dudik, M., Hsu, D., Kale, S., Karampatziakis, N., Langford, J., Reyzin, L., and
  Zhang, T.
\newblock Efficient optimal learning for contextual bandits.
\newblock In \emph{Conference on Uncertainty in Artificial Intelligence}, pp.\
  169--178, 2011.

\bibitem[Freedman(1975)]{Fre75}
Freedman, D.~A.
\newblock On tail probabilities for martingales.
\newblock \emph{The Annals of Probability}, 3\penalty0 (1):\penalty0 100--118,
  1975.

\bibitem[Joulani et~al.(2013)Joulani, Gy{\"o}rgy, and
  Szepesv{\'a}ri]{joulani2013online}
Joulani, P., Gy{\"o}rgy, A., and Szepesv{\'a}ri, C.
\newblock Online learning under delayed feedback.
\newblock In \emph{International Conference on Machine Learning}, pp.\
  1453--1461, 2013.

\bibitem[Lai \& Robbins(1985)Lai and Robbins]{lai1985asymptotically}
Lai, T. and Robbins, H.
\newblock Asymptotically efficient adaptive allocation rules.
\newblock \emph{Advances in Applied Mathematics}, 6\penalty0 (1):\penalty0
  4--22, 1985.

\bibitem[Mandel et~al.(2015)Mandel, Liu, Brunskill, and
  Popovic]{mandel2015queue}
Mandel, T., Liu, Y.-E., Brunskill, E., and Popovic, Z.
\newblock The queue method: Handling delay, heuristics, prior data, and
  evaluation in bandits.
\newblock In \emph{AAAI}, pp.\  2849--2856, 2015.

\bibitem[Neu et~al.(2010)Neu, Antos, Gy{\"o}rgy, and
  Szepesv{\'a}ri]{neu2010online}
Neu, G., Antos, A., Gy{\"o}rgy, A., and Szepesv{\'a}ri, C.
\newblock Online {M}arkov decision processes under bandit feedback.
\newblock In \emph{Advances in Neural Information Processing Systems}, pp.\
  1804--1812, 2010.

\bibitem[Perchet et~al.(2016)Perchet, Rigollet, Chassang, and
  Snowberg]{perchet2016batched}
Perchet, V., Rigollet, P., Chassang, S., and Snowberg, E.
\newblock Batched bandit problems.
\newblock \emph{The Annals of Statistics}, 44\penalty0 (2):\penalty0 660--681,
  2016.

\bibitem[Szita \& Szepesv{\'a}ri(2011)Szita and Szepesv{\'a}ri]{SziSze11}
Szita, I. and Szepesv{\'a}ri, C.
\newblock Agnostic {KWIK} learning and efficient approximate reinforcement
  learning.
\newblock In \emph{Conference on Learning Theory}, pp.\  739--772, July 2011.

\bibitem[Thompson(1933)]{thompson1933likelihood}
Thompson, W.
\newblock On the likelihood that one unknown probability exceeds another in
  view of the evidence of two samples.
\newblock \emph{Biometrika}, 25\penalty0 (3--4):\penalty0 285--294, 1933.

\bibitem[Vernade et~al.(2017)Vernade, Capp{\'e}, and
  Perchet]{vernade2017stochastic}
Vernade, C., Capp{\'e}, O., and Perchet, V.
\newblock Stochastic bandit models for delayed conversions.
\newblock In \emph{Conference on Uncertainty in Artificial Intelligence}, 2017.

\end{thebibliography}


\newpage
\onecolumn
\appendix
\section*{Appendix}

\section{Preliminaries}
\subsection{Table of Notation} \label{app:notation}
For ease of reading, we define here key notation that will be used in this Appendix.
\vspace{-5pt}
\begin{center}
\begin{tabular}{r c p{12cm}}
	$T$ &: & The horizon. \\
	$\Delta_j$ & : & The gap between the mean of the optimal arm and the mean of arm $j$, $\Delta_j = \mu^*-\mu_j$. \\
	$\tilde \Delta_m$ & : & The approximation to $\Delta_j$ at round $m$ of the \ODAAF algorithm, $\tilde \Delta_m = \frac{1}{2^{m}}$. \\
	$n_m$ & : & The number of samples of an active arm $j$ \ODAAF needs by the end of round $m$. \\
	$\nu_m$ & : & The number of times each arm is played in phase $m$, $\nu_m = n_m- n_{m-1}$. \\
	$d$ & : & The bound on the delay in the case of bounded delay. \\
	$m_j$ & : & The first round of the \ODAAF algorithm where $\tilde \Delta_m < \nicefrac{\Delta_j}{2}$. \\
	$M_j$ & : & The random variable representing the round arm $j$ is eliminated in. \\
	$ T_j(m)$ & : & The set of all time point where arm $j$ is played up to (and including) round $m$. \\
	$X_t$ & : & The reward received at time $t$ (from any possible past plays of the algorithm). \\
	$ R_{t,j}$ & : & The reward generated by playing arm $j$ at time $t$. \\
	$ \tau_{t,j}$ &:& The delay associated with playing arm $j$ at time $t$. \\
	$ \E[\tau]$ &:& The expected delay (assuming i.i.d. delays). \\
	$\Var(\tau)$ &:& The variance of the delay (assuming i.i.d. delays). \\
	$ \bar X_{m,j}$ &:& The estimated reward of arm $j$ in phase $m$. See Algorithm~\ref{alg:skeleton} for the definition. \\
	$S_m$ &:& The start point of the $m$th phase. See \cref{app:phases} for more details. \\
	$U_m$ &:& The end point of the $m$th phase. See \cref{app:phases} for more details. \\
	$S_{m,j}$ & : & The start point of phase $m$ of playing arm $j$. See \cref{app:phases} for more details.  \\
	$U_{m,j}$ & : & The end point of phase $m$ of playing arm $j$. See \cref{app:phases} for more details.  \\
	$\cA_m$ & : & The set of active arms in round $m$ of the \ODAAF algorithm. \\
	$A_{i,t}, B_{i,t}, C_{i,t}$ &:& The contribution of the reward generated at time $t$ in certain intervals relating to phase $i$ to the corruption. See \eqref{eqn:abcdef} for the exact definitions. \\
	$\mathcal{G}_t$ &:& The smallest $\sigma$-algebra containing all information up to time $t$, see \eqref{eqn:gt} for a definition. \\
\end{tabular}
\end{center}

\subsection{Beginning and End of Phases} \label{app:phases}
\begin{figure}[h]
 \centering
  \begin{tikzpicture}
	\draw[-, line width=1.7] (1,1)--(2.5,1);
	\draw[dotted, line width=1.7] (2.5,1)--(3.5,1);
	\draw[-, line width=1.7] (3.5,1)--(4.5,1);
	\draw[dotted, line width=1.7] (4.5,1)--(5,1);
	\draw[-, line width=1.7] (5,1)--(6,1);
	\draw[dotted, line width=1.7] (6,1)--(7,1);
	\draw[-,line width=1.7,color=red] (7,1)--(8,1);
	\draw[-,line width=1.7] (8,1)--(9,1);
	\draw[-, line width=1.7] (2,1.3)-- (2,0.7) node[below] {$S_i$};
		\draw[-, line width=1] (2.1,1.1)-- (2.1,0.9);
		\draw[-, line width=1] (2.2,1.1)-- (2.2,0.9);
		\draw[-, line width=1] (2.3,1.1)-- (2.3,0.9);
		\draw[-, line width=1] (2.4,1.1)-- (2.4,0.9);
		\draw[-, line width=1] (2.5,1.1)-- (2.5,0.9);
	\draw[-, line width=1.7] (4,0.8)-- (4,1.2) node[above] {$S_{i,j}$};
		\foreach \x in {3.5,3.6,3.7,3.8,3.9,4,4.1,4.2,4.3,4.4,4.5} 
			\draw[shift={(\x,0)},color=black, line width=1] (0,1.1) -- (0,0.9);
		\foreach \x in {5,5.1,5.2,5.3,5.4,5.5,5.6,5.7,5.8,5.9,6}
			\draw[shift={(\x,0)},color=black, line width=1] (0,1.1) -- (0,0.9);
	\draw[-, line width=1.7] (5.5,0.8)-- (5.5,1.2)  node[above] {$U_{i,j}$};
	\draw[-, line width=1.7, color=red, text=black] (7,0.8)-- (7,1.2) node[above] {$U_{i,j'}+1$};
		\foreach \x in {7,7.1,7.2,7.3,7.4,7.5,7.6,7.7,7.8,7.9,8}
			\draw[shift={(\x,0)},color=red, line width=1] (0,1.1) -- (0,0.9);
	\draw[-, line width=1.7, color=red, text=black] (8,1.3)-- (8,0.7) node[below] {$U_i$};
	
	\draw [decorate,decoration={brace,amplitude=10pt}]  (2,1.7) -- (8,1.7) node [midway, above, yshift=7.5] {Phase $i$};
	\draw [decorate,decoration={brace,mirror,amplitude=5pt}] (4,0.7) -- (5.5,0.7) node [midway, below, yshift=-5] {$T_j(i)\setminus T_j(i-1)$};
	\draw [decorate,decoration={brace,mirror,amplitude=5pt}, color=red]  (7,0.7) -- (8,0.7) node [midway, below, yshift=-9] {Bridge};
\end{tikzpicture}
  \caption{An example of phase $i$ of our algorithm. Here $j'$ is the last active arm played in phase $i$.}
  \label{fig:phasesS}
\end{figure}
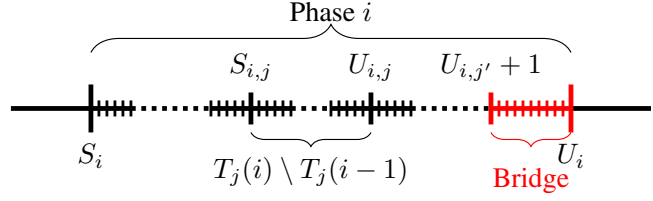

We formalize here some notation that will be used throughout the analysis to denote the start and end points of each phase.
 Define the random variables $S_i$ and $U_i$ for each phase $i=1,\dots, m$ to be the start and end points of the phase. Then let $S_{i,j}$, $U_{i,j}$ denote the start and end points of playing arm $j$ in phase $i$. See Figure~\ref{fig:phasesS} for details. By convention, let $S_{i,j}=U_{i,j}=\infty$ if arm $j$ is not active in phase $i$, $S_i=U_i =\infty$ if the algorithm never reaches phase $i$ and let $S_{0,j}=U_{0,j}=S_0=U_0=0$ for all $j$. It is important to point out that $n_m$ are deterministic so at the end of any phase $m-1$, once we have eliminated sub-optimal arms, we also know which arms are in $\cA_m$ and consequently the start and end points of phase $m$. Furthermore, since we play arms in a given order, we also know the specific rounds when we start and finish playing each active arm in phase $m$. Hence, at any time step $t$ in phase $m$, $S_m, U_m, S_{m+1}$ and $U_{m,j},S_{m,j}$ for all active arms $j \in \cA_m$ will be known. More formally, define the filtration $\{\mathcal{G}_t\}_{t=0}^\infty$ where 
 \begin{align}
 \mathcal{G}_t = \sigma(X_1, \dots, X_t, \tau_{1,J_1}, \dots, \tau_{t,J_t}, R_{1,J_1} ,\dots, R_{t,J_t}, J_1, \dots, J_t) \label{eqn:gt}
 \end{align}
 and $\mathcal{G}_0 = \{ \emptyset, \Omega\}$. 
 This means the joint events like $\{S_i \leq t\} \cap \{S_{i,j} = s' \} \in \cG_t$ for all $s'\in \mathbb{N}$, $j\in \cA$. 

\subsection{Useful Results}
For our analysis, we will need Freedman's version of Bernstein's inequality for the right-tail of martingales with bounded increments:
\begin{theorem}[Freedman's version of Bernstein's inequality; Theorem 1.6 of \citet{Fre75}]
\label{thm:freedman}
Let $\{Y_k\}_{k=0}^\infty$ be a real-valued martingale with respect to the filtration $\{\cF_k\}_{k=0}^\infty$ with increments $\{Z_k\}_{k=1}^\infty$:
$\E[Z_k|\cF_{k-1}]=0$ and $Z_k = Y_k - Y_{k-1}$, for $k=1,2,\dots$.
 Assume that the difference sequence is uniformly bounded on the right: $Z_k \le b$ almost surely for $k=1,2,\dots$. Define the predictable variation process $W_k = \sum_{j=1}^{k} \E[Z_j^2|\cF_{j-1}]$ for $k=1,2,\dots$. Then, for all $t\ge 0$, $\sigma^2>0$,
\begin{align*}
\Prob{ \exists k\ge 0\,: Y_k \ge t \text{ and } W_k\le \sigma^2 } \le \exp\left\{- \frac{t^2/2}{\sigma^2 + bt/3} \right\}\,.
\end{align*}
\end{theorem}
This result implies that if for some deterministic constant, $\sigma^2$, $W_k \le \sigma^2$ holds almost surely, then $\Prob{ Y_k \ge t } \le  \exp\left\{- \frac{t^2/2}{\sigma^2 + bt/3}\right\}$ holds for any $t\ge 0$. 

We will also make use of the following technical lemma which combines the Hoeffding-Azuma inequality and Doob's optional skipping theorem (Theorem~2.3 in Chapter VII of  \citet{Doob53})):
\begin{lemma}
\label{lem:concentration}
Fix the positive integers $m,n$ and let $a,c\in \mathbb{R}$.
Let $\mathcal{F}=\{\mathcal{F}_t\}_{t=0}^n$ be a filtration, 
  $(\epsilon_t,Z_t)_{t=1,2,\dots,n}$ be a sequence of $\{0,1\}\times \mathbb{R}$-valued 
  random variables
  such that for $t\in \{1,2,\dots,n\}$, 
 $\epsilon_t$ is $\mathcal{F}_{t-1}$-measurable, 
 $Z_t$ is $\mathcal{F}_t$-measurable, $\EE{Z_t|\mathcal{F}_{t-1}}=0$ and $Z_t\in [a,a+c]$.
Further, assume that $\sum_{s=1}^n \epsilon_s \le m$ with probability one.
Then, for any $\lambda >0$, 
\begin{align}
\PP \bigg( \sum_{t=1}^n \epsilon_t Z_t \geq \lambda \bigg) \leq \exp \bigg \{ - \frac{2\lambda^2}{c^2m} \bigg \}. 
\end{align}
\end{lemma}
\begin{proof}
This lemma appeared 
in a slightly more general form (where $n=\infty$ is allowed) 
as Lemma~A.1 in the paper by \citet{SziSze11} so we refer the reader to the proof there.
\end{proof}

\section{Results for Known and Bounded Expected Delay} \label{app:gen}
\subsection{High Probability Bounds} \label{app:boundsgen}
\lemebone*
\begin{proof}
Let 
\begin{align}
w_m = \frac{4\log(T\tilde \Delta_m^2)}{3n_m} + \sqrt{\frac{2\log(T \tilde \Delta_m^2)}{n_m}} + \frac{3m\E[\tau]}{n_m}. \label{eqn:wmgen}
\end{align}
We first show that with probability greater than $1-\frac{3}{T\tilde \Delta_m^2}$, $j \notin \cA_m$ or $\frac{1}{n_m}\sum_{t \in T_j(m)} (X_t - \mu_j) \leq  w_m$. 

For arm $j$ and phase $m$, assume $j \in \cA_m$. For notational simplicity we will use in the following 
$\one_i \{H \} := \one \{H \cap \{j \in \cA_i \}\} \leq \one\{H\}$ for any event $H$. If $j\in \mathcal{A}_m$ for a particular experiment $\omega$ then $\one_i(H)(\omega) = \one(H)(\omega)$. Then for any phase $i\leq m$ and time $t$, define,
\begin{align}
A_{i,t} = R_{t,J_t} \one\{ \tau_{t,J_t} + t \geq S_i\},
\quad B_{i,t} = R_{t,J_t} \one \{ \tau_{t,J_t} + t \geq S_{i,j} \}, 
\quad  C_{i,t} = R_{t,J_t} \one \{ \tau_{t,J_t} +t > U_{i,j} \}, \label{eqn:abcdef}
\end{align}
and note that since $S_{i,j}=U_{i,j} = \infty$ if arm $j$ is not active in phase $i$, we have the equalities $\one_i \{ \tau_{t,J_t} + t \geq S_{i,j} \} =  \one \{ \tau_{t,J_t} + t \geq S_{i,j} \}$ and $ \one_i \{ \tau_{t,J_t} +t > U_{i,j} \} = \one \{ \tau_{t,J_t} +t > U_{i,j} \}$.
Define the filtration $\{\cG_s\}_{s=0}^\infty$ by $
\cG_0=\{\Omega, \emptyset\}$ and
\begin{align}
\cG_t = \sigma(X_1, \dots, X_t, J_1, \dots, J_t, \tau_{1,J_1}, \dots, \tau_{t,J_t}, R_{1,J_1}, \dots R_{t,J_t}).\label{eqn:Gdef}
\end{align}

Then, we use the decomposition,
\begin{align}
\sum_{i=1}^m \sum_{t=S_{i,j}}^{U_{i,j}} (X_t - \mu_j) &\leq \sum_{i=1}^m \bigg( \sum_{t=S_{i-1,j}}^{S_{i,j}-1} R_{t,J_t} \one_i \{ \tau_{t,J_t} + t \geq S_{i,j}\} + \sum_{t=S_{i,j}}^{U_{i,j}} (R_{t,J_t} - \mu_j) - \sum_{t=S_{i,j}}^{U_{i,j}} R_{t,J_t} \one_i\{ \tau_{t,J_t} + t > U_{i,j}\} \bigg) \nonumber
\\ &\leq \sum_{i=1}^m \bigg( \sum_{t=S_{i-1,j}}^{S_i -1} R_{t,J_t} \one \{ \tau_{t,J_t} + t \geq S_i\} + \sum_{t=S_i }^{S_{i,j}-1} R_{t,J_t} \one \{ \tau_{t,J_t} + t \geq S_{i,j}\}  \nonumber
	\\ & \hspace{50pt} + \sum_{t=S_{i,j}}^{U_{i,j}} (R_{t,J_t} - \mu_j) - \sum_{t=S_{i,j}}^{U_{i,j}} R_{t,J_t} \one\{ \tau_{t,J_t} + t > U_{i,j}\} \bigg) \nonumber
\\ & = \sum_{i=1}^m \bigg( \sum_{t=S_{i-1,j}}^{S_i -1} A_{i,t} +  \sum_{t=S_i }^{S_{i,j}-1} B_{i,t} + \sum_{t=S_{i,j}}^{U_{i,j}} (R_{t,J_t} - \mu_j) - \sum_{t=S_{i,j}}^{U_{i,j}} C_{i,t} \bigg) \nonumber
\\ & = \sum_{i=1}^m \sum_{t=S_{i,j}}^{U_{i,j}}(R_{t,J_t} - \mu_j) + \sum_{t=1}^{S_{m,j}} Q_t - \sum_{t=1}^{U_{m,j}} P_t \nonumber
\\ & = \underbrace{\sum_{i=1}^m \sum_{t=S_{i,j}}^{U_{i,j}}(R_{t,J_t} - \mu_j)}_{\text{Term I.}} 
	+ \underbrace{ \sum_{t=1}^{S_{m,j}} (Q_t - \E[Q_t| \cG_{t-1}])}_{\text{Term II.}}
	 + \underbrace{ \sum_{t=1}^{U_{m,j}}( \E[P_t|\cG_{t-1}] - P_t)}_{\text{Term III.}} \label{eqn:decompgen}
	\\ & \hspace{50pt}  + \underbrace{\bigg( \sum_{t=1}^{S_{m,j}} \E[Q_t|\cG_{t-1}] - \sum_{t=1}^{U_{m,j}} \E[P_t|\cG_{t-1}]\bigg),}_{\text{Term IV.}} \nonumber
\end{align}
where,
\begin{align*}
Q_t &= \sum_{i=1}^m (A_{i,t} \one\{S_{i-1,j} \leq t \leq S_i -1\} + B_{i,t} \one \{S_i  \leq t \leq S_{i,j}-1\} )
\\ P_t &= \sum_{i=1}^m C_{i,t} \one\{S_{i,j} \leq t \leq U_{i,j}\}. 
\end{align*}
Recall that the filtration $\{\cG_s\}_{s=0}^\infty$ is defined by $\cG_0=\{\Omega, \emptyset\}$, $\cG_t = \sigma(X_1, \dots, X_t, J_1, \dots, J_t, \tau_{1,J_1}, \dots, \tau_{t,J_t}, R_{1,J_1}, \dots R_{t,J_t})$ and we have defined $S_{i,j}=\infty$ if arm $j$ is eliminated before phase $i$ and $S_i = \infty$ if the algorithm stops before reaching phase $i$. 

\paragraph{Outline of proof}
	We will bound each term of the above decomposition in \eqref{eqn:decompgen} in turn, however first we need to prove several intermediary results. For term II., we will use Freedman's inequality so we first need Lemma~\ref{lem:qmartgen} to show that $Z_t = Q_t - \E[Q_t|\cG_{t-1}]$ is a martingale difference and Lemma~\ref{lem:varqgen} to bound the variance of the sum of the $Z_t$'s. Similarly, for term III., in Lemma~\ref{lem:pmart}, we show that $Z_t' = \E[P_t|\cG_{t-1}]-P_t$ is a martingale difference and bound its variance in Lemma~\ref{lem:varp}. In Lemma~\ref{lem:abcboundsgen}, we consider term IV. and bound the conditional expectations of $A_{i,t},B_{i,t}, C_{i,t}$. Finally, in Lemma~\ref{lem:az}, we bound term I. using Lemma~\ref{lem:concentration}. We then combine the bounds on all terms together to conclude the proof.

\begin{lemma} \label{lem:qmartgen}
Let $Y_s = \sum_{t=1}^s (Q_t - \E[Q_t|\cG_{t-1}])$ for all $s\geq 1$, $Y_0=0$. Then $\{Y_s\}_{s=0}^\infty$ is a martingale with respect to the filtration $\{\cG_s\}_{s=0}^\infty$ with increments $Z_s = Y_s-Y_{s-1}=Q_s - \E[Q_s|\cG_{s-1}]$ satisfying $\E[Z_s|\cG_{s-1}]=0, Z_s \leq 1$ for all $s\geq 1$.
\end{lemma}
\begin{proof}
To show $\{Y_s\}_{s=0}^\infty$ is a martingale with respect to $\{\cG_s\}_{s=0}^\infty$, we need to show that $Y_s$ is $G_s$ measurable for all $s$ and $\E[Y_s|\cG_{s-1}]=Y_{s-1}$.

\noindent \underline{Measurability:} First note that by definition of $\cG_s$, $\tau_{t,J_t}, R_{t,J_t}$ are all $\cG_s$-measurable for $t \leq s$. Then, for each $i$, either $t$ is in a phase later than $i$ so $S_{i-1,j}$ and $S_i$ are $\cG_t$-measurable, or $S_{i-1,j}$ and $S_i$ are not $\cG_t$-measurable, but $\one \{t \geq S_{i,j}\}=0$ so $\one \{t \geq S_{i,j}\}$ is $\cG_t$-measurable. In the first case, since $S_{i-1,j}$ and $S_i$ are $\cG_t$-measurable $A_{i,t}  \one\{S_{i-1,j} \leq t \leq S_i - \nu_i\}$ is $\cG_t$-measurable. In the second case, $A_{i,t} \one \{S_{i-1,j} \leq t \leq S_i -1\}= A_{i,t} \one\{\{S_{i-1,j} \leq t\} \one\{ t \leq S_i -1\}=0$ so it is also $\cG_t$-measurable. Similarly, if $t$ is after $S_i$ , $S_i$ and $S_{i,j}$ will be $\cG$-measurable or $\one \{S_i  \leq t \leq S_{i,j}-1\}=0$. In both cases, $ B_{i,t} \one \{S_i  \leq t \leq S_{i,j}-1\}$ is $\cG_t$-measurable. Hence, $Q_t$ is $\cG_t$-measurable, and also $Q_t$ is $\cG_s$ measurable for any $s \geq t$. It then follows that $Y_s$ is $\cG_s$-measurable for all $s$.

\noindent \underline{Expectation:} Since $Q_t$ is $\cG_s$ measurable for all $t \leq s$,
\begin{align*}
\E[Y_s |\cG_{s-1}] &= \E\bigg[ \sum_{t=1}^s (Q_t - \E[Q_t|\cG_{t-1}]) |\cG_{s-1} \bigg] 
\\ &=  \E\bigg[ \sum_{t=1}^{s-1} (Q_t-\E[Q_t|\cG_{t-1}]) |\cG_{s-1} \bigg]  + \E [(Q_s-\E[Q_s|\cG_{s-1}])|\cG_{s-1}]
\\ & = \sum_{t=1}^{s-1} (Q_t-\E[Q_t|\cG_{t-1}]) +  \E[Q_s|\cG_{s-1}] - \E[Q_s|\cG_{s-1}]
\\ & = \sum_{t=1}^{s-1} (Q_t-\E[Q_t|\cG_{t-1}]) = Y_{s-1}
\end{align*}
Hence, $\{Y_s\}_{s=0}^\infty$ is a martingale with respect to the filtration $\{\cG_s\}_{s=0}^\infty$.

\noindent \underline{Increments:} For any $s=1, \dots$, we have that 
\[ Z_s = Y_s - Y_{s-1} = \sum_{t=1}^s (Q_t - \E[Q_t|\cG_{t-1}]) - \sum_{t=1}^{s-1} (Q_t - \E[Q_t|\cG_{t-1}]) = Q_s - \E[Q_s|\cG_{s-1}]. \]
Then,
\[ \E[Z_s |\cG_{s-1}] = \E[Q_s - \E[Q_s|\cG_{s-1}] |\cG_{s-1}] = \E[Q_s|\cG_{s-1}] - \E[Q_s|\cG_{s-1}] = 0. \]
Lastly, since for any $t$, there is only one $i$ where one of $\one\{S_{i-1,j} \leq t \leq S_i - 1\}=1$ or $\one \{S_i  \leq t \leq S_{i,j}-1\}=1$ (and they cannot both be one), and since $R_{t,J_t} \in [0,1]$, $A_{i,t}, B_{i,t} \leq 1$, so it follows that $Z_s = Q_s-\E[Q_s|\cG_{s-1}] \leq 1$ for all $s$.
\end{proof}

\begin{lemma} \label{lem:varqgen}
For any $t$, let $Z_t=Q_t - \E[Q_t|\cG_{t-1}]$, then, for any $s<S_{m,j}$,
\[ \sum_{t=1}^s\E[Z_t^2|\cG_{t-1}] \leq 2m\E[\tau]. \] 
\end{lemma}
\begin{proof}
First note that
\begin{align*}
\sum_{t=1}^s \E[Z_t^2|\cG_{t-1}] &= \sum_{t=1}^s \Var(Q_t|\cG_{t-1}) \leq \sum_{t=1}^s \E[Q_t^2|\cG_{t-1}]
\\ & = \sum_{t=1}^s \E\bigg[ \bigg(\sum_{i=1}^m (A_{i,t} \one\{S_{i-1,j} \leq t \leq S_i - 1\} + B_{i,t} \one \{S_i \leq t \leq S_{i,j}-1\} ) \bigg)^2 \bigg|\cG_{t-1} \bigg].
\end{align*}
Then, given $\cG_{t-1}$, all indicator terms $\one\{S_{i-1,j} \leq t \leq S_i -1\}$ and $\one \{S_i \leq S_{i,j}-1\}$ for all $i=1, \dots, m$ are measurable and only one can be non zero. Hence, all interaction terms in the expansion of the quadratic are 0 and so we are left with 
\begin{align*}
\sum_{t=1}^s \E[Z_t^2|\cG_{t-1}] & \leq \sum_{t=1}^s \E\bigg[ \bigg(\sum_{i=1}^m (A_{i,t} \one\{S_{i-1,j} \leq t \leq S_i -1\} + B_{i,t} \one \{S_i \leq t \leq S_{i,j}-1\} ) \bigg)^2 \bigg|\cG_{t-1} \bigg]
\\ & = \sum_{t=1}^s  \E\bigg[ \sum_{i=1}^m (A_{i,t}^2 \one\{S_{i-1,j} \leq t \leq S_i -1\}^2 + B_{i,t}^2 \one \{S_i \leq t \leq S_{i,j}-1\}^2 ) \bigg|\cG_{t-1} \bigg]
\\ & = \sum_{i=1}^m \sum_{t=1}^s \E[A_{i,t}^2  \one\{S_{i-1,j} \leq t \leq S_i -1\} |\cG_{t-1}] + \sum_{i=1}^m \sum_{t=1}^s \E[B_{i,t}^2 \one \{S_i \leq t \leq S_{i,j}-1\} |\cG_{t-1}]
\\ & \leq \sum_{i=1}^m \sum_{t=S_{i-1,j}}^{S_i-1} \E[A_{i,t}^2|\cG_{t-1}] + \sum_{i=1}^m \sum_{t=S_i}^{S_{i,j}-1} \E[B_{i,t}^2|\cG_{t-1}]. 
\end{align*}
Then, for any $i \geq 1$, 
\begin{align*}
\sum_{t=S_{i-1,j}}^{S_i-1} \E[A_{i,t}^2|\cG_{t-1}] & = \sum_{t=S_{i-1,j}}^{S_i-1} \E[R_{t,J_t}^2 \one\{\tau_{t,J_t} + t \geq S_i\}|\cG_{t-1}]
\\ & \leq \sum_{t=S_{i-1,j}}^{S_i-1} \E[ \one\{\tau_{t,J_t} + t \geq S_i\}|\cG_{t-1}]
\\ & = \sum_{s=0}^\infty \sum_{s'=s}^\infty \one\{S_{i-1,j}=s, S_i=s'\} \sum_{t=s}^{s'-1} \E[\one \{\tau_{t,J_t} +t \geq S_i \}|\cG_{t-1}]
\\ & = \sum_{s=0}^\infty \sum_{s'=s}^\infty \sum_{t=s}^{s'-1} \E[\one \{S_{i-1,j}=s, S_i=s', \tau_{t,J_t} +t \geq S_i \}|\cG_{t-1}]
	\\ \tag*{(\text{Since $\{ t \geq S_{i-1,j}, S_i=s'\} \in \cG_{t-1}$})}
\\ & = \sum_{s=0}^\infty \sum_{s'=s}^\infty \sum_{t=s}^{s'-1} \E[\one \{S_{i-1,j}=s, S_i=s', \tau_{t,J_t} +t \geq s' \}|\cG_{t-1}]
\\ & = \sum_{s=0}^ \infty \sum_{s'=s}^\infty \one\{S_{i-1,j}=s, S_i=s'\} \sum_{t=s}^{s'-1} \PP( \tau_{t,J_t} +t \geq s')
	\\ \tag*{(\text{Since $\{ t \geq S_{i-1,j}, S_i=s'\} \in \cG_{t-1}$})}
\\ & \leq \sum_{s=0}^ \infty \sum_{s'=s}^\infty \one\{S_{i-1,j}=s, S_i=s'\} \sum_{l=0}^\infty \PP(\tau > l)
\\ & \leq \E[\tau].
\end{align*}
Likewise, for any $i \geq 1$,
\begin{align*}
\sum_{t=S_i}^{S_{i,j}-1} \E[B_{i,t}^2|\cG_{t-1}] & = \sum_{t=S_i}^{S_{i,j}-1} \E[R_{t,J_t}^2 \one\{\tau_{t,J_t} + t \geq S_{i,j}\}|\cG_{t-1}]
\\ & \leq \sum_{t=S_i}^{S_{i,j}-1} \E[ \one\{\tau_{t,J_t} + t \geq S_{i,j}\}|\cG_{t-1}]
\\ & = \sum_{s=0}^\infty \sum_{s'=s}^\infty \one\{S_i=s, S_{i,j}=s'\} \sum_{t=s}^{s'-1} \E[\one \{\tau_{t,J_t} +t \geq S_{i,j} \}|\cG_{t-1}]
\\ & = \sum_{s=0}^\infty \sum_{s'=s}^\infty \sum_{t=s}^{s'-1} \E[\one \{S_i=s, S_{i,j}=s', \tau_{t,J_t} +t \geq S_{i,j} \}|\cG_{t-1}]
	\\ \tag*{(\text{Since $\{ t \geq S_i, S_{i,j}=s'\} \in \cG_{t-1}$})}
\\ & = \sum_{s=0}^\infty \sum_{s'=s}^\infty \sum_{t=s}^{s'-1} \E[\one \{S_i=s, S_{i,j}=s', \tau_{t,J_t} +t \geq s' \}|\cG_{t-1}]
\\ & = \sum_{s=0}^ \infty \sum_{s'=s}^\infty \one\{S_i=s, S_{i,j}=s'\} \sum_{t=s}^{s'-1} \PP( \tau_{t,J_t} +t \geq s')
	\\ \tag*{(\text{Since $\{ t \geq S_i, S_{i,j}=s'\} \in \cG_{t-1}$})}
\\ & \leq \sum_{s=0}^ \infty \sum_{s'=s}^\infty \one\{S_i=s, S_{i,j}=s'\} \sum_{l=0}^\infty \PP(\tau \geq l)
\\ & \leq \E[\tau]. 
\end{align*}
Hence, combining both terms and summing over the phases $m$ gives the result.
\end{proof}

\begin{lemma} \label{lem:pmart}
Let $Y'_s = \sum_{t=1}^s (\E[P_s|\cG_{s-1}] - P_s)$ for all $s\geq 1$, $Y'_0=0$. Then $\{Y'_s\}_{s=0}^\infty$ is a martingale with respect to the filtration $\{\cG_s\}_{s=0}^\infty$ with increments $Z'_s = Y'_s-Y'_{s-1}=\E[P_s|\cG_{s-1}] - P_s$ satisfying $\E[Z'_s|\cG_{s-1}]=0, Z'_s \leq 1$ for all $s\geq 1$.
\end{lemma}
\begin{proof}
The proof is similar to that of Lemma~\ref{lem:qmartgen}. To show $\{Y'_s\}_{s=0}^\infty$ is a martingale with respect to $\{\cG_s\}_{s=0}^\infty$, we need to show that $Y'_s$ is $G_s$ measurable for all $s$ and $\E[Y'_s|\cG_{s-1}]=Y'_{s-1}$. \\

\noindent \underline{Measurability:} As before,  by definition of $\cG_s$, $\tau_{t,J_t}, R_{t,J_t}$ are all $\cG_s$-measurable for $t \leq s$. Also,  we can reduce measurability again to measurability of $\one\{ \tau_{s,J_s} + s \geq U_{i,j}, S_{i,j} \leq s \leq U_{i,j}\}$. But, $\{U_{i,j} = s'\} \cap \{S_{i,j} \leq s\} \in \cG_{s}$ for all $s' \in \mathbb{N}$ and $Y'_s$ is adapted to $\cG_s$.  \\

\noindent \underline{Increments:} For any $s\geq 1$, we have that 
\[ Z'_s = Y'_s - Y'_{s-1} = \sum_{t=1}^s ( \E[P_t|\cG_{t-1}] - P_t) - \sum_{t=1}^{s-1} (\E[P_t|\cG_{t-1}] - P_t) =  \E[P_s|\cG_{s-1}] - P_s. \]
Then,
\[ \E[Z'_s |\cG_{s-1}] = \E[ \E[P_s|\cG_{s-1}] - P_s |\cG_{s-1}] = \E[P_s|\cG_{s-1}] - \E[P_s|\cG_{s-1}] = 0. \]
Lastly, since for any $t$ and $\omega \in \Omega$, there is at most one $i$ for which $\one\{S_{i,j} \leq t \leq U_{i,j}\}=1$, and by definition of $R_{t,J_t}$, $C_{i,t} \leq 1$, so it follows that $Z'_s = \E[P_s|\cG_{s-1}] - P_s \leq 1$ for all $s$.
\end{proof}

\begin{lemma} \label{lem:varp}
For any $t$, let $Z'_t= \E[P_t|\cG_{t-1}] - P_t$, then
\[ \sum_{t=1}^{U_{m,j}} \E[{Z'_t}^2|\cG_{t-1}] \leq m\E[\tau]. \] 
\end{lemma}
\begin{proof}
The proof is similar to that of Lemma~\ref{lem:varqgen}. First note that
\begin{align*}
\sum_{t=1}^{U_{m,j}} \E[{Z'_t}^2|\cG_{t-1}] &= \sum_{t=1}^{U_{m,j}} \Var(P_t|\cG_{t-1}) \leq \sum_{t=1}^{U_{m,j}} \E[P_t^2|\cG_{t-1}] \\
&= \sum_{t=1}^{U_{m,j}} \E\bigg[ \bigg(\sum_{i=1}^m (C_{i,t} \one\{S_{i,j} \leq t \leq U_{i,j}\} \bigg)^2 |\cG_{t-1} \bigg].
\end{align*}
Then, given $\cG_{t-1}$, all indicator terms $\one\{S_{i,j} \leq t \leq U_{i,j} \}$ for $i=1, \dots, m$ are measurable and at most one can be non zero. Hence, all interaction terms are 0 and so we are left with 
\begin{align*}
\sum_{t=1}^{U_{m,j}} \E[{Z'_t}^2|\cG_{t-1}] & \leq \sum_{t=1}^{U_{m,j}} \E\bigg[ \bigg(\sum_{i=1}^m (C_{i,t} \one\{S_{i,j} \leq t \leq U_{i,j} \}\bigg)^2 |\cG_{t-1} \bigg]
\\ 
& = \sum_{i=1}^m \sum_{t=1}^{U_{m,j}} \E[C_{i,t}^2  \one\{S_{i,j} \leq t \leq U_{i,j} \} |\cG_{t-1}]
\\ & \leq \sum_{i=1}^m \sum_{t=S_{i,j}}^{U_{i,j}} \E[C_{i,t}^2|\cG_{t-1}] 
\tag*{(since the indicator is $\cG_{t-1}$-measurable)} 
\\ & = \sum_{i=1}^m \sum_{t=S_{i,j}}^{U_{i,j}} \E[R_{t,J_t}^2 \one\{\tau_{t,J_t} + t > U_{i,j}\}|\cG_{t-1}]
\\ & \leq \sum_{i=1}^m  \sum_{t=S_{i,j}}^{U_{i,j}} \E[ \one\{\tau_{t,J_t} + t > U_{i,j}\}|\cG_{t-1}]
\\ & = \sum_{i=1}^m \sum_{s=0}^\infty \sum_{s'=s}^\infty \one\{S_{i,j}=s, U_{i,j}=s'\} \sum_{t=s}^{s'} \E[\one \{\tau_{t,J_t} +t > U_{i,j} \}|\cG_{t-1}]
\\ 
& = \sum_{i=1}^m \sum_{s=0}^\infty \sum_{s'=s}^\infty \sum_{t=s}^{s'} \E[\one \{S_{i,j}=s, U_{i,j}=s', \tau_{t,J_t} +t > U_{i,j} \}|\cG_{t-1}]
	\\ 
& = \sum_{i=1}^m \sum_{s=0}^\infty \sum_{s'=s}^\infty \sum_{t=s}^{s'} \E[\one \{S_{i,j}=s, U_{i,j}=s', \tau_{t,J_t} +t > s' \}|\cG_{t-1}]
\\ & = \sum_{i=1}^m \sum_{s=0}^ \infty \sum_{s'=s}^\infty \one\{S_{i,j}=s, U_{i,j}=s'\} \sum_{t=s}^{s'} \PP( \tau_{t,J_t} +t > s')
	\\ 
& \leq \sum_{i=1}^m \sum_{s=0}^ \infty \sum_{s'=s}^\infty \one\{S_{i,j}=s, U_{i,j}=s'\} \sum_{l=0}^\infty \PP(\tau > l)
\\ & \leq \sum_{i=1}^m \E[\tau] = m\E[\tau]. 
\end{align*} 
\end{proof}

\begin{lemma} \label{lem:abcboundsgen}
For $A_{i,t}, B_{i,t}$ and $C_{i,t}$ defined as in \eqref{eqn:abcdef}, let $\nu_i = n_i - n_{i-1}$ be the number of times each arm is played in phase $i$ and $j_i'$ be the arm played directly before arm $j$ in phase $i$. Then, it holds that, for any arm $j$ and phase $i \geq 1$,
\begin{enumerate}[(i)]
	\item $ \displaystyle  \sum_{t=S_{i-1,j}}^{S_i -1} \E[A_{i,t}|\cG_{t-1}] \leq \E[\tau]$
	\item $ \displaystyle \sum_{t=S_i}^{S_{i,j} -1} \E[B_{i,t}|\cG_{t-1}] \leq \E[\tau]  + \mu_{j_i'} \sum_{l=0}^{\nu_i} \PP(\tau > l)$
	\item $ \displaystyle \sum_{t=S_{i,j}}^{U_{i,j}} \E[C_{i,t}|\cG_{t-1}] = \mu_j \sum_{l=0}^{\nu_i} \PP(\tau > l)$
\end{enumerate}
\end{lemma}
\begin{proof}
We prove each statement individually. Several of the proofs are similar to those appearing in \cref{lem:varqgen,lem:varp}. 
\paragraph{Statement (i):}
\begin{align*}
\sum_{t=S_{i-1,j}}^{S_i -1} \E[A_{i,t}|\cG_{t-1}] & \leq \sum_{t=S_{i-1,j}}^{S_i-1} \E[ \one\{\tau_{t,J_t} + t \geq S_i\}|\cG_{t-1}]
\\ & = \sum_{s=0}^\infty \sum_{s'=s}^\infty \one\{S_{i-1,j}=s, S_i=s'\} \sum_{t=s}^{s'-1} \E[\one \{\tau_{t,J_t} +t \geq S_i \}|\cG_{t-1}]
\\ & = \sum_{s=0}^\infty \sum_{s'=s}^\infty \sum_{t=s}^{s'-1} \E[\one \{S_{i-1,j}=s, S_i=s', \tau_{t,J_t} +t \geq S_i \}|\cG_{t-1}]
	\\ \tag*{(\text{Since $\{ t \geq S_{i-1,j}, S_i=s'\} \in \cG_{t-1}$})}
\\ & = \sum_{s=0}^\infty \sum_{s'=s}^\infty \sum_{t=s}^{s'-1} \E[\one \{S_{i-1,j}=s, S_i=s', \tau_{t,J_t} +t \geq s' \}|\cG_{t-1}]
\\ & = \sum_{s=0}^ \infty \sum_{s'=s}^\infty \one\{S_{i-1,j}=s, S_i=s'\} \sum_{t=s}^{s'-1} \PP( \tau_{t,J_t} +t \geq s')
	\\ \tag*{(\text{Since $\{ t \geq S_{i-1,j}, S_i=s'\} \in \cG_{t-1}$})}
\\ & \leq \sum_{s=0}^ \infty \sum_{s'=s}^\infty \one\{S_{i-1,j}=s, S_i=s'\} \sum_{l=0}^\infty \PP(\tau> l)
\\ & = \sum_{l=0}^\infty \PP(\tau > l) = \E[\tau].
\end{align*}

\paragraph{Statement (iii):}
\begin{align*}
 \sum_{t=S_{i,j}}^{U_{i,j}} \E[C_{i,t}|\cG_{t-1}] & =  \sum_{t=S_{i,j}}^{U_{i,j}} \E[R_{t,J_t} \one\{\tau_{t,J_t} + t > U_{i,j}\}|\cG_{t-1}]
\\ & =  \sum_{s=0}^\infty \sum_{s'=s}^\infty \one\{S_{i,j}=s, U_{i,j}=s'\} \sum_{t=s}^{s'} \E[R_{t,J_t} \one \{\tau_{t,J_t} +t > U_{i,j} \}|\cG_{t-1}]
\\ & = \sum_{s=0}^\infty \sum_{s'=s}^\infty \sum_{t=s}^{s'} \E[R_{t,J_t} \one \{S_{i,j}=s, U_{i,j}=s', \tau_{t,J_t} +t > U_{i,j} \}|\cG_{t-1}]
	\\ \tag*{(\text{Since $\{ S_{i,j} =s, U_{i,j}=s'\} \in \cG_{t-1}$ for $s \leq t$})}
\\ & = \sum_{s=0}^\infty \sum_{s'=s}^\infty \sum_{t=s}^{s'} \E[R_{t,J_t} \one \{S_{i,j}=s, U_{i,j}=s', \tau_{t,J_t} +t > s' \}|\cG_{t-1}]
\\ & = \sum_{s=0}^ \infty \sum_{s'=s}^\infty \one\{S_{i,j}=s, U_{i,j}=s'\} \sum_{t=s}^{s'} \mu_j \PP( \tau_{t,J_t} +t > s')
	\\ \tag*{(\text{Since $\{ S_{i,j} = s, U_{i,j}=s'\} \in \cG_{t-1}$ and given $\cG_{t-1}$, $R_{t,J_t}$ and $\tau_{t,J_t}$ are independent})}
\\ & =  \sum_{s=0}^ \infty \sum_{s'=s}^\infty \one\{S_{i,j}=s, U_{i,j}=s'\} \mu_j \sum_{l=0}^{\nu_i}\PP(\tau > l)
\\ & = \mu_j \sum_{l=0}^{\nu_i}\PP(\tau > l)
\end{align*}

\paragraph{Statement (ii):} For statement (ii), we have that for $(i,j) \neq (1,1)$,
\begin{align*}
\sum_{t=S_i}^{S_{i,j} -1} \E[B_{i,t}|\cG_{t-1}] = \sum_{t=S_i}^{S_{i,j} -\nu_{i-1}-2} \E[B_{i,t}|\cG_{t-1}] + \sum_{t=S_{i,j} -\nu_{i-1} -1}^{S_{i,j} -1} \E[B_{i,t}|\cG_{t-1}].
\end{align*}
Then, $S_{i,j}$ is $\cG_{t-1}$ measurable for $t\geq S_i$, so we can use the same technique as for statement (i) to bound the first term. For the second term, since we will only be playing arm $j_i'$ for $S_{i,j}-\nu_{i-1}-1, \dots, S_{i,j}-1$, we can use the same technique as for statement (iii). Hence,
\[ \sum_{t=S_i}^{S_{i,j} -1} \E[B_{i,t}|\cG_{t-1}] \leq \sum_{l=\nu_{i-1}+1}^\infty \PP(\tau > l) + \mu_{j_i'} \sum_{l=0}^{\nu_{i-1}} \PP(\tau > l) \leq \E[\tau] + \mu_{j_i'} \sum_{l=0}^{\nu_i} \PP(\tau > l) . \]
Note that, for $(i,j)=(1,1)$, the amount seeping in will be 0, so using $\nu_0=0, \mu_{1_1}'=0$, the result trivially holds. Hence the result holds for all $i,j \geq 1$.
\end{proof}

\begin{restatable}{lemma}{lemaz}
\label{lem:az}
For any arm $j \in \{1, \dots, K\}$ and phase $m$, it holds that for any $\lambda>0$,
\[ \PP \bigg( \sum_{t \in T_j(m)} (R_{t,j} - \mu_j) \geq \lambda \bigg) \leq \exp \bigg\{ - \frac{2 \lambda^2}{n_m} \bigg \}.  \]
\end{restatable}
\begin{proof}
The result follows from Lemma~\ref{lem:concentration}.
When applying this lemma, we use $n=T$,  $m=n_m$, 
for $t=0,1,\dots,T$ set
$\cF_t = \sigma(X_1,\dots,X_t,R_{1,j},\dots,R_{t,j})$ 
and for $t=1,2,\dots,T$ define
$Z_t = R_{t,j}-\mu_j$ and
 $\epsilon_t = \one\{J_t=j,t\le U_{m,j}\}$. 
Note that 
	$T_j(m) = \{ t\in \{1,\dots,T\}\,: \epsilon_t = 1 \}$ and hence
	$\sum_{t\in T_j(m)} (R_{t,j}-\mu_j) = \sum_{t=1}^T \epsilon_t (R_{t,j}-\mu_j)$.
Further, $\sum_{t=1}^T \epsilon_t = |T_j(m)| \le n_m$ with probability one.

Fix $1\le t \le T$.
We now argue that $\epsilon_t$ is $\cF_{t-1}$-measurable.
First, notice that by the definition of \ODAAF, the index $M$ of the phase that $t$ belongs to can be calculated based on the observations $X_1,\dots,X_{t-1}$ up to time $t-1$.
Since $t\le U_{m,j}$ is equivalent to 
whether for this phase index $M$, the inequality $M\le m$ holds, 
it follows that $\{t\le U_{m,j}\}$ is $\cF_{t-1}$-measurable.
The same holds for $\{J_t=j\}$ for the same reason.
Hence, it follows that $\epsilon_t$ is indeed $\cF_{t-1}$-measurable.

Now, $Z_t$ is $\cF_t$-measurable as 
$R_{t,j}$ is clearly $\cF_t$-measurable.
Furthermore, by our assumptions on $(R_{t,j})_{t,j}$ and $(X_t)_t$, 
$\EE{ R_{t,j}| \cF_{t-1}} = \mu_j$ also holds, implying that
$Z_t$ also satisfies the conditions of the lemma with $a=-\mu_j$ and $c=1$.
Thus, the result follows by applying Lemma~\ref{lem:concentration}.
\end{proof}

We now bound each term of the decomposition in \eqref{eqn:decompgen} in turn.
\paragraph{Bounding Term I.:}
For Term I., we use Lemma~\ref{lem:az} to get that with probability greater than $1-\frac{1}{T\tilde \Delta_m^2}$,
 \[ \sum_{i=1}^m \sum_{t=S_{i,j}}^{U_{i,j}} (R_{t,J_t} - \mu_j) \leq \sqrt{\frac{n_m \log(T\tilde \Delta_m^2)}{2}}. \]
 
 \paragraph{Bounding Term II.:}
 For Term II., we will use Freedmans inequality (Theorem~\ref{thm:freedman}). From Lemma~\ref{lem:qmartgen}, $\{Y_s\}_{s=0}^\infty$ with $Y_s = \sum_{t=1}^s (Q_t- \E[Q_t|\cG_{t-1}])$ is a martingale with respect to $\{ \cG_s\}_{s=0}^\infty$ with increments $\{Z_s\}_{s=0}^\infty$ satisfying $\E[Z_s|\cG_{s-1}]=0$ and $Z_s \leq 1$ for all $s$. Further, by Lemma~\ref{lem:varqgen}, $\sum_{t=1}^s \E[Z_t^2 |\cG_{t-1}] \leq 2m \E[\tau] \leq \frac{6m\times 2^m \E[\tau]}{12} \leq n_m/12$ with probability 1. Hence we can apply Freedman's inequality to get that with probability greater than $1- \frac{1}{T\tilde \Delta_m^2}$,
 \[  \sum_{t=1}^{S_{m,j}} (Q_t - \E[Q_t|\cG_{t-1}]) \leq \frac{2}{3} \log(T\tilde \Delta_m^2) + \sqrt{\frac{1}{12} n_m \log(T\tilde \Delta_m^2)}. \] 
 
 \paragraph{Bounding Term III.:}
 For Term III., we again use Freedman's inequality (Theorem~\ref{thm:freedman}) but using Lemma~\ref{lem:pmart} to show that $\{Y'_s\}_{s=0}^\infty$ with $Y'_s = \sum_{t=1}^s (\E[P_t|\cG_{t-1}]- P_t)$ is a martingale with respect to $\{ \cG_s\}_{s=0}^\infty$ with increments $\{Z'_s\}_{s=0}^\infty$ satisfying $\E[Z'_s|\cG_{s-1}]=0$ and $Z'_s \leq 1$ for all $s$. Further, by Lemma~\ref{lem:varp}, $\sum_{t=1}^s \E[Z_t^2 |\cG_{t-1}] \leq m \E[\tau] \leq n_m/12$ with probability 1. Hence, with probability greater than $1- \frac{1}{T\tilde \Delta_m^2}$,
 \[  \sum_{t=1}^{U_{m,j}} (\E[P_t|\cG_{t-1}] - P_t) \leq \frac{2}{3} \log(T\tilde \Delta_m) + \sqrt{\frac{1}{12} n_m \log(T\tilde \Delta_m^2)}. \]
 
 \paragraph{Bounding Term IV.:}
We bound term IV. using Lemma~\ref{lem:abcboundsgen}, 
 \begin{align*}
 & \hspace{-10pt} \sum_{t=1}^{S_{m,j}} \E[Q_t|\cG_{t-1}] - \sum_{t=1}^{U_{m,j}} \E[P_t|\cG_{t-1}] 
 \\ & = \sum_{t=1}^{S_{m,j}} \E \bigg[ \sum_{i=1}^m (A_{i,t} \one\{S_{i-1,j} \leq t \leq S_i -1\} + B_{i,t} \one \{S_i  \leq t \leq S_{i,j}-1\}) \bigg| \cG_{t-1} \bigg] 
 	\\ & \hspace{50pt} - \sum_{t=1}^{U_{m,j}} \E \bigg[  \sum_{i=1}^m C_{i,t} \one\{S_{i,j} \leq t \leq U_{i,j}\} \bigg| \cG_{t-1}\bigg]
\\ & = \sum_{i=1}^m\sum_{t=1}^{S_{m,j}} \E[ A_{i,t} \one\{S_{i-1,j} \leq t \leq S_i -1\}|\cG_{t-1}] + \sum_{i=1}^m\sum_{t=1}^{S_{m,j}} \E[B_{i,t} \one \{S_i \leq t \leq S_{i,j}-1\}|\cG_{t-1}] 
	\\ & \hspace{50pt} - \sum_{i=1}^m \sum_{t=1}^{U_{m,j}} \E[C_{i,t} \one\{S_{i,j} \leq t \leq U_{i,j}\} |\cG_{t-1}]
\\ & = \sum_{i=1}^m \bigg( \sum_{t=S_{i-1,j}}^{S_i -1} \E[A_{i,t}|\cG_{t-1}] + \sum_{t=S_i }^{S_{i,j}-1} \E[B_{i,t}|\cG_{t-1}] - \sum_{t=S_{i,j}}^{U_{i,j}} \E[C_{i,t}|\cG_{t-1}] \bigg)
\\ &\leq \sum_{i=1}^m \bigg( 2\E[\tau] + \mu_{j_i'} \sum_{l=0}^{\nu_i} \PP(\tau > l) - \mu_j \sum_{l=0}^{\nu_i} \PP(\tau > l) \bigg) \leq 3m\E[\tau].
 \end{align*}
since $R_{t,j} \in [0,1]$.

\paragraph{Combining all terms:}
To get the final high probability bound, we sum the bounds for each term I.-IV.. Then, with probability greater than $1-\frac{3}{T\tilde \Delta_m^2}$, either $j \notin \cA_m$ or arm $j$ is played $n_m$ times by the end of phase $m$ and 
\begin{align*}
\frac{1}{n_m}\sum_{t \in T_j(m)} (X_t - \mu_j) &\leq \frac{4\log(T\tilde \Delta_m^2)}{3n_m} + \bigg(\frac{2}{\sqrt{12}} + \frac{1}{\sqrt{2}}\bigg)\sqrt{\frac{\log(T \tilde \Delta_m^2)}{n_m}} + \frac{3m\E[\tau]}{n_m} 
\\ &\leq \frac{4\log(T\tilde \Delta_m^2)}{3n_m} + \sqrt{\frac{2\log(T \tilde \Delta_m^2)}{n_m}} + \frac{3m\E[\tau]}{n_m} = w_m. 
\end{align*}

\paragraph{Defining ${\bf n_m}$:}
Setting
\begin{align}
n_m =  \bigg \lceil \frac{1}{\tilde \Delta_m^2} \bigg( \sqrt{2\log(T\tilde \Delta_m^2)} + \sqrt{2 \log(T\tilde \Delta_m^2) +\frac{8}{3}\tilde \Delta_m \log(T\tilde \Delta_m^2) + 6\tilde \Delta_m m \E[\tau]}\bigg)^2 \bigg \rceil.  \label{eqn:ngen}
\end{align}
ensures that $w_m \leq \frac{\tilde \Delta_m}{2}$ which concludes the proof.
\end{proof}

\subsection{Regret Bounds} \label{app:reggen}
Here we prove the regret bound in Theorem~\ref{thm:reggenNew} under Assumption~\ref{assum:1} and the choice of $n_m$ given by \eqref{eqn:ngen}. Under Assumption~\ref{assum:1}, the bridge period is not necessary so the results here hold for the version of Algorithm~\ref{alg:skeleton} with the bridge period omitted. Note that if we were to include the bridge period, we would be playing each arm at most $2n_m$ times by the end of phase $m$ so our regret would simply increase by a factor of 2. 

\thmreggenNew*
\begin{proof}
Our proof is a restructuring of the proof of \cite{auer2010ucb}.
For any arm $j$, define $M_j$ to be the random variable representing the phase when arm $j$ is eliminated in.
We set $M_j = \infty$ if the arm did not get eliminated before time step $T$.
Note that if $M_j$ is finite, $j\in \cA_{M_j}$  (this also means that $\cA_{M_j}$ is well-defined)
and if $\cA_{M_j+1}$ is also defined ($M_j$ is not the last phase) then $j\not\in \cA_{M_j+1}$. 
We also let $m_j$ denote the phase arm $j$ \emph{should} be eliminated in, that is $m_j= \min \{\,m\ge 1\,:\, \tilde{\Delta}_m < \frac{\Delta_j}{2}\,\}$. From the definition of $\tilde \Delta_m$ in our algorithm, we get the relations 
\begin{align}
2^{m_j} = \frac{1}{\tilde\Delta_{m_j}} \leq \frac{4}{\Delta_j} < \frac{1}{\tilde \Delta_{m_j+1}} \quad \text{ and } \quad  \frac{\Delta_j}{4} \leq \tilde \Delta_{m_j} \leq \frac{\Delta_j}{2}\,. \label{eqn:deltabounds}
\end{align} 
Define $N_j = \sum_{t=1}^T \one\{ J_t = j \}$ be the number of times arm $j$ is used and
let $\Reg_T^{(j)} = N_j \Delta_j$ be the ``pseudo''-regret contribution from each arm $1 \leq j \leq K$ so that $\E[\Reg_T] = \E \bigg[ \sum_{j=1}^K \Reg_T^{(j)} \bigg]$.
Let $M^*$ be the round when the optimal arm $j^*$ is eliminated. 
Hence, 
\begin{align*}
\E[\Reg_T] = \E\bigg[ \sum_{j=1}^K \Reg_T^{(j)} \bigg] 
= \underbrace{\E\bigg[ \sum_{j=1}^K \Reg_T^{(j)} \onep{M^*\ge m_j} \bigg]}_{\text{Term I.}}
   + \underbrace{\E\bigg[ \sum_{j=1}^K \Reg_T^{(j)} \onep{M^*<m_j} \bigg]}_{\text{Term II.}}\,.
\end{align*}

We will bound the regret in each of these cases in turn. To do so, we need the following results which consider the probabilities of confidence bounds failing and arms being eliminated in the incorrect rounds. 
\begin{lemma} \label{lem:probelim}
For any suboptimal arm $j$, 
\[ \PP(M_j > m_j \text{ and } M^* \geq m_j) \leq \frac{6}{T\tilde \Delta_{m_j}^2}\,. \]
\end{lemma}
\begin{proof}
Define
\[ E = \{\bar X_{m_j,j} \leq \mu_j + w_{m_j} \} \quad \text{ and } \quad H = \{\bar X_{m_j,j^*} > \mu^* - w_{m_j} \}\,. \]
If both $E$ and $F$ occur, it follows that,
\begin{align*}
\bar{X}_{m_j,j} &\leq \mu_j +  w_{m_j}
\\ & = \mu_j^* - \Delta_j + w_{m_j}  && \text{(since $\Delta_j = \mu_{j^*} - \mu_j$)}
\\ & \leq \bar{X}_{m_j,j^*} + w_{m_j} - \Delta_j +w_{m_j}
\\ & < \bar{X}_{m_j,j^*} -2 \tilde \Delta_{m_j} + 2w_{m_j}  && \text{(by \eqref{eqn:deltabounds})}
\\ & \leq \bar{X}_{m_j,j^*}  - \tilde\Delta_{m_j} && \text{(since $n_m$ is  such that $w_m \leq \tilde \Delta_m/2$)}
\end{align*}
and arm $j$ would be eliminated.
Hence, on the event $M^*\ge m_j$, $M_j\le m_j$.
Thus, $M^* \ge m_j$ and $M_j>m_j$ imply that either $E$ or $H$ does not occur
and so $\PP(M_j > m_j \text{ and } M^* \geq m_j)  \leq \PP( \{E^c \cup H^c\} \cap \{j,j^* \in \cA_{m_j}\}) 
\le \PP(E^c \cap j \in \cA_{m_j} ) + \PP(H^c \cap j^* \in \cA_{m_j})$.
Using \cref{lem:eb1}, we then get that,
\[
\PP(M_j \geq m_j \text{ and } M^* \geq m_j) \le \frac{6}{T \tilde \Delta_{m_j}^2}. 
 \]
\end{proof}

Note that the random set $\cA_m$ may not be defined for certain $\omega\in \Omega$.
That is, $\cA_m$ is a partially defined random element. For convenience, we
modify the definition of $\cA_m$ so that it is an emptyset for any $\omega$ when it is not defined by the previous definition. 
Define the event $F_j(m) = \{ \bar{X}_{m,j^*}  < \bar{X}_{m,j} - \tilde \Delta_m \} \cap \{j,j^* \in \mathcal{A}_m\} $ 
to be the event that arm $j^*$ is eliminated by arm $j$ in phase $m$ (given our note on $\cA_m$, this is well-defined).
The probability of this occurring is bounded in the following lemma.
\begin{lemma} \label{lem:probelimjstar}
The probability that the optimal arm $j^*$ is eliminated in round $m<\infty$ by the suboptimal arm $j$ is bounded by
\[ 
\PP( F_j(m) ) \leq \frac{6}{T \tilde \Delta_m^2}\,. 
\]
\end{lemma}
\begin{proof}
First note that for a suboptimal arm $j$ to eliminate arm $j^*$ in round $m$, both $j$ and $j^*$ must be active in round $m$ and $\bar X_{m,j} - w_m > \bar X_{m,j^*} + w_m$. Hence,
\begin{align*}
\PP(F_j(m)) &= \PP( j,j^* \in \cA_m \text{ and } \bar X_{m,j} - w_m > \bar X_{m,j^*} + w_m) 
\end{align*}
Then, observe that if
\[ E = \{\bar X_{m,j} \leq \mu_j + w_m \} \quad \text{ and } \quad H = \{\bar X_{m,j^*} > \mu^* - w_m \} \]
both hold in round $m$, it follows that,
\begin{align*}
\bar{X}_{m,j} - \tilde \Delta_m \leq \mu_j + w_m - \tilde \Delta_m \leq \mu_j - \frac{\tilde \Delta_m}{2} \leq \mu_{j^*} - \frac{\tilde \Delta_m}{2} \leq \bar X_{m,j^*} + w_m- \frac{\tilde \Delta_m}{2} \leq \bar{X}_{m,j^*}
\end{align*}
so arm $j^*$ will not be eliminated by arm $j$ in round $m$. Hence, for arm $j^*$ to be eliminated by arm $j$ in round $m$, one of $E$ or $H$ must not occur and the probability of this is bounded by \cref{lem:eb1} as,
\[\PP(F_j(m)) \leq \PP((E^C \cup H^C) \cap (j,j^* \in \mathcal{A}_m)) \leq  \PP(E^C \cap (j \in \cA_m)) + \PP(H^C \cap (j^* \in \cA_m)) \leq \frac{6}{T \tilde \Delta_m^2}. \]
\end{proof}

\noindent We now return to bounding the expected regret in each of the two cases.
\paragraph{Bounding Term I.}
To bound the first term, we consider the cases where arm $j$ is eliminated in or before the correct round ($M_j \leq m_j)$ and where arm $j$ is eliminated late ($M_j >m_j$). Then, by Lemma~\ref{lem:probelim},
\begin{align*}
\MoveEqLeft \E \bigg[ \sum_{j=1}^K \Reg_T^{(j)} \one\{M^* \geq m_j \} \bigg]   \\
&=  \E \bigg[ \sum_{j=1}^K \Reg_T^{(j)} \one\{M^* \geq m_j \} \one\{M_j \leq m_j\}\bigg] +  \E \bigg[ \sum_{j=1}^K \Reg_T^{(j)} \one\{M^* \geq m_j \} \one\{M_j >m_j\} \bigg] 
\\ 
&   \leq \sum_{j=1}^K \E[\Reg_T^{(j)} \one\{ M_j \leq m_j\} ] + \sum_{j=1}^K \E[ T \Delta_j \one\{ M^* \geq m_j , M_j > m_j\}]  \\ 
&  \leq \sum_{j=1}^K \Delta_j n_{m_j} + \sum_{j=1}^K T\Delta_j  \PP(M_j >m_j \text{ and } M^* \geq m_j) \\ 
&  \leq \sum_{j=1}^K \Delta_jn_{m_j} + \sum_{j=1}^K T\Delta_j \frac{6}{T\tilde \Delta_{m_j}^2} \\ 
& \leq \sum_{j=1}^K \bigg( \Delta_j n_{m_j} + \frac{24}{\tilde \Delta_{m_j}} \bigg) \leq \sum_{j=1}^K \bigg( \frac{96}{\Delta_j} + \Delta_j n_{m_j} \bigg)\,.
\end{align*}

\paragraph{Bounding Term II}
For the second term, let $m_{\max}=\max_{j \neq j^*} m_j$.
and recall that $N_j$ is the total number of times arm $j$ is played. 
Then, 
\begin{align*}
\E\bigg[ \sum_{j=1}^K \Reg_T^{(j)} \onep{M^*<m_j} \bigg]
&  =  \E \bigg[ \sum_{m=1}^{m_{\max}} \sum_{j: m<m_j } \Reg_T^{(j)} \one\{M^*=m\} \bigg] \\
 & = \sum_{m=1}^{m_{\max}} \E \bigg[ \one\{M^*=m\}\sum_{j: m_j > m} \Reg_T^{(j)} \bigg] \\ 
&  = \sum_{m=1}^{m_{\max}} \E \bigg[ \one\{M^*=m\}\sum_{j: m_j > m} N_j \Delta_j \bigg] \\ 
& \leq \sum_{m=1}^{m_{\max}} \E \bigg[ \one\{M^*=m\} T\max_{j: m_j > m} \Delta_j \bigg]\\ 
&  \leq \sum_{m=1}^{m_{\max}} 4\PP(M^*=m) T\tilde \Delta_m\,.
\end{align*}

Now consider the probability that arm $j^*$ is eliminated in round $m$. This includes the probability that it is eliminated by any suboptimal arm. For arm $j^*$ to be eliminated in round $m$ by a suboptimal arm with $m_j<m$, arm $j$ must be active ($M_j>m_j$) and the optimal arm must also have been active in round $m_j$ ($M^* \geq m_j$). Using this, it follows that
\begin{align*}
\PP(M^* = m) &= \sum_{j=1}^K \PP(F_j(m)) = \sum_{j: m_j <m} \PP(F_j(m)) + \sum_{j: m_j \geq m} \PP(F_j(m)) 
\\ &\leq \sum_{j: m_j < m} \PP(M_j >m_j \text{ and } M^*\geq m_j ) + \sum_{j: m_j \geq m} \PP(F_j(m)).
\end{align*}
Then, using \cref{lem:probelim,lem:probelimjstar} and summing over all $m\leq M$ gives,
\begin{align*}
&\sum_{m=1}^{m_{\max}} \bigg(\sum_{j: m_j <m} 4 \PP(M_j >m_j \text{ and } M^*\geq m_j ) T \tilde \Delta_m + \sum_{j: m_j \geq m} 4 \PP(F_j(m)) T \tilde \Delta_{m} \bigg) 
\\ &\leq \sum_{m=1}^{m_{\max}} \bigg(\sum_{j: m_j <m} 4 \frac{6}{T \tilde \Delta_{m_j}^2} T \frac{ \tilde \Delta_{m_j}}{2^{m-m_j}} + \sum_{j: m_j \geq m} \frac{24}{T \tilde \Delta_m^2} T \tilde \Delta_{m} \bigg) 
\\ & \leq \sum_{j=1}^K \frac{24}{\tilde \Delta_{m_j}} \sum_{m=m_j}^{m_{\max}} 2^{-(m-m_j)} + \sum_{j=1}^K \sum_{m=1}^{m_j} \frac{24}{2^{-m}}
\\ & \leq \sum_{j=1}^K \frac{96 \cdot 2 }{\Delta_j} + \sum_{j=1}^K 24 \cdot 2^{m_j +1}
\\ & \leq \sum_{j=1}^K \frac{192}{\Delta_j} + \sum_{j=1}^K 48 \cdot \frac{4}{\Delta_j} = \sum_{j=1}^K \frac{384}{\Delta_j}. 
\end{align*}

\noindent Combining the regret from terms I and II gives,
 \[ \E [\Reg_T] \leq \sum_{j=1}^K \bigg( \frac{480}{\Delta_j} + \Delta_j n_{m_j} \bigg). \]
Hence, all that remains is to bound $n_m$ in terms of $\Delta_j, T$ and $d$,
\begin{align*}
n_{m_j} &=  \bigg \lceil \frac{1}{\tilde \Delta_{m_j}^2} \bigg( \sqrt{2\log(T\tilde \Delta_{m_j}^2)} + \sqrt{2 \log(T\tilde \Delta_{m_j}^2) +\frac{8}{3}\tilde \Delta_{m_j} \log(T\tilde \Delta_{m_j}^2) + 6\tilde \Delta_{m_j} m_j \E[\tau]}\bigg)^2 \bigg \rceil 
	\\ & \leq  \left \lceil \frac{1}{\tilde \Delta_{m_j}^2} \left( 8 \log(T\tilde \Delta_{m_j}^2)+ \frac{16}{3} \tilde \Delta_{m_j} \log(T\tilde \Delta_{m_j}^2) + 12 \tilde \Delta_{m_j} m_j \E[\tau]\right) \right \rceil
	\\ & \leq 1 + \frac{8 \log(T\Delta_j^2/4)}{\tilde \Delta_{m_j}^2} +\frac{16\log(T\Delta_j^2/4)}{3 \tilde \Delta_{m_j}} + \frac{12 \log_2(4/\Delta_j)\E[\tau]}{\tilde \Delta_{m_j}}
	\\ & \leq 1 + \frac{128\log(T \Delta_j^2)}{ \Delta_j^2} +\frac{32\log(T \Delta_j^2)}{3 \Delta_j} +\frac{96\log(4/\Delta_j)\E[\tau]}{\Delta_j},
\end{align*}
where we have used $(a+b)^2 \leq 2(a^2 + b^2)$ for $a,b \geq 0$ and $\log_2(x) \leq 2\log(x)$ for $x>0$.

Hence, the total expected regret from \ODAAF with bounded delays can be bounded by,
\begin{align}
 \E[\Reg_t] \leq \sum_{j=1:j \neq j^*}^K \bigg( \frac{128\log(T \Delta_j^2)}{\Delta_j}  +\frac{32}{3}\log(T \Delta_j^2) + 96\log(4/\Delta_j)\E[\tau] + \frac{480}{\Delta_j} + \Delta_j \bigg).  \label{eqn:regfull}
 \end{align}

\end{proof}

We now prove the problem independent regret bound,
\correggen*
\begin{proof}
Let
\[ \lambda = \sqrt{\frac{K\log(K) e^2}{T}} \]
and note that for $\Delta>\lambda$, $\log(T\Delta^2)/\Delta$ is a decreasing function of $\Delta$. Then, for some constants $C_1, C_2$, and using the previous theorem, we can bound the regret by,
\[ \E[\Reg_T] \leq \sum_{j: \Delta_j \leq \lambda} \E[\Reg_t^{(j)}] + \sum_{j: \Delta_j >\lambda} \E[\Reg_T^{(j)}] \leq \frac{KC_1 \log(T \lambda^2 )}{\lambda} + KdC_2 \log(1/\lambda) + T\lambda. \]
Then, subsituting the above value of $\lambda$ gives a worst case regret bound that scales with $O(\sqrt{KT\log(K)} + K\E[\tau]\log(T))$.

\end{proof}

\section{Results for Delays with Bounded Support} \label{app:bounded}

\subsection{High Probability Bounds} \label{app:cbbounded}
\lemebtwo*
\begin{proof}
Let 
\begin{align}
 w_m = \frac{4\log(T\tilde \Delta_m^2)}{3n_m} + \sqrt{\frac{2\log(T \tilde \Delta_m^2)}{n_m}} + \frac{2\E[\tau]}{n_m}. \label{eqn:wmbounded}
 \end{align}
We show that with probability greater than $1-\frac{12}{T\tilde \Delta_m^2}$, either $j \notin \cA_m$ or  $\frac{1}{n_m}\sum_{t \in T_j(m)} (X_t - \mu_j) \leq  w_m$.  For now, assume that $n_m \geq md$.

For arm $j$ and phase $m$, assume $j \in \cA_m$ and define $p_i$ to be the probability of the confidence bounds on arm $j$ failing at the end of each phase $i \leq m$, ie. $p_i \doteq \PP(\sum_{t \in T_j(i)} (X_t - \mu_j) \geq n_iw_i)$ with $p_0=0$.
Again, let $B_{i,t} = R_t \one \{\tau_{t,J_t} +t \geq S_{i,j}\}$ and $C_{i,t} = R_t \one\{ \tau_{t,J_t} +t > U_{i,j}\}$ (note that we don't need to consider $A_{i,t}$ since $\nu_i = n_i - n_{i-1} \geq d$ so all reward entering $[S_{i,j},U_{i,j}]$ will be from the last $\nu_i \geq d$ plays) and for any event $H$, let $\one_i \{H \} := \one \{H \cap \{j \in \cA_i \}\}$. Recall the filtration $\{\mathcal{G}_t\}_{t=0}^\infty$ from \eqref{eqn:Gdef} where $\mathcal{G}_t = \sigma(X_1,\dots, X_t, J_1, \dots, J_t, \tau_{1,J_1}, \dots, \tau_{t,J_t}, R_{1,J_1}, \dots, R_{t,J_t})$ and $\mathcal{G}_0 = \{\emptyset, \Omega\}$. 
Now, defining,
\begin{align*}
Q_t &= \sum_{i=1}^m B_{i,t} \one \{S_{i,j} - d -1 \leq t \leq S_{i,j}-1\} ),
\\ P_t &= \sum_{i=1}^m C_{i,t} \one\{S_{i,j} \leq t \leq U_{i,j}\},
\end{align*}
we use the decomposition
\begin{align*}
\sum_{t \in T_j(m)} (X_t - \mu_j) &= \sum_{i=1}^m \sum_{t=S_{i,j}}^{U_{i,j}} (X_t - \mu_j)
\\ &\leq  \sum_{i=1}^m \bigg( \sum_{t=S_{i-1,j}}^{S_{i,j}-1} R_{t,J_t} \one_i \{ \tau_{t,J_t} + t \geq S_{i,j}\} + \sum_{t=S_{i,j}}^{U_{i,j}} (R_{t,J_t} - \mu_j) - \sum_{t=S_{i,j}}^{U_{i,j}} R_{t,J_t} \one_i\{ \tau_{t,J_t} + t > U_{i,j}\} \bigg)
\\ & \leq \sum_{i=1}^m \bigg( \sum_{t=S_{i,j}-d}^{S_{i,j}-1} B_{i,t} + \sum_{t=S_{i,j}}^{U_{i,j}} (R_t -\mu_j)- \sum_{t=S_{i,j}}^{U_{i,j}} C_{i,t} \bigg)
\\ & = \sum_{i=1}^m \sum_{t=S_{i,j}}^{U_{i,j}}(R_{t,J_t} - \mu_j) + \sum_{t=1}^{S_{m,j}} Q_t - \sum_{t=1}^{U_{m,j}} P_t \nonumber
\\ & = \underbrace{\sum_{i=1}^m \sum_{t=S_{i,j}}^{U_{i,j}}(R_{t,J_t} - \mu_j)}_{\text{Term I.}} 
	+ \underbrace{\sum_{t=1}^{S_{m,j}} (Q_t - \E[Q_t| \cG_{t-1}])}_{\text{Term II.}}
	 + \underbrace{\sum_{t=1}^{U_{m,j}}( \E[P_t|\cG_{t-1}] - P_t)}_{\text{Term III.}} 
	\\ & \hspace{50pt}  + \underbrace{\sum_{t=1}^{S_{m,j}} \E[Q_t|\cG_{t-1}] - \sum_{t=1}^{U_{m,j}} \E[P_t|\cG_{t-1}].}_{\text{Term IV.}}
\end{align*}

\paragraph{Outline of proof}
Again, the proof continues by bounding each term of this decomposition in turn.
Note that we do not have the $A_{i,t}$ terms in this decomposition since there will be no reward from phase $i-1$ (before the bridge period) received in $[S_{i,j},U_{i,j}]$. We bound each of these terms with high probability. For terms I. and III., this is the same as in the general case (see the proof of \cref{lem:eb1}, Appendix~\ref{app:gen}),. For term II. we need the following results to show that $Z_t = Q_t - \E[Q_s|\cG_{t-1}]$ is a martingale difference (Lemma~\ref{lem:qmartbound}) and to bound its variance (Lemma~\ref{lem:varqbound}) before we can apply Freedman's inequality. The bound for term IV. is also different due to the bridge period and boundedness of the delay. After bounding each term, we collect them together and recursively calculate the probability with which the bounds hold.  

\begin{lemma} \label{lem:qmartbound}
Let $Y_s = \sum_{t=1}^s (Q_t - \E[Q_t|\cG_{t-1}])$ for all $s\geq 1$, and $Y_0=0$. Then $\{Y_s\}_{s=0}^\infty$ is a martingale with respect to the filtration $\{\cG_s\}_{s=0}^\infty$ with increments $Z_s = Y_s-Y_{s-1}=Q_s - \E[Q_s|\cG_{s-1}]$ satisfying $\E[Z_s|\cG_{s-1}]=0, |Z_s| \leq 1$ for all $s\geq 1$.
\end{lemma}
\begin{proof}
To show $\{Y_s\}_{s=0}^\infty$ is a martingale we need to show that $Y_s$ is $\cG_s$-measurable for all $s$ and $\E[Y_s|\cG_{s-1}]=Y_{s-1}$. 

\noindent \underline{Measurability:} 
We show that $B_{i,s} \one\{S_{i,j} - d-1 \leq s \leq S_{i,j} -1 \}$ is $\cG_s$-measurable. This then suffices to show that $Y_s$ is $\cG_s$-measurable since the filtration $\cG_s$ is non-decreasing in $s$.

First note that by definition of $\cG_s$, $\tau_{t,J_t}, R_{t,J_t}$ are all $\cG_s$-measurable for $t \leq s$. Hence, it is sufficient to show that $\one\{\tau_{s,J_s} + s \geq S_{i,j}, S_{i,j} - d-1  \leq s \leq S_{i,j} -1 \}$  is $\cG_s$-measurable since the product of measurable functions is measurable. 
For any $s' \in \mathbb{N} \cup \{\infty\}$, $\{S_{i,j}=s', s'-d-1 \leq s\} \in \cG_s$ for $s \geq S_i - \nu_{i-1}$ and so the union $\bigcup_{s' \in \mathbb{N} \cup \{\infty\}} \{\tau_{s,J_s} + s \geq s', s'-d-1 \leq s \leq s'-1, S_{i,j} =s' \} = \{\tau_{s,J_s} + s \geq S_{i,j}, S_{i,j} -d -1\leq s \leq S_{i,j} -1 \} $ is an element of $\cG_s$. 

\noindent \underline{Increments:} Hence, $\{Y_s\}_{s=0}^\infty$ is a martingale with respect to the filtration $\{\cG_s\}_{s=0}^\infty$ if the increments conditional on the past are zero. For any $s\geq 1$, we have that 
\[ Z_s = Y_s - Y_{s-1} = \sum_{t=1}^s (Q_t - \E[Q_t|\cG_{t-1}]) - \sum_{t=1}^{s-1} (Q_t - \E[Q_t|\cG_{t-1}]) = Q_s - \E[Q_s|\cG_{s-1}]. \]
Then,
\[ \E[Z_s |\cG_{s-1}] = \E[Q_s - \E[Q_s|\cG_{s-1}] |\cG_{s-1}] = \E[Q_s|\cG_{s-1}] - \E[Q_s|\cG_{s-1}] = 0 \]
and so $\{Y_s\}_{s=0}^\infty$ is a martingale.

Lastly, since for any $t$ and $\omega \in \Omega$, there is at most one $i$ where $\one\{S_{i,j} -d \leq t \leq S_{i,j}-1\}(\omega)=1$, and by definition of $R_{t,J_t}$, $B_{i,t} \leq 1$, it follows that $|Z_s| = |Q_s-\E[Q_s|\cG_{s-1}]| \leq 1$ for all $s$.
\end{proof}

\begin{lemma} \label{lem:varqbound}
For any $t\geq 1$, let $Z_t=Q_t - \E[Q_t|\cG_{t-1}]$, then
\[ \sum_{t=1}^{S_{m,j} -1} \E[Z_t^2|\cG_{t-1}] \leq m\E[\tau]. \] 
\end{lemma}
\begin{proof}
Let us denote $S' \doteq S_{m,j}-1$. Observe that 
\begin{align*}
&\sum_{t=1}^{S'} \E[Z_t^2|\cG_{t-1}] = \sum_{t=1}^{S'} \Var(Q_t|\cG_{t-1}) \leq \sum_{t=1}^{S'} \E[Q_t^2|\cG_{t-1}]
 = \sum_{t=1}^{S'} \E\bigg[ \bigg(\sum_{i=1}^m (B_{i,t} \one \{S_{i,j}-d \leq t \leq S_{i,j}-1\} ) \bigg)^2 \Big|\cG_{t-1} \bigg].
\end{align*}
Then for all $i=1, \dots, m$, all indicator terms $\one \{S_{i,j}-d \leq t \leq S_{i,j}-1\}$ are $\cG_{t-1}$-measurable and only one can be non zero for any $\omega \in \Omega$.  Hence, for any $i,i' \leq m$, $i \not = i'$,  
\begin{align*}
B_{i,t} \times \one \{S_{i,j}-d-1 \leq t \leq S_{i,j}-1\} \times B_{i',t} \times  \one \{S_{i',j} -d- 1 \leq t \leq S_{i',j}-1\} = 0&, \\
\end{align*}
Using the above we see that 
\begin{align*}
\sum_{t=1}^{S'} \E[Z_t^2|\cG_{t-1}] & \leq \sum_{t=1}^{S'} \E\bigg[ \bigg(B_{i,t} \one \{S_{i,j}-d -1\leq t \leq S_{i,j}-1\} \bigg)^2 \Big|\cG_{t-1} \bigg]
\\ & = \sum_{t=1}^{S'}  \E\bigg[ \sum_{i=1}^m B_{i,t}^2 \one \{S_{i,j}-d-1 \leq t \leq S_{i,j}-1\}^2 \Big|\cG_{t-1} \bigg]
\\ & = \sum_{i=1}^m \sum_{t=1}^{S'} \E[B_{i,t}^2 \one \{S_{i,j}-d -1\leq t \leq S_{i,j}-1\}|\cG_{t-1}]
\\ 
\tag*{(using that the indicator is $\cG_{t-1}$-measurable)} \\
& \leq  \sum_{i=1}^m \sum_{t=S_{i,j}-d-1}^{S_{i,j}-1} \E[B_{i,t}^2|\cG_{t-1}]. 
\end{align*}
Then, for any $i \geq 1$,
\begin{align*}
\sum_{t=S_{i,j} - d-1}^{S_{i,j}-1} \E[B_{i,t}^2|\cG_{t-1}] & = \sum_{t=S_{i,j}-d-1}^{S_{i,j}-1} \E[R_{t,J_t}^2 \one\{\tau_{t,J_t} + t \geq S_{i,j}\}|\cG_{t-1}]
\\ & \leq \sum_{t=S_{i,j}-d-1}^{S_{i,j}-1} \E[ \one\{\tau_{t,J_t} + t \geq S_{i,j}\}|\cG_{t-1}]
\\ & = \sum_{s=0}^\infty \one\{S_{i,j}=s\} \sum_{t=s-d-1}^{s-1} \E[\one \{\tau_{t,J_t} +t \geq S_{i,j} \}|\cG_{t-1}]
\\ & = \sum_{s=0}^\infty \sum_{t=s-d-1}^{s-1} \E[\one \{S_{i,j}=s, \tau_{t,J_t} +t \geq S_{i,j} \}|\cG_{t-1}]
	\\ \tag*{(\text{Since $S_{i,j} \geq S_i$ and so, due to the bridge period, $\{S_{i,j}=s\} \in \cG_{t-1}$} for any  $t \geq s-d$)}
\\ & = \sum_{s=0}^\infty \sum_{t=s-d-1}^{s-1} \E[\one \{S_{i,j}=s, \tau_{t,J_t} +t \geq s \}|\cG_{t-1}]
\\ & = \sum_{s=0}^ \infty  \one\{S_{i,j}=s\} \sum_{t=s-d-1}^{s-1} \PP( \tau_{t,J_t} +t \geq s)
	\\ \tag*{(\text{Since $\{S_{i,j}=s\} \in \cG_{t-1}$} for any $t \geq s- d$)}
\\ & \leq \sum_{s=0}^ \infty \one\{S_{i,j}=s\} \sum_{l=0}^\infty \PP(\tau > l)
\\ & \leq \E[\tau]. 
\end{align*}
Combining all terms gives the result.
\end{proof}

We now return to bounding each term of the decomposition
\paragraph{Bounding Term I.:}
 For term II., as in \cref{lem:eb1}, we can use Lemma~\ref{lem:az} to get that with probability greater than $1-\frac{1}{T\tilde \Delta_m^2}$,
 \[ \sum_{i=1}^m \sum_{t=S_{i,j}}^{U_{i,j}} (R_{t,J_t} - \mu_j) \leq \sqrt{\frac{n_m \log(T\tilde \Delta_m^2)}{2}}. \]
 
 
 \paragraph{Bounding Term II.:}
 For Term II., we will use Freedmans inequality (Theorem~\ref{thm:freedman}). From Lemma~\ref{lem:qmartbound}, $\{Y_s\}_{s=0}^\infty$ with $Y_s = \sum_{t=1}^s (Q_t- \E[Q_t|\cG_{t-1}])$ is a martingale with respect to $\{ \cG_s\}_{s=0}^\infty$ with increments $\{Z_s\}_{s=0}^\infty$ satisfying $\E[Z_s|\cG_{s-1}]=0$ and $Z_s \leq 1$ for all $s$. Further, by Lemma~\ref{lem:varqbound}, $\sum_{t=1}^s \E[Z_t^2 |\cG_{t-1}] \leq m \E[\tau] \leq \frac{4 \times 2^m \E[\tau]}{8} \leq n_m/8$ with probability 1. Hence we can apply Freedman's inequality to get that with probability greater than $1- \frac{1}{T\tilde \Delta_m^2}$,
 \[  \sum_{t=1}^{S_{m,j}} (Q_t - \E[Q_t|\cG_{t-1}]) \leq \frac{2}{3} \log(T\tilde \Delta_m^2) + \sqrt{\frac{1}{8} n_m \log(T\tilde \Delta_m^2)}. \] 
 
 \paragraph{Bounding Term III.:}
 For Term III., we again use Freedman's inequality (Theorem~\ref{thm:freedman}). As in \cref{lem:eb1}, we use Lemma~\ref{lem:pmart} to show that $\{Y'_s\}_{s=0}^\infty$ with $Y'_s = \sum_{t=1}^s (\E[P_t|\cG_{t-1}] - P_t)$ is a martingale with respect to $\{ \cG_s\}_{s=0}^\infty$ with increments $\{Z'_s\}_{s=0}^\infty$ satisfying $\E[Z'_s|\cG_{s-1}]=0$ and $Z'_s \leq 1$ for all $s$. Further, by Lemma~\ref{lem:varp}, $\sum_{t=1}^s \E[Z_t^2 |\cG_{t-1}] \leq m \E[\tau] \leq n_m/8$ with probability 1. Hence, with probability greater than $1- \frac{1}{T\tilde \Delta_m^2}$,
 \[  \sum_{t=1}^{U_{m,j}} (\E[P_t|\cG_{t-1}] - P_t) \leq \frac{2}{3} \log(T\tilde \Delta_m^2) + \sqrt{\frac{1}{8} n_m \log(T\tilde \Delta_m^2)}. \]
 
 \paragraph{Bounding Term IV.:}
 For term IV., we consider the expected difference at each round $ 1\leq i \leq m$ and exploit the independence of $\tau_{t,J_t}$ and $R_{t,J_t}$. 
Consider first $i\geq 2$
and  let $j'_i$ be the arm played just before arm $j$ is played in the $i$th phase (allowing for $j'_i$ to be the last arm played in phase $i-1$). Then, much in the same way as Lemma~\ref{lem:varqbound},
\begin{align*}
\sum_{t=S_{i,j} - d-1}^{S_{i,j}-1} \E[B_{i,t}|\cG_{t-1}] & = \sum_{t=S_{i,j}-d-1}^{S_{i,j}-1} \E[R_{t,J_t} \one\{\tau_{t,J_t} + t \geq S_{i,j}\}|\cG_{t-1}]
\\ & = \sum_{s'=d+1}^\infty\sum_{s=s'}^\infty \one\{S_i = s', S_{i,j}=s\} \sum_{t=s-d}^{s-1} \E[R_{t,J_t} \one \{\tau_{t,J_t} +t \geq S_{i,j} \}|\cG_{t-1}]
\\ 
 & = \sum_{s'=d+1}^\infty\sum_{s=s'}^\infty  \sum_{t=s-d}^{s-1} \sum_{k=1}^K \E[R_{t,J_t} \one\{S_i = s', S_{i,j}=s, \tau_{t,J_t} +t \geq S_{i,j}, J_t =k \}|\cG_{t-1}]
	\\ \tag*{(Due to the bridge period $\{S_i=s', S_{i,j}=s\} \in \cG_{t-1}$ for $t \geq s-d \geq s'-d$)} \\
 & = \sum_{s'=d+1}^\infty\sum_{s=s'}^\infty  \sum_{t=s-d}^{s-1} \sum_{k=1}^K \one\{S_i = s', S_{i,j}=s, J_t=k\}
 \E[R_{t,k}  \one\{\tau_{t,k} +t \geq s\}|\cG_{t-1}] \\
&= \sum_{s'=d+1}^\infty\sum_{s=s'}^\infty  \sum_{t=s-d}^{s-1} \sum_{k=1}^K \mu_k \one\{S_i = s', S_{i,j}=s, J_t=k\}  \PP(\tau  \geq s-t) \\
&= \mu_{j'_i} \sum_{l=0}^{d-1} \PP(\tau > l).
\end{align*}
A similar argument works for $i=1, j>1$ with the simplification that $S_{i,j}$ is not a random quantity but known . Finally, for $i=1, j= 1$ the sum is $0$.
Furthermore, using a similar argument, for all $i,j$,
\begin{align*}
\sum_{t=S_{i,j}}^{U_{i,j}} \E[C_{i,t}|\mathcal{G}_{t-1}] &= \sum_{t=U_{i,j}-d + 1}^{U_{i,j}} \E[C_{i,t}|\mathcal{G}_{t-1}] \\
&= \sum_{s' = d+1}^\infty \sum_{s=s'}^\infty \sum_{t=s-d}^{s} \E[R_{t,j} \one\{\tau_{t,j} +t > s  \}  \one\{U_{i,j}=s, S_i = s' \}|\mathcal{G}_{t-1}] \\
&= \mu_j \sum_{s = d+1}^\infty \one\{U_{i,j}=s,S_i=s' \} \sum_{t=s-d}^{s}  \PP(\tau +t > s ) \\
&= \mu_j \sum_{l=0}^{d-1}  \PP(\tau > l ).
\end{align*}

Combining these we get the following bound for term IV for all $(i,j) \not= (1,1)$, 
 \begin{align*}
 \sum_{t=S_{i,j}-d-1}^{S_{i,j}-1} \E[B_{i,t}|\mathcal{G}_{t-1}] - \sum_{t=S_{i,j}}^{U_{i,j}} \E[C_{i,t}|\mathcal{G}_{t-1}] 
& \leq \mu_{j'_i} \sum_{l=0}^{d-1} \PP(\tau > l) - \mu_j \sum_{l=0}^{d-1} \PP(\tau > l)
 \\ & \leq |\mu_{j'_i} - \mu_j|  \E[\tau]. 
 \end{align*}
 If $(i,j) = (1,1)$ then we have the upper bounded by $\mu_1 \E[\tau] \leq \E[\tau] = \tilde \Delta_0 \E[\tau]$ since no pay-off seeps in and we define $\tilde \Delta_0=1$. 
 
Let $p_i$ be the probability that the confidence bounds for one arm hold in phase $i$ and $p_0=0$. Then, the probability that either arm $j'_i$ or $j$ is active in phase $i$ when it should have been eliminated in or before phase $i-1$ is less than $2p_{i-1}$. 
If neither arm should have been eliminated by phase $i$, this means that their mean rewards are within $\tilde \Delta_{i-1}$ of each other. Hence, with probability greater than $1-2p_{i-1}$,
 \[  \sum_{t=S_{i,j}-d-1}^{S_{i,j}-1} \E[B_{i,t}|\mathcal{G}_{t-1}] - \sum_{t=S_{i,j}}^{U_{i,j}} \E[C_{i,t}|\mathcal{G}_{t-1}]  \leq \tilde \Delta_{i-1} \E[\tau]. \]
Then, summing over all phases gives that with probability greater than $1- 2 \sum_{i=0}^{m-1} p_i $,
\[ \sum_{i=1}^m \bigg( \sum_{t=S_{i,j}-d-1}^{S_{i,j}-1} \E[B_{i,t}|\mathcal{G}_{t-1}] - \sum_{t=S_{i,j}}^{U_{i,j}} \E[C_{i,t}|\mathcal{G}_{t-1}] \bigg) \leq \E[\tau] \sum_{i=1}^m \tilde \Delta_{i-1} = \E[\tau] \sum_{i=0}^{m-1} \frac{1}{2^i} \leq 2 \E[\tau]. \]

\paragraph{Combining all Terms:}
To get the final high probability bound, we sum the bounds for each term I.-IV.. Then, with probability greater than $1-(\frac{3}{T\tilde \Delta_m^2} + 2\sum_{i=1}^{m-1} p_i)$ either $j \notin \cA_m$ or arm $j$ is played $n_m$ times by the end of phase $m$ and 
\begin{align*}
\frac{1}{n_m}\sum_{t \in T_j(m)} (X_t - \mu_j) &\leq \frac{4\log(T\tilde \Delta_m^2)}{3n_m} + \bigg(\frac{2}{\sqrt{8}} + \frac{1}{\sqrt{2}}\bigg)\sqrt{\frac{\log(T \tilde \Delta_m^2)}{n_m}} + \frac{2\E[\tau]}{n_m} 
\\ &\leq \frac{4\log(T\tilde \Delta_m^2)}{3n_m} + \sqrt{\frac{2\log(T \tilde \Delta_m^2)}{n_m}} + \frac{2\E[\tau]}{n_m} = w_m. 
\end{align*}
Using the fact that $p_0=0$ and substituting the other $p_i$'s using the recursive relationship $p_i = \frac{3}{T\tilde \Delta_i^2} + 2\sum_{l=1}^{i-1} p_l$ gives,
\begin{align*}
\frac{3}{T\tilde \Delta_m^2} + 2\sum_{i=0}^{m-1} p_i &= \frac{3}{T\tilde \Delta_m^2} + 2(\frac{3}{T \tilde \Delta_{m-1}^2} + 2(p_{m-2} + \dots + p_1) + p_{m-2} + \dots + p_1)
\\ &= \frac{3}{T \tilde \Delta_m^2} + 2(\frac{3}{T \tilde \Delta_{m-1}^2} + 3(p_{m-2} + \dots + p_1))
\\ & = \frac{3}{T \tilde \Delta_m^2} + 2(\frac{3}{T \tilde \Delta_{m-1}^2} + 3(\frac{3}{T\tilde \Delta_{m-2}^2} + 3(p_{m-3} + \dots + p_1))
\\ & \leq \sum_{i=1}^m 3^{m-i} \frac{3}{T\tilde \Delta_i^2}
\\ & = \frac{3}{T} \sum_{i=1}^m 3^{m-i} 2^{2i}
\\ & = \frac{3}{T} \sum_{i=1}^m 3^{m-i} 4i
\\ & = \frac{3}{T} \sum_{i=1}^m (\frac{3}{4})^{m-i} 4^{m-i} 4^i
\\ & = \frac{3 \times 4^m }{T} \sum_{i=1}^m (\frac{3}{4})^{m-i} 
\\ & \leq \frac{12}{T \tilde \Delta_m^2}.
\end{align*}
Hence, with probability greater than $1-\frac{12}{T \tilde \Delta_m^2}$, either $j \notin \cA_m$ or $\frac{1}{n_m}\sum_{t \in T_j(m)} (X_t - \mu_j)  \leq w_m$.

\paragraph{Defining ${\bf n_m}$:}
The above results rely on the assumption that $n_m \geq md$, so that only the previous arm can corrupt our observations. In practice, if $d$ is too large then we will not want to play each active arm $d$ times per phase because we will end up playing sub-optimal arms too many times. In this case, it is better to ignore the bound on the delay and use the results from \cref{lem:eb1} to set $n_m$ as in \eqref{eqn:ngen}. Formalizing this gives
\begin{align}
n_m = \max\bigg\{ m \tilde d_m, \bigg \lceil \frac{1}{\tilde \Delta_m^2} \bigg( \sqrt{2\log(T\tilde \Delta_m^2)} + \sqrt{2 \log(T\tilde \Delta_m^2) +\frac{8}{3}\tilde \Delta_m \log(T\tilde \Delta_m^2) + 4\tilde \Delta_m \E[\tau]}\bigg)^2 \bigg \rceil \bigg \} \label{eqn:nbounded}
\end{align}
where $\tilde d_m = \min\{ d, \frac{\eqref{eqn:ngen}}{m}\}$. This ensures that if $d$ is small, we play each active arm enough times to ensure that $w_m \leq \frac{\tilde \Delta_m}{2}$ for $w_m$ in \eqref{eqn:wmbounded}. Similarly, for large $d$, by \cref{lem:eb1}, we know that $n_m$ is suffiently large to guarantee $w_m \leq \frac{\tilde \Delta_m}{2}$ for $w_m$ from \eqref{eqn:wmgen}. 
\end{proof}

\subsection{Regret Bounds} \label{app:regbounded}
We now prove the regret bound given in Theorem~\ref{thm:regboundedNew}. Note that for these results, it is necessary to use the bridge period of the algorithm.
\thmregboundedNew*
\begin{proof}
For any sub-optimal arm $j$, define $M_j$ to be the random variable representing the phase arm $j$ is eliminated in and note that if $M_j$ is finite, $j\in \cA_{M_j}$ but $j\not\in \cA_{M_j+1}$. Then let $m_j$ be the phase arm $j$ \emph{should} be eliminated in, that is $m_j= \min \{m| \tilde{\Delta}_m < \frac{\Delta_j}{2}\}$ and note that, from the definition of $\tilde \Delta_m$ in our algorithm, we get the relations
\begin{align}
2^{m} = \frac{1}{\tilde\Delta_{m}}, \quad 2\tilde \Delta_{m_j} = \tilde\Delta_{m_j-1} \geq \frac{\Delta_j}{2} \quad \text{ and so, } \quad  \frac{\Delta_j}{4} \leq \tilde \Delta_{m_j} \leq \frac{\Delta_j}{2}. \label{eqn:deltabounds2}
\end{align} 
Define $\Reg_T^{(j)}$ to be the regret contribution from each arm $1 \leq j \leq K$ and let $M^*$ be the round where the optimal arm $j^*$is eliminated. Hence,
\begin{align*}
\E[\Reg_T] &= \E \bigg[ \sum_{j=1}^K \Reg_T^{(j)} \bigg] =\E \bigg[ \sum_{m=0}^{\infty}  \sum_{j=1}^K \Reg_T^{(j)} \one\{M^*=m\} \bigg] 
\\ & =  \E \bigg[\sum_{m=0}^{\infty}  \sum_{j : m_j <m} \Reg_T^{(j)} \one\{M^*=m\} + \sum_{j : m_j \geq m} \Reg_T^{(j)} \one \{M^*=m\}  \bigg] 
\\ &= \underbrace{\E \bigg[ \sum_{m=0}^{\infty}\sum_{j: m_j<m} \Reg_T^{(j)} \one \{M^*=m\}  \bigg]}_{\text{I.}} 
	+ \underbrace{ \E \bigg[ \sum_{m=0}^\infty \sum_{j: m_j \geq m} \Reg_T^{(j)}\one\{M^*=m \}\bigg] }_{\text{II.}}
\end{align*}

We will bound the regret in each of these cases in turn. First, however, we need the following results. 
\begin{lemma} \label{lem:probelim3}
For any suboptimal arm $j$, if $j^* \in \cA_{m_j}$, then the probability arm $j$ is \emph{not} eliminated by round $m_j$ is,
\[ \PP(M_j > m_j \text{ and } M^* \geq m_j) \leq \frac{24}{T\tilde \Delta_{m_j}^2} \]
\end{lemma}
\begin{proof}
The proof is exactly that of Lemma~\ref{lem:probelim} but using \cref{lem:eb2} to bound the probability of the confidence bounds on either arm $j$ or $j^*$ failing.
\end{proof}

Define the event $F_j(m) = \{ \bar{X}_{m,j^*}  < \bar{X}_{m,j} - \tilde \Delta_m \} \cap\{j,j^* \in \cA_m\}$ to be the event that arm $j^*$ is eliminated by arm $j$ in phase $m$. The probability of this occurring is bounded in the following lemma.
\begin{lemma} \label{lem:probelimjstar3}
The probability that the optimal arm $j^*$ is eliminated in round $m<\infty$ by the suboptimal arm $j$ is bounded by
\[ \PP( F_j(m) ) \leq \frac{24}{T \tilde \Delta_m^2} \]
\end{lemma}
\begin{proof}
Again, the proof follows from Lemma~\ref{lem:probelimjstar} but using \cref{lem:eb2} to bound the probability of the confidence bounds failing.
\end{proof}

\noindent We now return to bounding the expected regret in each of the two cases.
\paragraph{Bounding Term I.}
To bound the first term, we consider the cases where arm $j$ is eliminated in or before the correct round ($M_j \leq m_j)$ and where arm $j$ is eliminated late ($M_j >m_j$). Then,
\begin{align*}
\E \bigg[ \sum_{m=0}^\infty \sum_{j: m_j<m} \Reg_T^{(j)} \one\{M^*=m\} \bigg] &= \E \bigg[ \sum_{j=1}^K \Reg_T^{(j)} \one\{M^* \geq m_j \} \bigg]  
\\ & \hspace{-60pt}  =  \E \bigg[ \sum_{j=1}^K \Reg_T^{(j)} \one\{M^* \geq m_j \} \one\{M_j \leq m_j\}\bigg] +  \E \bigg[ \sum_{j=1}^K \Reg_T^{(j)} \one\{M^* \geq m_j \} \one\{M_j >m_j\} \bigg] 
\\ & \hspace{-60pt}  \leq \sum_{j=1}^K \E[\Reg_T^{(j)} \one\{ M_j \leq m_j\} ] + \sum_{j=1}^K \E[ T \Delta_j \one\{ M^* \geq m_j , M_j > m_j\}] 
\\ & \hspace{-60pt}  \leq \sum_{j=1}^K 2\Delta_j n_{m_j,j} + \sum_{j=1}^K T\Delta_j \PP(M_j >m_j \text{ and } M^* \geq m_j)
\\ & \hspace{-60pt}  \leq \sum_{j=1}^K 2\Delta_jn_{m_j,j} + \sum_{j=1}^K T\Delta_j \frac{24}{T\tilde \Delta_{m_j}^2}
\\ & \hspace{-60pt} \leq \sum_{j=1}^K \bigg( 2\Delta_jn_{m_j,j} + \frac{384}{\Delta_j} \bigg),
\end{align*}
where the extra factor of 2 comes from the fact that each arm will be played $n_m$ times by the end of phase $m$ to get the data for the estimated mean, then in the worst case, arm $j$ is chosen as the arm to be played in the bridge period of each phase that it is active, and thus is played another $n_m$ times.

\paragraph{Bounding Term II}
For the second term, we use the results from Theorem~\ref{thm:reggenNew}, but using Lemma~\ref{lem:probelim3} to bound the probability a suboptimal arm is eliminated in a later round and Lemma~\ref{lem:probelimjstar3} to bound the probability $j^*$ is eliminated by a suboptimal arm. Hence, 
\[  \E \bigg[ \sum_{m=0}^\infty \sum_{j: m_j \geq m} \Reg_T^{(j)}\one\{M^*=m \}\bigg] \leq  \sum_{j=1}^K \frac{1536}{\Delta_j}. \]

\noindent Combining the regret from terms I and II gives,
 \begin{align*}
  \E [\Reg_T] &\leq \sum_{j=1}^K \bigg( \frac{1920}{\Delta_j} + 2\Delta_j n_{m_j,j} \bigg)
\end{align*}
Hence, all that remains is to bound $n_m$ in terms of $\Delta_j, T$ and $d$.
 Using $L_{m,T} = \log(T\tilde \Delta_m^2)$, we have that,
\begin{align*}
n_{m_j,j} & =  \max\bigg\{ m_j \tilde d_{m_j}, \bigg \lceil \frac{1}{\tilde \Delta_m^2} \bigg( \sqrt{2\log(T\tilde \Delta_m)} + \sqrt{2 \log(T\tilde \Delta_m) +\frac{8}{3}\tilde \Delta_m \log(T\tilde \Delta_m) + 4\tilde \Delta_m \E[\tau]}\bigg)^2 \bigg \rceil \bigg\}
	\\ & \leq \max \bigg\{ m_j \tilde d_{m_j},  \left \lceil \frac{1}{\tilde \Delta_{m_j}^2} \left(  8L_{m_j,T} + \frac{16}{3} \tilde \Delta_{m_j} L_{m_j,T} + 8 \tilde \Delta_{m_j} \E[\tau] \right) \right \rceil \bigg \}
	\\ & \leq \max \bigg\{ m_j \tilde d_{m_j}, 1 + \frac{8L_{m_j,T}}{\tilde \Delta_{m_j}^2} +\frac{8L_{m_j,T}}{3 \tilde \Delta_{m_j}} + \frac{8\E[\tau]}{\tilde \Delta_{m_j}} \bigg \}
	\\ & \leq \max \bigg\{ m_j \tilde d_{m_j}, 1 + \frac{128 L_{m_j,T}}{\Delta_j^2} +\frac{32L_{m_j,T}}{\Delta_j} + \frac{32\E[\tau]}{\Delta_j}. \bigg\}
\end{align*}
where we have used $(a+b)^2 \leq 2(a^2 + b^2)$ for $a,b \geq 0$.

Hence, using the definition of $\tilde d_m = \min\{d, \frac{\eqref{eqn:ngen}}{m} \}$ and the results from \cref{thm:reggenNew}, the total expected regret from \ODAAF with bounded delays can be bounded by,
\begin{align}
 \E[\Reg_t] &\leq \sum_{\substack{j=1;j \neq j^*}}^K \max \bigg \{ \min \{d, \eqref{eqn:regfull}\}, \bigg( \frac{256 \log(T\Delta_j^2)}{\Delta_j} + 64 \E[\tau] +  \frac{1920}{\Delta_j} + 64 \log(T\Delta_j^2) + 2\Delta_j \bigg) \bigg\}. \label{eqn:regbounded}
 \\ & \leq \sum_{\substack{j=1;j \neq j^*}}^K \bigg(  \frac{256 \log(T \Delta_j^2)}{\Delta_j} + 64 \E[\tau] +  \frac{1920}{\Delta_j} + 64 \log(T\Delta_j^2) + 2\Delta_j \nonumber
 	\\ & \hspace{80pt} + \min \bigg\{d, \frac{128\log(T\Delta_j^2)}{\Delta_j}  + 96\log(4/\Delta_j)\E[\tau] \bigg \} \bigg) \nonumber
 \end{align}
\end{proof}

Note that the constants in these regret bounds can be improved by only requiring the confidence bounds in phase $m$ to hold with probability $\frac{1}{T\tilde \Delta_m}$ rather than $\frac{1}{T\tilde \Delta_m^2}$. This comes at a cost of increasing the logarithmic term to $\log(T \Delta_j)$.
We now prove the problem independent regret bound,
\corregboundedNew*
\begin{proof}
We consider the maximal value each part of the regret in \eqref{eqn:regbounded} can take. From \cref{cor:reggen}, the first term is bounded by
\[ O(\min\{ Kd, \sqrt{KT\log K} + K \log (T) \E[\tau] \}). \]
For the first term, we again set $\lambda = \sqrt{\frac{K \log(K) e^2}{T}}$. Then, as in corollary \cref{cor:reggen}, for constants $C_1,C_2>0$, we bound the regret contribution by
\[ \sum_{j: \Delta_j \leq \lambda} \E[\Reg_t^{(j)}] + \sum_{j: \Delta_j >\lambda} \E[\Reg_T^{(j)}] \leq \frac{KC_1 \log(T\lambda^2)}{\lambda} + C_2 K \E[\tau] + T\lambda. \]
Then, substituting in for $\lambda$ implies that the second term of \eqref{eqn:regbounded} is $O(\sqrt{KT\log K} + K\E[\tau])$.

For $d \leq \sqrt{\frac{T \log K}{K}} + \E[ \tau]$, $\min\{ Kd, \sqrt{KT\log K} + K \log T \E[\tau] \} \leq \sqrt{KT \log K} + K\E[\tau]$. Hence the bound in \eqref{eqn:regbounded} gives
\[ \E[\Reg_T] \leq O(\sqrt{KT \log K} + K\E[\tau] + \sqrt{KT \log K} + K\E[\tau]) = O(\sqrt{KT \log K} + K\E[\tau]). \]
\end{proof}


\section{Results for Delay with Known and Bounded Variance and Expectation} \label{app:var}

\subsection{High Probability Bounds} \label{app:cbvar}
\begin{lemma}
\label{lem:eb3}
Under Assumption \ref{assum:1} of known expected value and \ref{assum:3} of known (bound on) the expectation and variance of the delay, and choice of $n_m$ given in \eqref{eq:nvar}, the estimates $\bar{X}_{m,j}$ obtained by Algorithm \ref{alg:skeleton} satisfy the following: For any arm $j$ and phase $m$, with probability at least $1-\frac{12}{T\tilde{\Delta}_m^2}$, either $j \notin \cA_m$ or
$$\bar{X}_{m,j} -\mu_j \le \tilde{\Delta}_{m}/2.$$
\end{lemma}
\begin{proof}
Let 
\begin{align}
w_m = \frac{4\log(T\tilde \Delta_m^2)}{3n_m} + \sqrt{\frac{2\log(T \tilde \Delta_m^2)}{n_m}} + \frac{2\E[\tau] + 4 \Var(\tau)}{n_m}. \label{eqn:wmvar}
\end{align}
We show that with probability greater than $1-\frac{12}{T\tilde \Delta_m^2}$, $j \notin \cA_m$ or $\frac{1}{n_m}\sum_{t \in T_j(m)} (X_t - \mu_j) \leq  w_m$. 

For any arm $j$, phase $i$ and time $t$, define,
\begin{align}
A_{i,t} = R_{t,J_t} \one\{ \tau_{t,J_t} + t \geq S_i\},
\quad B_{i,t} = R_{t,J_t} \one \{ \tau_{t,J_t} + t \geq S_{i,j} \}, 
\quad  C_{i,t} = R_{t,J_t} \one \{ \tau_{t,J_t} +t > U_{i,j} \} \label{eqn:abcdefvar}
\end{align}
as in \eqref{eqn:abcdef} and
\begin{align*}
Q_t &= \sum_{i=1}^m (A_{i,t}  \one\{S_{i-1,j} \leq t \leq S_i - \nu_{i-1}-1\} + B_{i,t}  \one \{S_i - \nu_{i-1} \leq t \leq S_{i,j}-1\} ),
\\ P_t &= \sum_{i=1}^m C_{i,t}  \one\{S_{i,j} \leq t \leq U_{i,j}\},
\end{align*}
where $\nu_i = n_i - n_{i-1}$ is the number of times each active arm is played in phase $i \geq 1$ (assume $n_0=0$). 
Recall from the proof of Theorem~\ref{thm:reggenNew}, $\one_i \{H \} := \one \{H \cap \{j \in \cA_i \}\} \leq \one\{H\}$ and for all arms $j$ and phases $i$, $\one_i \{ \tau_{t,J_t} + t \geq S_{i,j} \} =  \one \{ \tau_{t,J_t} + t \geq S_{i,j} \}$ and $ \one_i \{ \tau_{t,J_t} +t > U_{i,j} \} = \one \{ \tau_{t,J_t} +t > U_{i,j} \}$.

Then, using the convention $S_0=S_{0,j}=0$ for all arms $j$, we use the decomposition,
\begin{align}
\sum_{i=1}^m \sum_{t=S_{i,j}}^{U_{i,j}} (X_t - \mu_j) &\leq \sum_{i=1}^m \bigg( \sum_{t=S_{i-1,j}}^{S_{i,j}-1} R_{t,J_t} \one_i \{ \tau_{t,J_t} + t \geq S_{i,j}\} + \sum_{t=S_{i,j}}^{U_{i,j}} (R_{t,J_t} - \mu_j) - \sum_{t=S_{i,j}}^{U_{i,j}} R_{t,J_t} \one_i\{ \tau_{t,J_t} + t > U_{i,j}\} \bigg) \nonumber
\\ &\leq \sum_{i=1}^m \bigg( \sum_{t=S_{i-1,j}}^{S_i - \nu_{i-1}-1} R_{t,J_t} \one \{ \tau_{t,J_t} + t \geq S_i\} + \sum_{t=S_i - \nu_{i-1} }^{S_{i,j}-1} R_{t,J_t} \one \{ \tau_{t,J_t} + t \geq S_{i,j}\}  \nonumber
	\\ & \hspace{50pt} + \sum_{t=S_{i,j}}^{U_{i,j}} (R_{t,J_t} - \mu_j) - \sum_{t=S_{i,j}}^{U_{i,j}} R_{t,J_t} \one\{ \tau_{t,J_t} + t > U_{i,j}\} \bigg) \nonumber
\\ & = \sum_{i=1}^m \bigg( \sum_{t=S_{i-1,j}}^{S_i - \nu_{i-1}-1} A_{i,t} +  \sum_{t=S_i - \nu_{i-1} }^{S_{i,j}-1} B_{i,t} + \sum_{t=S_{i,j}}^{U_{i,j}} (R_{t,J_t} - \mu_j) - \sum_{t=S_{i,j}}^{U_{i,j}} C_{i,t} \bigg) \nonumber
\\ & = \sum_{i=1}^m \sum_{t=S_{i,j}}^{U_{i,j}}(R_{t,J_t} - \mu_j) + \sum_{t=1}^{S_{m,j}} Q_t - \sum_{t=1}^{U_{m,j}} P_t \nonumber
\\ & = \underbrace{\sum_{i=1}^m \sum_{t=S_{i,j}}^{U_{i,j}}(R_{t,J_t} - \mu_j)}_{\text{Term I.}} 
	+ \underbrace{\sum_{t=1}^{S_{m,j}} (Q_t - \E[Q_t| \cG_{t-1}])}_{\text{Term II.}}
	 + \underbrace{\sum_{t=1}^{U_{m,j}}( \E[P_t|\cG_{t-1}]- P_t)}_{\text{Term III.}} \label{eqn:decompvar}
	\\ & \hspace{50pt}  + \underbrace{\sum_{t=1}^{S_{m,j}} \E[Q_t|\cG_{t-1}] - \sum_{t=1}^{U_{m,j}} \E[P_t|\cG_{t-1}],}_{\text{Term IV.}} \nonumber
\end{align}
Recall that the filtration $\{\cG_s\}_{s=0}^\infty$ is defined by $
\cG_0=\{\Omega, \emptyset\}$ and
\[\cG_t = \sigma(X_1, \dots, X_t, J_1, \dots, J_t, \tau_{1,J_1}, \dots, \tau_{t,J_t}, R_{1,J_1}, \dots R_{t,J_t}).\]
Furthermore, we have defined $S_{i,j}=\infty$ if arm $j$ is eliminated before phase $i$ and $S_i = \infty$ if the algorithm stops before reaching phase $i$. 

\paragraph{Outline of proof:}
We will bound each term of the above decomposition in turn. We first show in Lemma~\ref{lem:getvar} how the bounded second moment information can be incorporated using Chebychev's inequality. In Lemma~\ref{lem:qmart}, we show that $Z_t = Q_t - \E[Q_t|\cG_{t-1}]$ is a martingale difference sequence and bound its variance in Lemma~\ref{lem:varq} before using Freedman's inequality. Then in Lemma~\ref{lem:abcbounds}, we provide alternative (tighter) bounds on $A_{i,t},B_{i,t}, C_{i,t}$ which are used to bound term IV.. All these results are then combined to give a high probability bound on the entire decomposition.

\begin{lemma} \label{lem:getvar}
For any $a > \lfloor \E[\tau]\rfloor +1$, $a \in \mathbb{N}$, 
\[ \sum_{l=a}^\infty \PP(\tau \geq l) \leq \frac{\Var(\tau)}{a-\lfloor \E[\tau] \rfloor -1}.\]
\end{lemma}
\begin{proof}
For any $b>a$, $b \in \mathbb{N}$,  and by denoting $\xi \doteq \lfloor \E(\tau) \rfloor$, 
\begin{align*}
\sum_{l=a}^{b}\PP(\tau \geq l) &=  \sum_{l=a}^{b}\PP(\tau - \xi \geq l -\xi)  =  \sum_{l=a-\xi}^{b-\xi}\PP(\tau-\xi \geq l) 
\\ &\leq  \sum_{l=a-\xi}^{b-\xi} \frac{\Var(\tau)}{l^2} \tag*{(\text{by Chebychev's inequality since $l+ \xi > \E[\tau]$ for $l \geq a - \xi$})}
\\ & \leq \Var(\tau) \sum_{l=a-\xi-1}^{b-\xi-1} \frac{1}{l(l+1)}
\\ &= \Var(\tau) \sum_{l=a-\xi-1}^{b-\xi-1} \bigg( \frac{1}{l} - \frac{1}{l+1} \bigg)
\\ &= \Var(\tau) \bigg(\frac{1}{a-\xi-1} - \frac{1}{b-\xi} \bigg).
\end{align*}
Hence, taking $b \to \infty$ gives
\[\sum_{l=a}^\infty \PP(\tau \geq l )\leq \Var(\tau) \frac{1}{a-\xi-1}. \]
\end{proof}

\begin{lemma} \label{lem:qmart}
Let $Y_s = \sum_{t=1}^s (Q_t - \E[Q_t|\cG_{t-1}])$ for all $s\geq 1$, and $Y_0=0$. Then $\{Y_s\}_{s=0}^\infty$ is a martingale with respect to the filtration $\{\cG_s\}_{s=0}^\infty$ with increments $Z_s = Y_s-Y_{s-1}=Q_s - \E[Q_s|\cG_{s-1}]$ satisfying $\E[Z_s|\cG_{s-1}]=0, |Z_s| \leq 1$ for all $s\geq 1$.
\end{lemma}
\begin{proof}
To show $\{Y_s\}_{s=0}^\infty$ is a martingale we need to show that $Y_s$ is $\cG_s$-measurable for all $s$ and $\E[Y_s|\cG_{s-1}]=Y_{s-1}$. 

\noindent \underline{Measurability:} 
We show that $A_{i,s} \one\{S_{i-1,j} \leq s \leq S_i- \nu_{i-1}\} + B_{i,s} \one\{S_i - \nu_{i-1} + 1 \leq s \leq S_{i,j} -1 \}$ is $\cG_s$-measurable for every $i \leq m$. This then suffices to show that $Y_s$ is $\cG_s$-measurable since each $Q_t$ is a sum of such terms and the filtration $\cG_s$ is non-decreasing in $s$.

First note that by definition of $\cG_s$, $\tau_{t,J_t}, R_{t,J_t}$ are all $\cG_s$-measurable for $t \leq s$. It is sufficient to show that $\one\{\tau_{s,J_s} + s \geq S_i, S_{i-1,j} \leq s \leq S_i- \nu_i \}  + \one\{\tau_{s,J_s} + s \geq S_{i,j}, S_i - \nu_{i-1} + 1 \leq s \leq S_{i,j} -1 \}$  is $\cG_s$-measurable since the product of measurable functions is measurable. 
The first summand is $\cG_s$ measurable since $\{S_{i-1,j} \leq s\} \in \cG_s$ and $\{S_i =s', S_{i-1,j} \leq s\} \in \cG_s$ for all $s'\in \mathbb{N} \cup \{\infty\}$. So the union $\bigcup_{s' \in \mathbb{N} \cup \{\infty\}} \{\tau_{s,J_s} + s \geq s', S_{i-1,j} \leq s \leq s' - \nu_i, S_i =s' \} = \{\tau_{s,J_s} + s \geq S_i, S_{i-1,j} \leq s \leq S_i- \nu_{i-1} \} $ is an element of $\cG_s$. The same argument works for the second summand since $\{S_{ij}=s',S_i- \nu_{i-1} \leq s \} \in \cG_s$ for all $s'\in\mathbb{N} \cup \{\infty \}$

\noindent \underline{Increments:} Hence, to show that $\{Y_s\}_{s=0}^\infty$ is a martingale with respect to the filtration $\{\cG_s\}_{s=0}^\infty$ it just remains to show that the increments conditional on the past are zero. For any $s\geq 1$, we have that 
\[ Z_s = Y_s - Y_{s-1} = \sum_{t=1}^s (Q_t - \E[Q_t|\cG_{t-1}]) - \sum_{t=1}^{s-1} (Q_t - \E[Q_t|\cG_{t-1}]) = Q_s - \E[Q_s|\cG_{s-1}]. \]
Then,
\[ \E[Z_s |\cG_{s-1}] = \E[Q_s - \E[Q_s|\cG_{s-1}] |\cG_{s-1}] = \E[Q_s|\cG_{s-1}] - \E[Q_s|\cG_{s-1}] = 0 \]
and so $\{Y_s\}_{s=0}^\infty$ is a martingale.

Lastly, since for any $t$ and $\omega \in \Omega$, there is only one $i$ where one of $\one\{S_{i-1,j} \leq t \leq S_i - \nu_{i-1}\}$ or $\one \{S_i - \nu_{i-1} +1 \leq t \leq S_{i,j}-1\}$ is equal to one (they cannot both be one), and by definition of $R_{t,J_t}$, $A_{i,t}, B_{i,t} \leq 1$, it follows that $|Z_s| = |Q_s-\E[Q_s|\cG_{s-1}]| \leq 1$ for all $s$.
\end{proof}

\begin{lemma} \label{lem:varq}
For any $t\geq 1$, let $Z_t=Q_t - \E[Q_t|\cG_{t-1}]$, then
\[ \sum_{t=1}^{S_{m,j} -1} \E[Z_t^2|\cG_{t-1}] \leq m\E[\tau] + m \Var(\tau). \] 
\end{lemma}
\begin{proof}
Let us denote $S' \doteq S_{m,j}-1$. Observe that 
\begin{align*}
&\sum_{t=1}^{S'} \E[Z_t^2|\cG_{t-1}] = \sum_{t=1}^{S'} \Var(Q_t|\cG_{t-1}) \leq \sum_{t=1}^{S'} \E[Q_t^2|\cG_{t-1}]
\\ & = \sum_{t=1}^{S'} \E\bigg[ \bigg(\sum_{i=1}^m (A_{i,t} \one\{S_{i-1,j} \leq t \leq S_i - \nu_{i-1}-1\} + B_{i,t} \one \{S_i - \nu_{i-1} \leq t \leq S_{i,j}-1\} ) \bigg)^2 \Big|\cG_{t-1} \bigg].
\end{align*}
Then all indicator terms $\one\{S_{i-1,j} \leq t \leq S_i - \nu_{i-1}-1\}$ and $\one \{S_i - \nu_{i-1} \leq t \leq S_{i,j}-1\}$ for all $i=1, \dots, m$ are $\cG_{t-1}$-measurable and only one can be non zero for any $\omega \in \Omega$. Hence, for any $\omega \in \Omega$, their product must be 0. Furthermore, for any $i,i' \leq m$, $i \not = i'$,  
\begin{align*}
A_{i,t}  \one\{S_{i-1,j} \leq t \leq S_i - \nu_{i-1}-1\} \times A_{i',t}  \one\{S_{i'-1,j} \leq t \leq S_{i'} - \nu_{i'-1}-1\} = 0&, \\
B_{i,t}  \one \{S_i - \nu_{i-1}  \leq t \leq S_{i,j}-1\} \times B_{i',t}   \one \{S_{i'} - \nu_{i'-1} \leq t \leq S_{i',j}-1\} = 0&, \\
A_{i,t} \one\{S_{i-1,j} \leq t \leq S_i - \nu_{i-1}-1\} \times B_{i',t} \one \{S_{i'} - \nu_{i'-1} \leq t \leq S_{i',j}-1\} = 0&, \\
A_{i',t} \one\{S_{i'-1,j} \leq t \leq S_{i'} - \nu_{i'-1}-1\} \times B_{i,t}  \times \one \{S_{i} - \nu_{i-1}  \leq t \leq S_{i,j}-1\} = 0&. \\
\end{align*}
Using the above we see that, 
\begin{align*}
\sum_{t=1}^{S'} \E[Z_t^2|\cG_{t-1}] & \leq \sum_{t=1}^{S'} \E\bigg[ \bigg(\sum_{i=1}^m (A_{i,t} \one\{S_{i-1,j} \leq t \leq S_i - \nu_{i-1}-1\} + B_{i,t} \one \{S_i - \nu_{i-1} \leq t \leq S_{i,j}-1\} ) \bigg)^2 \Big|\cG_{t-1} \bigg]
\\ & = \sum_{t=1}^{S'}  \E\bigg[ \sum_{i=1}^m (A_{i,t}^2 \one\{S_{i-1,j} \leq t \leq S_i - \nu_{i-1}-1\}^2 + B_{i,t}^2 \one \{S_i - \nu_{i-1} \leq t \leq S_{i,j}-1\}^2 )\Big|\cG_{t-1} \bigg]
\\ & = \sum_{i=2}^m \sum_{t=1}^{S'} \E[A_{i,t}^2  \one\{S_{i-1,j} \leq t \leq S_i - \nu_{i-1}-1\} |\cG_{t-1}]  \\
&\quad\quad\quad + \sum_{i=1}^m \sum_{t=1}^{S'} \E[B_{i,t}^2 \one \{S_i - \nu_i \leq t \leq S_{i,j}-1\} |\cG_{t-1}]
\\ 
\tag*{(using that both indicators are $\cG_{t-1}$-measurable)} \\
& \leq \sum_{i=2}^m \sum_{t=S_{i-1,j}}^{S_i-\nu_{i-1}-1} \E[A_{i,t}^2|\cG_{t-1}] + \sum_{i=1}^m \sum_{t=S_i-\nu_{i-1}}^{S_{i,j}-1} \E[B_{i,t}^2|\cG_{t-1}]. 
\end{align*}
Then, for any $i \geq 2$,
\begin{align*}
\sum_{t=S_{i-1,j}}^{S_i-\nu_{i-1}-1} \E[A_{i,t}^2|\cG_{t-1}] & = \sum_{t=S_{i-1,j}}^{S_i-\nu_{i-1}-1} \E[R_{t,J_t}^2 \one\{\tau_{t,J_t} + t \geq S_i\}|\cG_{t-1}]
\\ & \leq \sum_{t=S_{i-1,j}}^{S_i-\nu_{i-1}-1} \E[ \one\{\tau_{t,J_t} + t \geq S_i\}|\cG_{t-1}]
\\ & = \sum_{s=0}^\infty \sum_{s'=s}^\infty \one\{S_{i-1,j}=s, S_i=s'\} \sum_{t=s}^{s'-\nu_{i-1}-1} \E[\one \{\tau_{t,J_t} +t \geq S_i \}|\cG_{t-1}]
\\ & = \sum_{s=0}^\infty \sum_{s'=s}^\infty \sum_{t=s}^{s'-\nu_{i-1}-1} \E[\one \{S_{i-1,j}=s, S_i=s', \tau_{t,J_t} +t \geq S_i \}|\cG_{t-1}]
	\\ \tag*{(Since $\{S_i=s', S_{i-1,j} = s\} \in \cG_{t-1}$ for $t \geq s$)}
\\ & = \sum_{s=0}^\infty \sum_{s'=s}^\infty \sum_{t=s}^{s'-\nu_{i-1}-1} \E[\one \{S_{i-1,j}=s, S_i=s', \tau_{t,J_t} +t \geq s' \}|\cG_{t-1}]
\\ & = \sum_{s=0}^ \infty \sum_{s'=s}^\infty \one\{S_{i-1,j}=s, S_i=s'\} \sum_{t=s}^{s'-\nu_{i-1}-1} \PP( \tau_{t,J_t} +t \geq s')
	\\ \tag*{(Since $\{S_i=s', S_{i-1,j} =s \} \in \cG_{t-1}$ for $t\geq s$)}
\\ & \leq \sum_{s=0}^ \infty \sum_{s'=s}^\infty \one\{S_{i-1,j}=s, S_i=s'\} \sum_{l=\nu_{i-1}+1}^\infty \PP(\tau > l)
\\ & \leq \Var[\tau],
\end{align*}
by Lemma~\ref{lem:getvar} since $\nu_i \geq \lfloor \E[\tau] \rfloor +2$ for all $i$.
Likewise, for any $i \geq 2$, 
\begin{align*}
\sum_{t=S_i - \nu_{i-1}}^{S_{i,j}-1} \E[B_{i,t}^2|\cG_{t-1}] & = \sum_{t=S_i - \nu_{i-1}}^{S_{i,j}-1} \E[R_{t,J_t}^2 \one\{\tau_{t,J_t} + t \geq S_{i,j}\}|\cG_{t-1}]
\\ & \leq \sum_{t=S_i - \nu_{i-1}}^{S_{i,j}-1} \E[ \one\{\tau_{t,J_t} + t \geq S_{i,j}\}|\cG_{t-1}]
\\ & = \sum_{s=\nu_{i-1}+1}^\infty \sum_{s'=s}^\infty \one\{S_i=s, S_{i,j}=s'\} \sum_{t=s-\nu_{i-1}}^{s'-1} \E[\one \{\tau_{t,J_t} +t \geq S_{i,j} \}|\cG_{t-1}]
\\ & = \sum_{s=\nu_{i-1}+1}^\infty \sum_{s'=s}^\infty \sum_{t=s-\nu_{i-1}}^{s'-1} \E[\one \{S_i=s, S_{i,j}=s', \tau_{t,J_t} +t \geq s' \}|\cG_{t-1}]
	\\ \tag*{(Since $\{S_{i,j}=s', S_i = s\} \in \cG_{t-1}$  for $t \geq s- \nu_i -1$)}
\\ & = \sum_{s=\nu_{i-1}+1}^ \infty \sum_{s'=s}^\infty \one\{S_i=s, S_{i,j}=s'\} \sum_{t=s-\nu_{i-1}}^{s'-1} \PP( \tau_{t,J_t} +t \geq s')
\\ & \leq \sum_{s=\nu_{i-1}+1}^ \infty \sum_{s'=s}^\infty \one\{S_i=s, S_{i,j}=s'\} \sum_{l=0}^\infty \PP(\tau > l)
\\ & \leq \E[\tau]  
\end{align*}
and for $i=1$ the derivation simplifies since we need to some over $1$ to $S_{1,j}-1$ only. Combining all terms gives the result.
\end{proof}

\begin{lemma} \label{lem:abcbounds}
For $A_{i,t}, B_{i,t}$ and $C_{i,t}$ defined as in \eqref{eqn:abcdefvar}, let $\nu_i = n_i - n_{i-1}$ be the number of times each arm is played in phase $i$ and $j_i'$ be the arm played directly before arm $j$ in phase $i$. Then, it holds that, for any arm $j$ and phase $i \geq 1$,
\begin{enumerate}[(i)]
	\item $ \displaystyle \sum_{t=S_{i-1,j}}^{S_i - \nu_{i-1}-1} \E[A_{i,t}|\cG_{t-1}] \leq \sum_{l=\nu_{i-1}+1}^\infty \PP(\tau \geq l)$.
	\item $ \displaystyle \sum_{t=S_i- \nu_{i-1}}^{S_{i,j} -1} \E[B_{i,t}|\cG_{t-1}] \leq \sum_{l=\nu_{i-1}+1}^\infty \PP(\tau \geq l) + \mu_{j_i'} \sum_{l=0}^{\nu_{i-1}} \PP(\tau > l)$.
	\item $ \displaystyle \sum_{t=S_{i,j}}^{U_{i,j}} \E[C_{i,t}|\cG_{t-1}] = \mu_j \sum_{l=0}^{\nu_{i-1}} \PP(\tau > l)$.
\end{enumerate}
\end{lemma}
\begin{proof}
The proof is very similar to that of Lemma~\ref{lem:varq}. We prove each statement individually. 
\paragraph{Statement (i):} This is similar to the proof of Lemma~\ref{lem:varq},
\begin{align*}
\sum_{t=S_{i-1,j}}^{S_i - \nu_{i-1}-1} \E[A_{i,t}|\cG_{t-1}] & \leq \sum_{t=S_{i-1,j}}^{S_i-\nu_{i-1}-1} \E[ \one\{\tau_{t,J_t} + t \geq S_i\}|\cG_{t-1}]
\\ & = \sum_{s=0}^\infty \sum_{s'=s}^\infty \one\{S_{i-1,j}=s, S_i=s'\} \sum_{t=s}^{s'-\nu_{i-1}-1} \E[\one \{\tau_{t,J_t} +t \geq S_i \}|\cG_{t-1}]
	\\ \tag*{(Since $\{S_{i}=s', S_{i-1,j} = s\} \in \cG_{t-1}$ for $t \geq s$)}
& = \sum_{s=0}^\infty \sum_{s'=s}^\infty \sum_{t=s}^{s'-\nu_{i-1}-1} \E[\one \{S_{i-1,j}=s, S_i=s', \tau_{t,J_t} +t \geq s' \}|\cG_{t-1}]
\\ & = \sum_{s=0}^ \infty \sum_{s'=s}^\infty \one\{S_{i-1,j}=s, S_i=s'\} \sum_{t=s}^{s'-\nu_{i-1}-1} \PP( \tau_{t,J_t} +t \geq s')
\\ & \leq \sum_{s=0}^ \infty \sum_{s'=s}^\infty \one\{S_{i-1,j}=s, S_i=s'\} \sum_{l=\nu_{i-1}+1}^\infty \PP(\tau >l)
\\ & = \sum_{l=\nu_{i-1}+1}^\infty \PP(\tau >l).
\end{align*}

\paragraph{Statement (ii):} For statement (ii), we have that for $(i,j) \neq (1,1)$,
\begin{align*}
\sum_{t=S_i- \nu_{i-1}}^{S_{i,j} -1} \E[B_{i,t}|\cG_{t-1}] = \sum_{t=S_i- \nu_{i-1}}^{S_{i,j} -\nu_{i-1}-2} \E[B_{i,t}|\cG_{t-1}] + \sum_{t=S_{i,j} -\nu_{i-1} -1}^{S_{i,j} -1} \E[B_{i,t}|\cG_{t-1}].
\end{align*}
Then,  since$\{S_{i,j}=s'\} \cap \{S_i- \nu_{i-1} \leq t\} \in \cG_{t-1}$  so we can use the same technique as for statement (i) to bound the first term. 
For the second term, since we will be playing only arm $j_i'$ for $S_{i,j}-\nu_{i-1}-1, \dots, S_{i,j}-1$, so,
\begin{align*}
 \sum_{t=S_{i,j}-\nu_{i-1}-1}^{S_{i,j}-1} \E[B_{i,t}|\cG_{t-1}] & =  \sum_{t=S_{i,j}-\nu_{i-1}-1}^{S_{i,j}-1}  \E[R_{t,J_t} \one\{\tau_{t,J_t} + t \geq S_{i,j}\}|\cG_{t-1}]
\\ & =  \sum_{s=0}^\infty\one\{S_{i,j}=s\} \sum_{t=s-\nu_{i-1}-1}^{s-1} \E[R_{t,J_t} \one \{\tau_{t,J_t} +t \geq S_{i,j} \}|\cG_{t-1}] 
\\ & = \sum_{s=0}^\infty  \sum_{t=s-\nu_{i-1}-1}^{s-1} \E[R_{t,J_t} \one \{S_{i,j}=s, \tau_{t,J_t} +t \geq S_{i,j} \}|\cG_{t-1}]
	\\ \tag*{(Since $\{S_{i,j}=s', S_{i,j} - \nu_{i-1} \leq t\} \in \cG_{t-1}$ )}
\\ & = \sum_{s=0}^\infty \sum_{t=s- \nu_{i-1}-1}^{s-1} \E[R_{t,J_t} \one \{S_{i,j}=s, \tau_{t,J_t} +t \geq s \}|\cG_{t-1}]
\\ & = \sum_{s=0}^ \infty \one\{S_{i,j}=s\} \sum_{t=s-\nu_{i-1}-1}^{s-1} \mu_{j_i'} \PP( \tau_{t,J_t} +t \geq s)
	\\ \tag*{(\text{Since $\{S_{i,j}=s\}  \in \cG_{t-1}$ for $t \geq s-\nu_{i-1}-1$ and given $\cG_{t-1}$, $R_{t,J_t}$ and $\tau_{t,J_t}$ are independent})}
\\ & =  \sum_{s=0}^ \infty \one\{S_{i,j}=s\} \mu_{j_i'} \sum_{l=0}^{\nu_{i-1}}\PP(\tau > l)
\\ & = \mu_{j_i'} \sum_{l=0}^{\nu_{i-1}}\PP(\tau > l).
\end{align*}
Then, for $(i,j)=(1,1)$, the amount seeping in will be 0, so using $\nu_0=0, \mu_{1_1}'=0$, the result trivially holds.
Hence,
\[ \sum_{t=S_i- \nu_{i-1}}^{S_{i,j} -1} \E[B_{i,t}|\cG_{t-1}] \leq \sum_{l=\nu_{i-1}+1}^\infty \PP(\tau \geq l) + \mu_{j_i'} \sum_{l=0}^{\nu_{i-1}} \PP(\tau > l). \]

\paragraph{Statement (iii):} This is the same as in Lemma~\ref{lem:abcboundsgen}.
\end{proof}

We now bound each term of the decomposition in \eqref{eqn:decompvar}.
\paragraph{Bounding Term I.:}
For Term I., we can again use Lemma~\ref{lem:az} as in the proof of \cref{lem:eb1} to get that with probability greater than $1-\frac{1}{T\tilde \Delta_m^2}$,
 \[ \sum_{i=1}^m \sum_{t=S_{i,j}}^{U_{i,j}} (R_{t,J_t} - \mu_j) \leq \sqrt{\frac{n_m \log(T\tilde \Delta_m^2)}{2}}. \]
 
 \paragraph{Bounding Term II.:}
 For Term II., we will use Freedmans inequality (Theorem~\ref{thm:freedman}). From Lemma~\ref{lem:qmart}, $\{Y_s\}_{s=0}^\infty$ with $Y_s = \sum_{t=1}^s (Q_t- \E[Q_t|\cG_{t-1}])$ is a martingale with respect to $\{ \cG_s\}_{s=0}^\infty$ with increments $\{Z_s\}_{s=0}^\infty$ satisfying $\E[Z_s|\cG_{s-1}]=0$ and $Z_s \leq 1$ for all $s$. Further, by Lemma~\ref{lem:varq}, $\sum_{t=1}^s \E[Z_t^2 |\cG_{t-1}] \leq m \E[\tau] + m \Var(\tau) \leq \frac{4 \times 2^m}{8}(\E[\tau] +\Var(\tau)) \leq n_m/8$ with probability 1. Hence we can apply Freedman's inequality to get that with probability greater than $1- \frac{1}{T\tilde \Delta_m^2}$,
 \[  \sum_{t=1}^{S_{m,j}} (Q_t - \E[Q_t|\cG_{t-1}]) 
 = \sum_{s=1}^\infty \one\{S_{m,j}=s\} \times Y_s
 \leq \frac{2}{3} \log(T\tilde \Delta_m^2) + \sqrt{\frac{1}{8} n_m \log(T\tilde \Delta_m^2)}, \]
 using that Freedman's inequality applies simultaneously to all $s \geq 1$.
 
 \paragraph{Bounding Term III.:}
 For Term III., we again use Freedman's inequality (Theorem~\ref{thm:freedman}), using Lemma~\ref{lem:pmart} to show that $\{Y'_s\}_{s=0}^\infty$ with $Y'_s = \sum_{t=1}^s (\E[P_t|\cG_{t-1}] - P_t)$ is a martingale with respect to $\{ \cG_s\}_{s=0}^\infty$ with increments $\{Z'_s\}_{s=0}^\infty$ satisfying $\E[Z'_s|\cG_{s-1}]=0$ and $Z'_s \leq 1$ for all $s$. Further, by Lemma~\ref{lem:varp}, $\sum_{t=1}^s \E[Z_t^2 |\cG_{t-1}] \leq m \E[\tau] \leq n_m/8$ with probability 1. Hence, with probability greater than $1- \frac{1}{T\tilde \Delta_m^2}$,
 \[  \sum_{t=1}^{U_{m,j}} ( \E[P_t|\cG_{t-1}] - P_t) = \sum_{s=1}^\infty \one\{U_{m,j} = s\} \times Y'_s \leq \frac{2}{3} \log(T\tilde \Delta_m^2) + \sqrt{\frac{1}{8} n_m \log(T\tilde \Delta_m^2)}. \]
 
 \paragraph{Bounding Term IV.:}
 To begin with, observe that,
 \begin{align}
 &\sum_{t=1}^{S_{m,j}} \E[Q_t|\cG_{t-1}] - \sum_{t=1}^{U_{m,j}} \E[P_t|\cG_{t-1}] \nonumber
 \\ & = \sum_{t=1}^{S_{m,j}} \E \bigg[ \sum_{i=1}^m (A_{i,t} \times \one\{S_{i-1,j} \leq t \leq S_i - \nu_{i-1}-1\} + B_{i,t} \times \one \{S_i - \nu_{i-1} \leq t \leq S_{i,j}-1\}) \bigg| \cG_{t-1} \bigg] \nonumber
 	\\ & \hspace{50pt} - \sum_{t=1}^{U_{m,j}} \E \bigg[  \sum_{i=1}^m C_{i,t} \times \one\{S_{i,j} \leq t \leq U_{i,j}\} \bigg| \cG_{t-1}\bigg] \nonumber
\\ & = \sum_{i=1}^m\sum_{t=1}^{S_{m,j}} \E[ A_{i,t} \times \one\{S_{i-1,j} \leq t \leq S_i - \nu_{i-1}-1\}|\cG_{t-1}]  \nonumber \\
	& \hspace{50pt} + \sum_{i=1}^m\sum_{t=1}^{S_{m,j}} \E[B_{i,t} \times \one \{S_i - \nu_{i-1}  \leq t \leq S_{i,j}-1\}|\cG_{t-1}]  \nonumber
	\\ & \hspace{50pt} - \sum_{i=1}^m \sum_{t=1}^{U_{m,j}} \E[C_{i,t} \times \one\{S_{i,j} \leq t \leq U_{i,j}\} |\cG_{t-1}] \nonumber
\\ 
& = \sum_{i=1}^m \bigg( \sum_{t=S_{i-1,j}}^{S_i - \nu_{i-1}-1} \E[A_{i,t}|\cG_{t-1}] + \sum_{t=S_i - \nu_{i-1}}^{S_{i,j}-1} \E[B_{i,t}|\cG_{t-1}] - \sum_{t=S_{i,j}}^{U_{i,j}} \E[C_{i,t}|\cG_{t-1}] \bigg) \nonumber
\\ 
\tag*{(using that the indicators are $\cG_{t-1}$-measurable)}\\ \nonumber
&\leq \sum_{i=1}^m \bigg( \sum_{l=\nu_{i-1}+1}^\infty \PP(\tau \geq l) + \mu_{j_i'} \sum_{l=0}^{\nu_{i-1}} \PP(\tau > l) - \mu_j \sum_{l=0}^{\nu_{i}} \PP(\tau > l) \bigg), \nonumber
\\ &\leq \sum_{i=1}^m \bigg( \frac{2\Var(\tau)}{\nu_{i-1} - \E[\tau]} + (\mu_{j_i'}-\mu_j) \sum_{l=0}^{\nu_{i}} \PP(\tau > l) \bigg), \nonumber
\\ &\leq \sum_{i=1}^m \bigg( \frac{2\Var(\tau)}{2^{i-1}} + (\mu_{j_i'}-\mu_j) \sum_{l=0}^{\nu_{i}} \PP(\tau > l) \bigg), \label{eqn:remark}
 \end{align}
by Lemma~\ref{lem:abcbounds} and Lemma~\ref{lem:getvar} where we have used the fact that since $n_m \leq T$, the maximal number of rounds of the algorithm is $\frac{1}{2} \log_2(T/4)$ and for $m \leq \frac{1}{2} \log_2(T/4)$, $\frac{\log(T\tilde \Delta_m^2)}{\tilde \Delta_m^2} \geq \frac{2 \log(T \tilde \Delta_{m-1}^2)}{\tilde \Delta_{m-1}^2}$ so $n_m \geq 2n_{m-1}$ and $\nu_m \geq n_{m-1}$. Then for $\E[\tau] \geq 1$, $\nu_{i-1} - \E[\tau] \geq 2/\tilde \Delta_{i-1} \E[\tau] - \E[\tau] \geq (2\times 2^{i-1}-1) \E[\tau] \geq 2^{i-1} \E[\tau] \geq 2^{i-1}$ and for $\E[\tau] \leq 1$, $\nu_{i-1}- \E[\tau] \geq \nu_{i-1}-1 \geq 2\log(4)/\tilde \Delta_{i-1}-1 \geq 2^{i-1}$ so $\nu_{i-1} - \E[\tau] \geq 2^{i-1}$. 
Then, the probability that either arm $j'_i$ or $j$ is active in phase $i$ when it should have been eliminated in or before phase $i-1$ is less than $2p_{i-1}$, where $p_i$ is the probability that the confidence bounds for one arm holds in phase $i$ and $p_0=0$. 
If neither arm should have been eliminated by phase $i$, this means that their mean rewards are within $\tilde \Delta_{i-1}$ of each other. Hence, with probability greater than $1-2p_{i-1}$,
 \[   \mu_{j_i'} \sum_{l=0}^{\nu_{i}} \PP(\tau > l) - \mu_j \sum_{l=0}^{\nu_{i}} \PP(\tau > l)  \leq \tilde \Delta_{i-1} \sum_{l=0}^{\nu_{i}} \PP(\tau > l)  \leq \tilde \Delta_{i-1} \E[\tau]. \]
Then, summing over all phases gives that with probability greater than $1- 2 \sum_{i=0}^{m-1} p_i $,
\begin{align*}
\sum_{t=1}^{S_{m,j}} \E[Q_t|\cG_{t-1}] - \sum_{t=1}^{U_{m,j}} \E[P_t|\cG_{t-1}]  &\leq 2\Var(\tau) \sum_{i=1}^m \frac{1}{2^{i-1}} + \E[\tau] \sum_{i=1}^m \tilde \Delta_{i- 1} = (2\Var(\tau) +\E[\tau])  \sum_{i=0}^{m-1} \frac{1}{2^i} 
\\ &\leq  4\Var(\tau) +2\E[\tau].
\end{align*}

\paragraph{Combining all terms:}
To get the final high probability bound, we sum the bounds for each term I.-IV.. Then, with probability greater than $1-(\frac{3}{T\tilde \Delta_m^2} + 2\sum_{i=1}^{m-1} p_i)$,  either $j \notin \cA_m$ or arm $j$ is played $n_m$ times by the end of phase $m$ and 
\begin{align*}
\frac{1}{n_m}\sum_{t \in T_j(m)} (X_t - \mu_j) &\leq \frac{4\log(T\tilde \Delta_m^2)}{3n_m} + \bigg(\frac{2}{\sqrt{8}} + \frac{1}{\sqrt{2}}\bigg)\sqrt{\frac{\log(T \tilde \Delta_m^2)}{n_m}} + \frac{2\E[\tau] + 4 \Var(\tau)}{n_m} 
\\ &\leq \frac{4\log(T\tilde \Delta_m^2)}{3n_m} + \sqrt{\frac{2\log(T \tilde \Delta_m^2)}{n_m}} + \frac{2\E[\tau] + 4 \Var(\tau)}{n_m} = w_m. 
\end{align*}
Using the fact that $p_0=0$ and substituting the other $p_i$'s using the same recursive relationship $p_i = \frac{3}{T\tilde \Delta_i^2} + 2\sum_{l=1}^{i-1} p_l$ as in the case for bounded delays (see the proof of \cref{lem:eb2}) gives, $p_m = \frac{12}{T \tilde \Delta_m^2}$ so the above bound holds with probability greater than $1-\frac{12}{T \tilde \Delta_m^2}$.

\paragraph{Defining ${\bf n_m}$:}
Setting
\begin{align}
n_m &=\bigg \lceil \frac{1}{\tilde \Delta_m^2} \bigg( \sqrt{2\log(T\tilde \Delta_m^2)} + \sqrt{2 \log(T\tilde \Delta_m^2) +\frac{8}{3}\tilde \Delta_m \log(T\tilde \Delta_m^2) + 4\tilde \Delta_m (\E[\tau] + 2\Var(\tau))}\bigg)^2 \bigg \rceil. \label{eqn:nvarlong}
\end{align}
ensures that $w_m \leq \frac{\tilde \Delta_m}{2}$ which concludes the proof.
\end{proof}

\paragraph{Remark:} Note that if $\E[\tau] \geq 1$, then the confidence bounds can be tightened by replacing \eqref{eqn:remark} with 
\[ \sum_{i=1}^m \bigg( \frac{2\Var(\tau)}{2^{i-1}\E[\tau]} + (\mu_{j_i'}-\mu_j) \sum_{l=0}^{\nu_{i}} \PP(\tau > l) \bigg) \]
This is obtained by noting that for $\E[\tau] \geq 1$. $\nu_{i-1} - \E[\tau] \geq 2/\tilde \Delta_{i-1} \E[\tau] - \E[\tau] \geq (2\times 2^{i-1}-1) \E[\tau] \geq 2^{i-1} \E[\tau]$. This leads to replacing the $\Var(\tau)$ term in the definition of $n_m$ by $\Var(\tau)/\E[\tau]$.


\subsection{Regret Bounds} \label{app:regvar}

\thmregvar*
\begin{proof}
The proof is very similar to that of Theorem~\ref{thm:reggenNew}, however, for clarity, we repeat the main arguments here. For any sub-optimal arm $j$, define $M_j$ to be the random variable representing the phase arm $j$ is eliminated in and note that if $M_j$ is finite, $j\in \cA_{M_j}$ but $j\not\in \cA_{M_j+1}$. Then let $m_j$ be the phase arm $j$ \emph{should} be eliminated in, that is $m_j= \min \{m| \tilde{\Delta}_m < \frac{\Delta_j}{2}\}$ and note that, from the new definition of $\tilde \Delta_m$ in our algorithm, we get the relations
\begin{align}
2^{m} = \frac{1}{\tilde\Delta_{m}}, \quad 2\tilde \Delta_{m_j} = \tilde\Delta_{m_j-1} \geq \frac{\Delta_j}{2} \quad \text{ and so, } \quad  \frac{\Delta_j}{4} \leq \tilde \Delta_{m_j} \leq \frac{\Delta_j}{2}. \label{eqn:deltabounds3}
\end{align} 
Define $\Reg_T^{(j)}$ to be the regret contribution from each arm $1 \leq j \leq K$ and let $M^*$ be the round where the optimal arm $j^*$is eliminated. Hence,
\begin{align*}
\E[\Reg_T] &= \E \bigg[ \sum_{j=1}^K \Reg_T^{(j)} \bigg] =\E \bigg[ \sum_{m=0}^{\infty}  \sum_{j=1}^K \Reg_T^{(j)} \one\{M^*=m\} \bigg] 
\\ & =  \E \bigg[\sum_{m=0}^{\infty}  \sum_{j : m_j <m} \Reg_T^{(j)} \one\{M^*=m\} + \sum_{j : m_j \geq m} \Reg_T^{(j)} \one \{M^*=m\}  \bigg] 
\\ &= \underbrace{\E \bigg[ \sum_{m=0}^{\infty}\sum_{j: m_j<m} \Reg_T^{(j)} \one \{M^*=m\}  \bigg]}_{\text{I.}} 
	+ \underbrace{ \E \bigg[ \sum_{m=0}^\infty \sum_{j: m_j \geq m} \Reg_T^{(j)}\one\{M^*=m \}\bigg] }_{\text{II.}}
\end{align*}

We will bound the regret in each of these cases in turn. First, however, we need the following results. 
\begin{lemma} \label{lem:probelim4}
For any suboptimal arm $j$, if $j^* \in \cA_{m_j}$, then the probability arm $j$ is \emph{not} eliminated by round $m_j$ is,
\[ \PP(M_j > m_j \text{ and } M^* \geq m_j) \leq \frac{24}{T\tilde \Delta_{m_j}^2} \]
\end{lemma}
\begin{proof}
The proof is exactly that of Lemma~\ref{lem:probelim} but using \cref{lem:eb3} to bound the probability of the confidence bounds on either arm $j$ or $j^*$ failing.
\end{proof}

Define the event $F_j(m) = \{ \bar{X}_{m,j^*}  < \bar{X}_{m,j} - \tilde \Delta_m \} \cap\{j,j^* \in \cA_m\}$ to be the event that arm $j^*$ is eliminated by arm $j$ in phase $m$. The probability of this event is bounded in the following lemma.
\begin{lemma} \label{lem:probelimjstar4}
The probability that the optimal arm $j^*$ is eliminated in round $m<\infty$ by the suboptimal arm $j$ is bounded by
\[ \PP( F_j(m) ) \leq \frac{24}{T \tilde \Delta_m^2} \]
\end{lemma}
\begin{proof}
Again, the proof follows from Lemma~\ref{lem:probelimjstar} but using \cref{lem:eb3} to bound the probability of the confidence bounds failing.
\end{proof}

\noindent We now return to bounding the expected regret in each of the two cases.
\paragraph{Bounding Term I.}
As in the proof of Theorem~\ref{thm:reggenNew}, to bound the first term, we consider the cases where arm $j$ is eliminated in or before the correct round ($M_j \leq m_j)$ and where arm $j$ is eliminated late ($M_j >m_j$). Then, using Lemma~\ref{lem:probelim3},
\begin{align*}
\E \bigg[ \sum_{m=0}^\infty \sum_{j: m_j<m} \Reg_T^{(j)} \one\{M^*=m\} \bigg] 
\leq \sum_{j=1}^K \bigg( 2\Delta_jn_{m_j,j} + \frac{384}{\Delta_j} \bigg)
\end{align*}


\paragraph{Bounding Term II}
For the second term, we again use the results from Theorem~\ref{thm:reggenNew}, but using Lemma~\ref{lem:probelim4} to bound the probability a suboptimal arm is eliminated in a later round and Lemma~\ref{lem:probelimjstar4} to bound the probability $j^*$ is eliminated by a suboptimal arm. Hence, 
\[  \E \bigg[ \sum_{m=0}^\infty \sum_{j: m_j \geq m} \Reg_T^{(j)}\one\{M^*=m \}\bigg] \leq  \sum_{j=1}^K \frac{1920}{\Delta_j}. \]

\noindent Combining the regret from terms I and II gives,
 \begin{align*}
  \E [\Reg_T] &\leq \sum_{j=1}^K \bigg( \frac{1920}{\Delta_j} + 2\Delta_j n_{m_j,j} \bigg)
\end{align*}
Hence, all that remains is to bound $n_m$ in terms of $\Delta_j, T$ and $\E[\tau], \Var(\tau)$.
 Using $L_{m,T} = \log(T\tilde \Delta_m^2)$, we have that,
\begin{align*}
n_{m_j,j} & = \bigg \lceil \frac{1}{\tilde \Delta_m^2} \bigg( \sqrt{2\log(T\tilde \Delta_m^2)} + \sqrt{2 \log(T\tilde \Delta_m^2) +\frac{8}{3}\tilde \Delta_m \log(T\tilde \Delta_m) + 4\tilde \Delta_m (\E[\tau] + 2\Var(\tau))}\bigg)^2 \bigg \rceil 
	\\ & \leq  \left \lceil \frac{1}{\tilde \Delta_{m_j}^2} \left( 8 L_{m_j,T} + \frac{16}{3} \tilde \Delta_{m_j} L_{m_j,T} + 8 \tilde \Delta_{m_j} \E[\tau]  +  16 \tilde \Delta_{m_j} \Var(\tau) \right) \right \rceil
	\\ & \leq 1 + \frac{8L_{m_j,T}}{\tilde \Delta_{m_j}^2} +\frac{16L_{m_j,T}}{3 \tilde \Delta_{m_j}} + \frac{8\E[\tau]}{\tilde \Delta_{m_j}} + \frac{16\Var(\tau)}{\tilde \Delta_{m_j}}
	\\ & \leq 1 + \frac{128 L_{m_j,T}}{\Delta_j^2} +\frac{32L_{m_j,T}}{\Delta_j} + \frac{32\E[\tau]}{\Delta_j} +\frac{64\Var(\tau)}{\Delta_j} .
\end{align*}
where we have used $(a+b)^2 \leq 2(a^2 + b^2)$ for $a,b \geq 0$.

Hence, the total expected regret from \ODAAF with bounded delays can be bounded by,
\[ \E[\Reg_t] \leq \sum_{j=1}^K \bigg( \frac{256 \log(T \Delta_j^2)}{\Delta_j} + 64 \E[\tau] + 128 \Var(\tau) + \frac{1920}{\Delta_j} + 64 \log(T) + 2\Delta_j \bigg). \]
\end{proof}

Note that again, these constants can be improved at a cost of increasing $\log(T\Delta_j^2)$ to $\log(T\Delta_j)$. We now prove the problem independent regret bound.

\corregvar*
\begin{proof}
Let $\lambda=\sqrt{\frac{K \log(K) e^2}{T}}$ and note that for $\Delta>\lambda$, $\log(T\Delta^2)/\Delta$ is decreasing in $\Delta$. Then, for constants $C_1,C_2>0$ we can bound the regret in the previous theorem by
\[ \E[\Reg_T] \leq \sum_{j: \Delta_j \leq \lambda} \E[\Reg_t^{(j)}] + \sum_{j: \Delta_j >\lambda} \E[\Reg_T^{(j)}] \leq \frac{KC_1 \log(T\lambda^2)}{\lambda} + KC_2(\E[\tau] + \Var(\tau)) + T\lambda. \]
substituting in the above value of $\lambda$ gives a worst case regret bound that scales with $O(\sqrt{KT\log(K)} + K(\E[\tau] + \Var(\tau)))$.
\end{proof}

\paragraph{Remark:} If $\E[\tau] \geq 1$, we can replace the $\Var(\tau)$ terms in the regret bounds with $\Var(\tau)/\E[\tau]$. This follows by using the alternative definition of $n_m$ suggested in the remark at the end of Section~\ref{app:cbvar}.

\section{Additional Experimental Results} \label{app:experiments}
\subsection{Increasing the Expected Delay}
\begin{figure}
	\centering
	\includegraphics[width=0.42\textwidth]{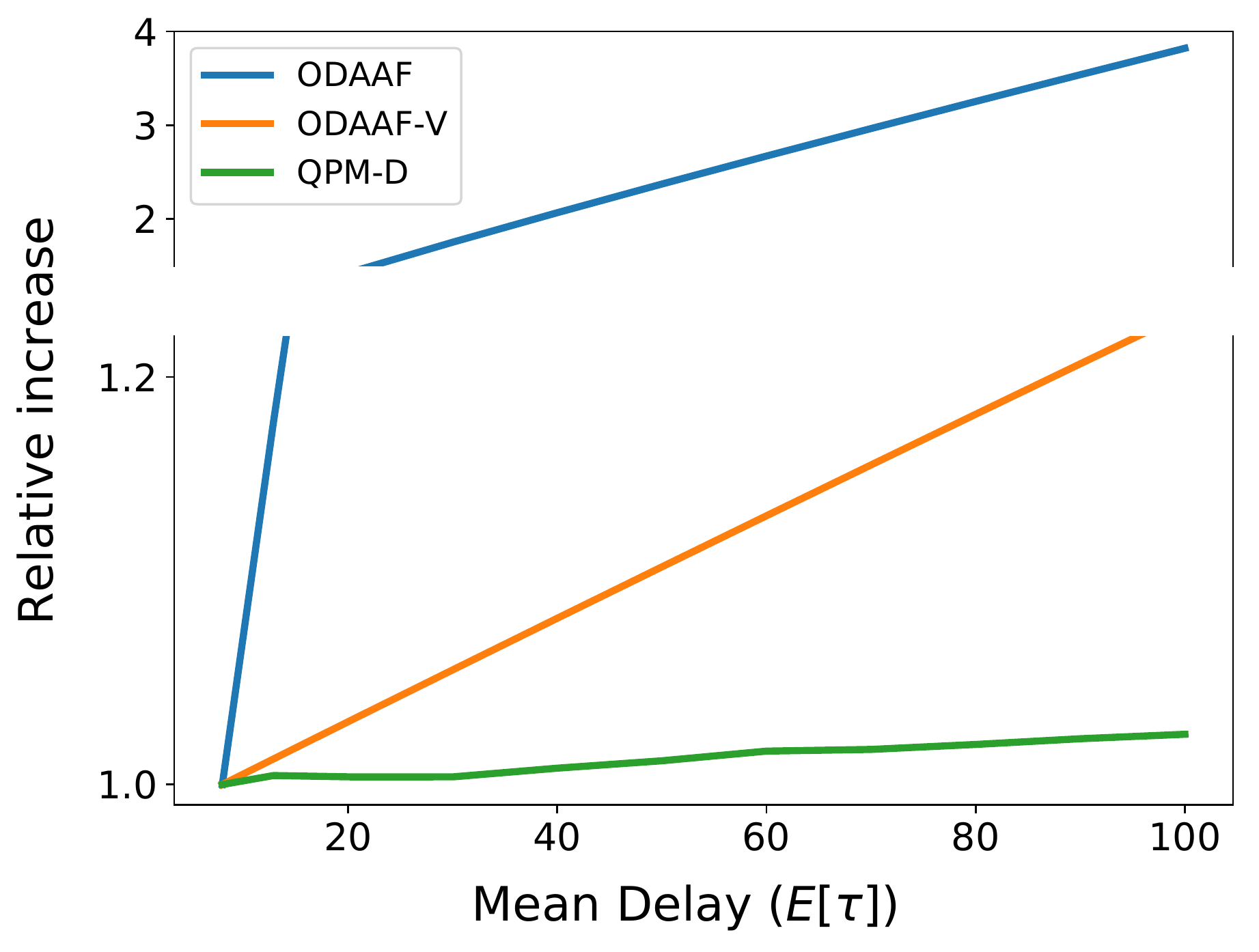}
        \caption{The relative increase in regret at horizon $T=250000$ for increasing mean delay when the delay is $\mathcal{N}_{+}$ with variance $100$.}
        \label{fig:multiplemeans}
\end{figure}

Here we investigate the effect of increasing the mean delay on both our algorithm and QPM-D \citep{joulani2013online} and demonstrate that the regret of both algorithms increases linearly with $\E[\tau]$, as indicated by our theoretical results. We use the same experimental set up as described in Section~\ref{sec:experiments}.
 In Figure~\ref{fig:multiplemeans}, we are interested in the impact of the mean delay  on the regret so we kept the delay distribution family the same, using a $\mathcal{N}_+(\mu,100)$ (Normal distribution with mean $\mu$, variance 100, truncated at 0) as the delay distribution. We then ran the algorithms for increasing mean delays and plotted the ratio of the regret at $T$ to the regret of the same algorithm when the delay distribution was $\mathcal{N}_+(0,100)$. In this case, the regret was averaged over 1000 replications for \ODAAF and \ODAAF-V, and 5000 for QPM-D (this was necessary since the variance of the regret of QPM-D was significant).
Here, it can be seen that increasing the mean delay causes the regret of all three algorithms to increase linearly. This is in accordance with the regret bounds which all include a linear factor of $\E[\tau]$ (since here $\log(T)$ is kept constant). It can also be seen that \ODAAF-V scales better with $\E[\tau]$ than \ODAAF (for constant variance). Particularly, at $\E[\tau]=100$, the relative increase in \ODAAF-V is only 1.2 whereas that of \ODAAF is 4 (QPM-D has the best relative increase of 1.05).

\subsection{Comparison with \citet{vernade2017stochastic}}

\begin{figure}[t]
    \centering
    \begin{subfigure}{0.42\textwidth}
        \centering
        \includegraphics[width=\textwidth]{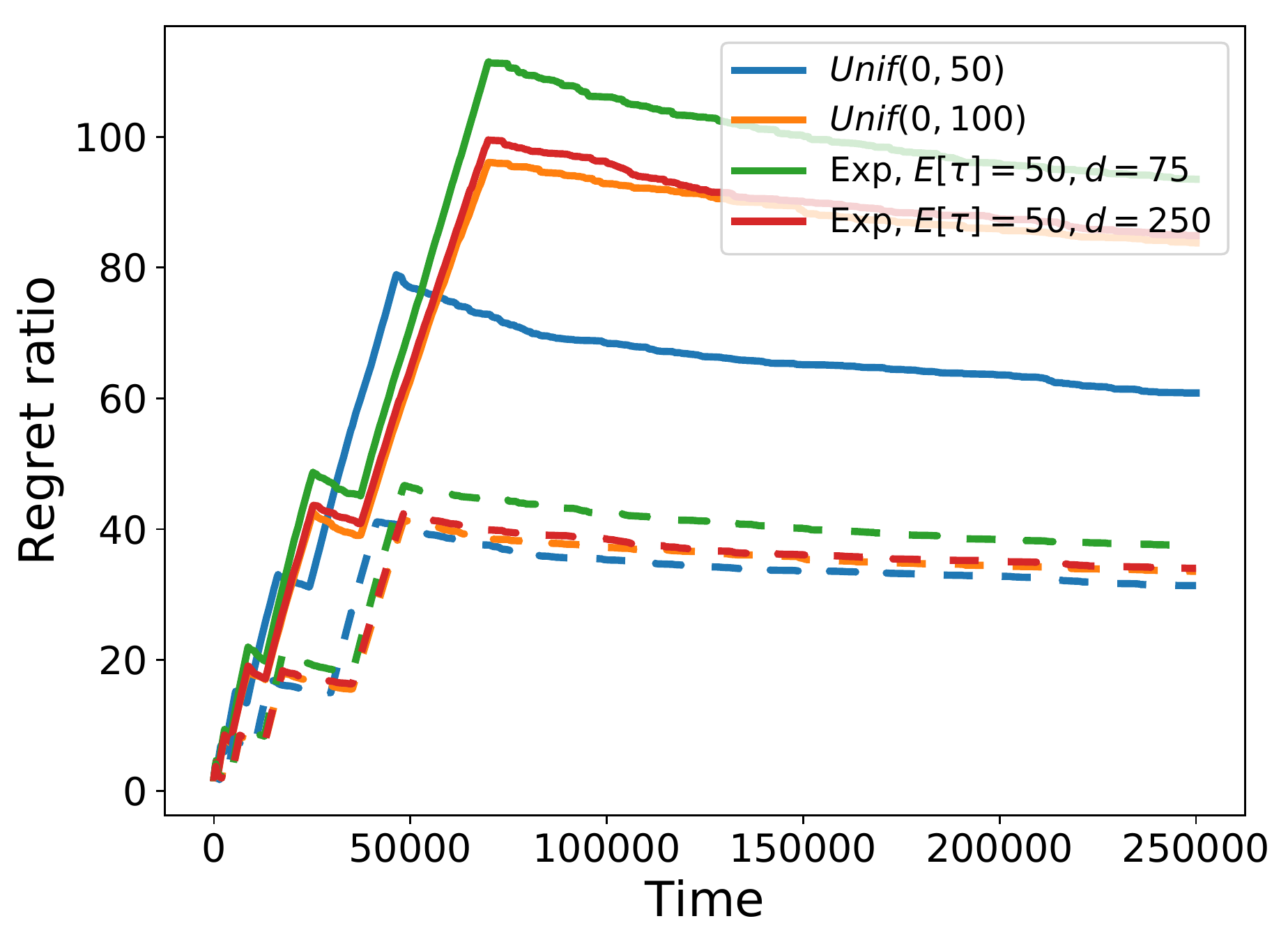}
        \caption{Bounded delays. Ratios of regret of \ODAAF (solid lines) and \ODAAF-B (dotted lines) to that of DUCB.}
        \label{fig:ducbbounded}
    \end{subfigure}
    \quad
    \begin{subfigure}{0.42\textwidth}
        \centering
        \includegraphics[width=\textwidth]{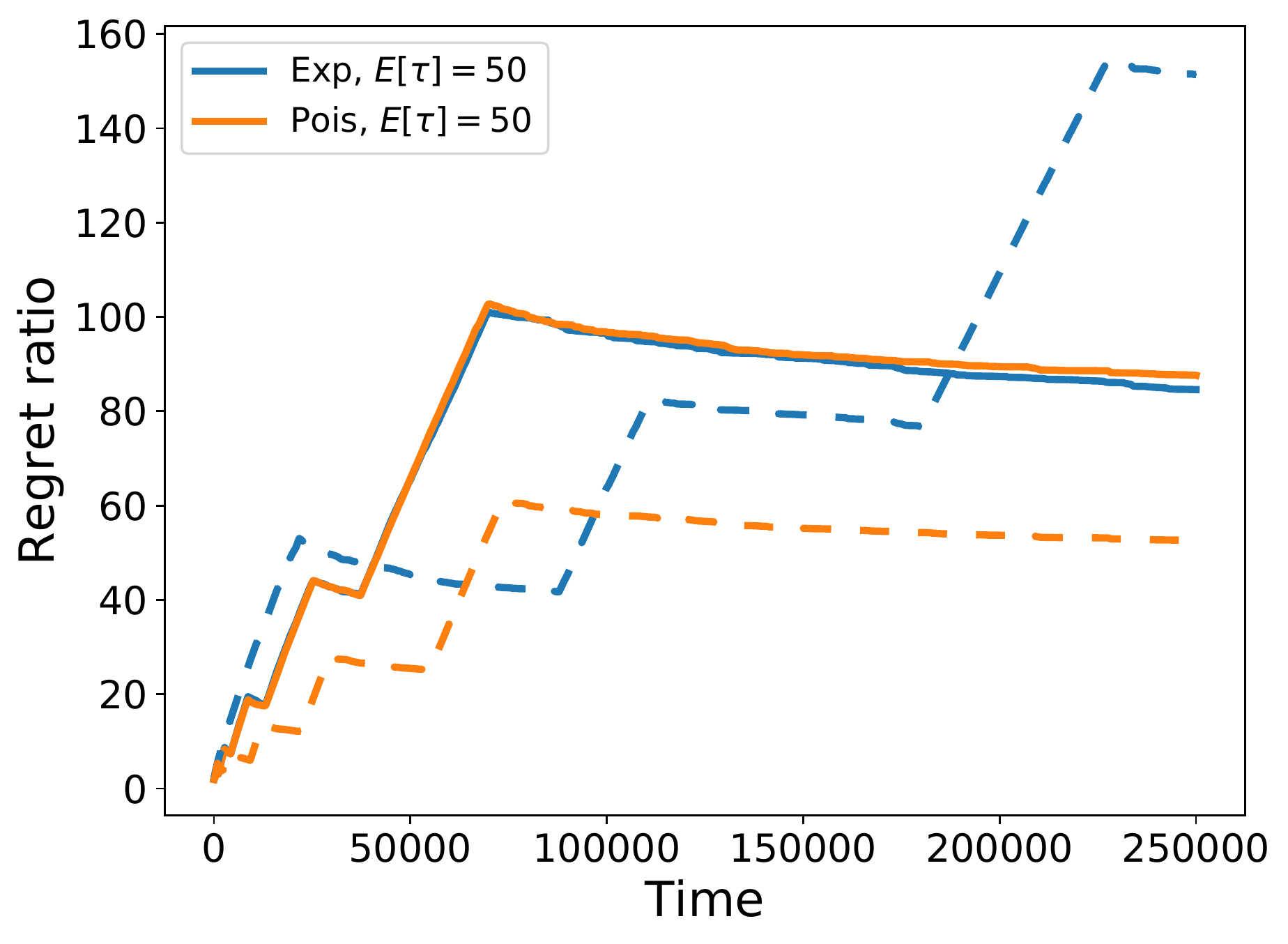}
        \caption{Unbounded delays. Ratios of regret of \ODAAF (solid lines) and \ODAAF-V (dotted lines) to that of DUCB.}
        \label{fig:ducbunbounded}
    \end{subfigure}
    \caption{The ratios of regret of variants of our algorithm to that of DUCB for different delay distributions. } \label{fig:ducb}
\end{figure}

Here we compare our algorithms, \ODAAF, \ODAAF-B and \ODAAF-V, to the (non-censored) DUCB algorithm of \citet{vernade2017stochastic}. We use the same experimental setup as described in Section~\ref{sec:experiments}. As in the comparison to QPM-D, in Figure~\ref{fig:ducb} we plot the ratios of the cumulative regret of our algorithms to that of DUCB for different delay distributions. In Figure~\ref{fig:ducbbounded}, we consider bounded delay distributions and in Figure~\ref{fig:ducbunbounded}, we consider unbounded delay distributions. From these plots, we observe that, as in the comparison to QPM-D in Figure~\ref{fig:results}, the regret ratios all converge to a constant. Thus we can conclude that the order of regret of our algorithms match that of DUCB, even though the DUCB algorithm of \citet{vernade2017stochastic} has considerably more information about the delay distribution. In particular, along with knowledge on the individual rewards of each play (non-anonymous observations), DUCB also uses complete knowledge of the cdf of the delay distribution to re-weigh the average reward for each arm. Thus, our algorithms are able to match the rate of regret of \citet{vernade2017stochastic} and QPM-D of \citet{joulani2013online} while just receiving aggregated, anonymous observations and using only knowledge of the expected delay rather than the entire cdf.

We ran the DUCB algorithm with parameter $\epsilon = 0$. As pointed out in \citet{vernade2017stochastic}, the computational bottleneck in the DUCB algorithm is evaluating the cdf at all past plays of the arms in every round. For bounded delay distributions, this can be avoided using the fact that the cdf will be 1 for plays more than $d$ steps ago. In the case of unbounded distributions, in order to make our experiments computationally feasible, we used the approximation $\PP(\tau \leq d) =1 $ for $d \geq 200$. Another nuance of the DUCB algorithm is due to the fact that in the early stages, the upper confidence bounds are dominated by the uncertainty terms, which themselves involve dividing by the cdf of the delay distributions. 
The arm that is played last in the initialization period will have the highest cdf and so it's confidence bound will be largest and DUCB will play this arm at time $K+1$ (and possibly in subsequent rounds unless the cdf increases quickly enough). 
In order to overcome this, we randomize the order that we play the arms in during the initialization period in each replication of the experiment.
Note that we did not run DUCB with half normal delays as DUCB divides by the cdf of the delay distribution and in this case the cdf would be 0 at some points.

\section{Naive Approach for Bounded Delays} \label{app:basicd}
In this section we describe a naive approach to defining the confidence intervals when the delay is bounded by some $d\geq 0$ and show that this leads to sub-optimal regret.
Let
\[w_m = \sqrt{\frac{\log(T\tilde \Delta_m^2)}{2n_m}} + \frac{md}{n_m}\,.\]
denote the width of the confidence intervals used in phase $m$ for any arm $j$.
 We start by showing that the confidence bounds hold with high probability:
\begin{lemma}\label{lem:bdhpbound}
For any phase $m$ and arm, $j$,
\[
\PP(|\bar{X}_{m,j} - \mu_j | >w_m) \leq \frac{2}{T \tilde{\Delta}_m^2}\,.
\]
\end{lemma}
\begin{proof}
First note that since the delay is bounded by $d$, at most $d$ rewards from other arms can seep into 
phase $i$ of playing arm $j$ and at most $d$ rewards from arm $j$ can be lost. 
Defining $S_{i,j}$ and $U_{i,j}$ as the start and end points of playing arm $j$ in phase $i$, respectively, we have
\begin{align}
\left| \sum_{t=S_{i,j}}^{U_{i,j}} R_{j,t} - \sum_{t=S_{i,j}}^{U_{i,j}} X_t \right| \leq d\,,
\label{eq:bddiff}
\end{align}
because we can pair up some of the missing and extra rewards, and in each pair the difference is at most one.
Then, since $T_j(m)  = \cup_{i=1}^m \{ S_{i,j}, S_{i,j}+1, \dots, U_{i,j} \}$ and using \eqref{eq:bddiff} we get
\begin{align*}
 \frac{1}{n_m} \left| \sum_{t \in T_j(m)} R_{j,t} - \sum_{t \in T_j(m)} X_t \right| \leq \frac{md}{n_m}\,.
\end{align*}
Define $\bar{R}_{m,j} = \frac{1}{|T_j(m)|} \sum_{t \in T_j(m) } R_{j,t}$ and recall that $\bar{X}_{m,j} = \frac{1}{|T_j(m)|} \sum_{t \in T_j(m)} X_t$. 
For any $a>\frac{md}{n_m}$,
\begin{align*}
\Prob{ |\bar{X}_{m,j} - \mu_j| > a} &\leq \Prob{|\bar{X}_{m,j} - \bar{R}_{m,j}| + |\bar{R}_{m,j} - \mu_j| >a}  \leq  
\Prob{|\bar{R}_{m,j} - \mu_j| > a- \frac{md}{n_m} }\\ 
&\leq 2\exp \left\{ -2n_m\left(a-\frac{md}{n_m}\right)^2 \right\}\,,
\end{align*}
where the first inequality is from the triangle inequality and the last from Hoeffding's inequality 
 since $R_{j,t} \in [0,1]$ are independent samples from $\nu_j$, the reward distribution of arm $j$. In particular, taking $a= \sqrt{\frac{\log(T\tilde \Delta_m^2)}{2n_m}} + \frac{md}{n_m}$ 
guarantees that $\Prob{|\bar{X}_j - \mu_j | >a} \leq \frac{2}{T \tilde{\Delta}_m^2}$, finishing the proof. 
\end{proof}

Observe that setting
\begin{align}
 n_m = \bigg \lceil \frac{1}{2 \tilde \Delta_m^2} \left( \sqrt{ \log(T\tilde \Delta_m^2) } + \sqrt{ \log(T\tilde \Delta_m^2) + 4 \tilde{\Delta}_m md} \right) ^2 \bigg \rceil. \label{eqn:nd}
 \end{align}
ensures that $w_m \leq \frac{\tilde \Delta_m}{2}$. Using this, we can substitute this value of $n_m$ into Improved UCB and use the analysis from \cite{auer2010ucb} to get the following bound on the regret.

\begin{theorem}
Assume there exists a bound $d\geq 0$ on the delay. Then for all $\lambda>0$, the expected regret of the Improved UCB algorithm run with $n_m$ defined as in \eqref{eqn:nd} can be upper bounded by
\[ \sum_{\substack{j \in A\\ \Delta_j > \lambda}} \Bigg( \Delta_j + \frac{64\log(T\Delta_j^2)}{\Delta_j}  + 64\log(2/\Delta_j)d + \frac{96}{\Delta_j} \Bigg)   + \sum_{\substack{j \in A \\  0<\Delta_j<\lambda}} \frac{64}{\lambda} + T \max_{\substack{j \in A \\  \Delta_j\leq\lambda}} \Delta_j \]
\end{theorem}
\begin{proof}
The result follows from the proof of Theorem 3.1 of \cite{auer2010ucb} using the above definition of $n_m$.
\end{proof} 

In particular, optimizing with respect to $\lambda$ gives worst case regret of $O(\sqrt{KT \log K} + K d\log T)$. This is a suboptimal dependence on the delay, particularly when $d >> \E[\tau]$.

\end{document}